\documentclass[10pt,twocolumn,letterpaper]{article}

\usepackage[margin=1in]{geometry}

\usepackage[utf8]{inputenc}
\usepackage[T1]{fontenc}
\usepackage{times}

\usepackage{amsmath,amssymb,amsthm}

\usepackage{graphicx}
\graphicspath{{figures/}}
\usepackage{subcaption}
\usepackage{cuted}     %
\usepackage{capt-of}   %
\usepackage{needspace}  %

\usepackage[normalem]{ulem}   %
\usepackage{booktabs}
\usepackage{multirow}
\usepackage{makecell}
\usepackage{enumitem}
\usepackage{siunitx}
\usepackage{etoc}

\usepackage{algorithm}
\usepackage{algorithmic}

\usepackage{xcolor}

\newcommand{\reductoname}{Reducto}
\newcommand{\extendname}{Extend}
\newcommand{\landingainame}{LandingAI}
\newcommand{\qualitycostpricingcites}{\cite{openai_pricing,gemini_pricing,anthropic_pricing,llamaparse_pricing,textract_pricing,google_docai_pricing,azure_doc_intel_pricing,reducto_pricing,extend_pricing,landingai_pricing}}
\newcommand{\qualitycostpricingproviders}{all evaluated providers}

\newcommand{\cmark}{$\checkmark$}

\newcommand{\pmark}{$\circ$}

\usepackage[colorlinks=true,linkcolor=blue,citecolor=blue,urlcolor=blue]{hyperref}
\usepackage[nameinlink,capitalize]{cleveref}

\usepackage[numbers,sort&compress]{natbib}
\usepackage{xfp}
\newcommand{\onedec}[1]{\num[round-mode=places,round-precision=1,round-integer-to-decimal,detect-weight=true,detect-family=true]{#1}}

\newcommand{\costeffectivetablesraw}{73.16}
\newcommand{\costeffectivechartsraw}{66.66}
\newcommand{\costeffectivecontentraw}{88.02}
\newcommand{\costeffectivesemanticraw}{73.04}
\newcommand{\costeffectivegroundingraw}{58.56}
\newcommand{\costeffectiveoverallraw}{\fpeval{round((\costeffectivetablesraw+\costeffectivechartsraw+\costeffectivecontentraw+\costeffectivesemanticraw+\costeffectivegroundingraw)/5,2)}}

\newcommand{\agentictablesraw}{90.74}
\newcommand{\agenticchartsraw}{78.11}
\newcommand{\agenticcontentraw}{89.68}
\newcommand{\agenticsemanticraw}{85.24}
\newcommand{\agenticgroundingraw}{80.62}
\newcommand{\agenticoverallraw}{\fpeval{round((\agentictablesraw+\agenticchartsraw+\agenticcontentraw+\agenticsemanticraw+\agenticgroundingraw)/5,2)}}

\newcommand{\costeffectivetables}{\onedec{\costeffectivetablesraw}}
\newcommand{\costeffectivecharts}{\onedec{\costeffectivechartsraw}}
\newcommand{\costeffectivecontent}{\onedec{\costeffectivecontentraw}}
\newcommand{\costeffectivesemantic}{\onedec{\costeffectivesemanticraw}}
\newcommand{\costeffectivegrounding}{\onedec{\costeffectivegroundingraw}}
\newcommand{\costeffectiveoverall}{\onedec{\costeffectiveoverallraw}}

\newcommand{\agentictables}{\onedec{\agentictablesraw}}
\newcommand{\agenticcharts}{\onedec{\agenticchartsraw}}
\newcommand{\agenticcontent}{\onedec{\agenticcontentraw}}
\newcommand{\agenticsemantic}{\onedec{\agenticsemanticraw}}
\newcommand{\agenticgrounding}{\onedec{\agenticgroundingraw}}
\newcommand{\agenticoverall}{\onedec{\agenticoverallraw}}

\theoremstyle{definition}

\title{ParseBench: A Document Parsing Benchmark for AI Agents}
\author{
  Boyang Zhang \quad
  Sebasti\'{a}n G. Acosta \quad
  Preston Carlson \quad
  Sacha Bron \\[2pt]
  Pierre-Lo\"{i}c Doulcet \quad
  Daniel B. Ospina \quad
  Simon Suo \\[4pt]
  \normalsize \texttt{\{boyang, sebas, preston, sacha, pierre, daniel, simon\}@runllama.ai}
}

\date{}

\begin{document}
\twocolumn[{%
\maketitle
\vspace{-1.0em}
\begin{center}
\includegraphics[width=\textwidth]{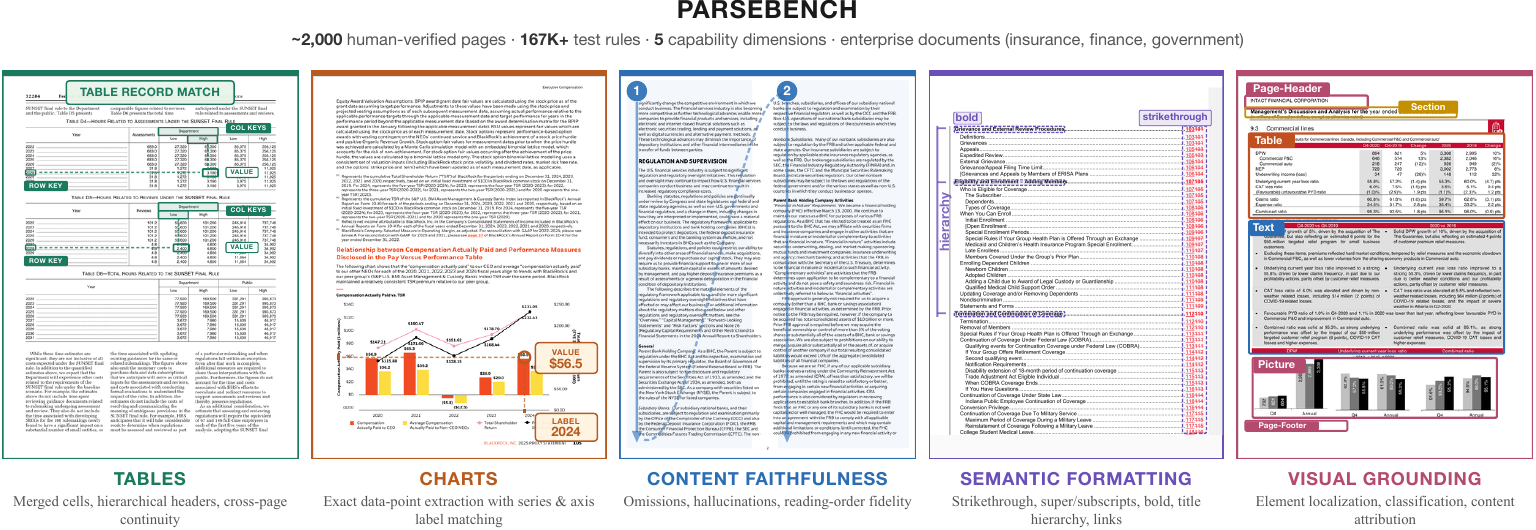}
\captionof{figure}{ParseBench comprises ${\sim}2{,}000$ human-verified enterprise document pages and more than 169K test rules, organized around five capability dimensions: tables, charts, content faithfulness, semantic formatting, and visual grounding. Each dimension targets a distinct class of agent-critical parsing failures.}
\label{fig:teaser}
\end{center}
\vspace{0.2em}
}]

\begin{abstract}

AI agents are changing the requirements for document parsing.
What matters is \emph{semantic correctness}: parsed output must preserve the structure and meaning needed for autonomous decisions, including correct table structure, precise chart data, semantically meaningful formatting, and visual grounding.
Existing benchmarks do not fully capture this setting for enterprise automation, relying on narrow document distributions and text-similarity metrics that miss agent-critical failures.
We introduce \textbf{ParseBench}, a benchmark of ${\sim}2{,}000$ human-verified pages from enterprise documents spanning insurance, finance, and government, organized around five capability dimensions: tables, charts, content faithfulness, semantic formatting, and visual grounding.
Across 14 methods spanning vision-language models, specialized document parsers, and LlamaParse, the benchmark reveals a fragmented capability landscape: no method is consistently strong across all five dimensions.
LlamaParse Agentic achieves the highest overall score at \agenticoverall\%, and the benchmark highlights the remaining capability gaps across current systems.
Dataset and evaluation code are available on \href{https://huggingface.co/datasets/llamaindex/ParseBench}{HuggingFace} and \href{https://github.com/run-llama/ParseBench}{GitHub}.

\end{abstract}

\section{Document Parsing for AI Agents}
\label{sec:intro}

\begin{table*}[t]
\centering
\small
\setlength{\tabcolsep}{3.5pt}
\begin{tabular}{@{}llrcl ccccc l@{}}
\toprule
& \textbf{Benchmark} & \textbf{Year} & \textbf{Scale} & \textbf{Domain} & \textbf{Tables} & \textbf{Charts} & \textbf{Content} & \textbf{Formatting} & \textbf{Grounding} & \textbf{Evaluation} \\
\midrule
\parbox[t]{2mm}{\multirow{4}{*}{\rotatebox[origin=c]{90}{\scriptsize Subtask}}}
& PubTabNet~\cite{zhong2020pubtabnet}        & 2020 & 568K  & Academic    & \cmark         &                &                &                &                    & TEDS \\
& FinTabNet~\cite{zheng2020gte}              & 2020 & 113K  & Financial   & \cmark         &                &                &                &                    & TEDS \\
& DocLayNet~\cite{pfitzmann2022doclaynet}    & 2022 & 80K   & Mixed$^a$   &                &                &                &                & \pmark\,$^b$       & mAP \\
& ChartQA~\cite{masry2022chartqa}            & 2022 & 21K   & Web         &                & \pmark\,$^c$   &                &                &                    & Relaxed accuracy \\
\midrule
\parbox[t]{2mm}{\multirow{4}{*}{\rotatebox[origin=c]{90}{\scriptsize E2E}}}
& OmniDocBench~\cite{ouyang2024omnidocbench} & 2024 & 1,355 & Mixed$^d$   & \cmark         &                & \cmark         &                & \pmark\,$^b$       & TEDS, edit dist. \\
& OCRBench~v2~\cite{fu2025ocrbenchv2}       & 2025 & 10K   & Mixed       & \cmark         & \pmark\,$^c$   & \cmark         &                &                    & TEDS, accuracy, F1 \\
& olmOCR-Bench~\cite{poznanski2025olmocr}    & 2025 & 1,402 & Academic$^e$ & \pmark\,$^f$  &                & \cmark         &                &                    & Binary rules \\
& \textbf{ParseBench (ours)}                 & \textbf{2026} & \textbf{2,078} & \textbf{Enterprise} & \cmark & \cmark & \cmark & \cmark & \cmark & Struct.\ match + rules \\
\bottomrule
\end{tabular}

\vspace{2pt}
{\scriptsize
\raggedright
\cmark\,=\,supported \quad \pmark\,=\,partial \\[2pt]
$^a$\,62\% enterprise (financial, legal, gov.); bounding-box detection only, no content extraction.
$^b$\,Layout detection only; no content attribution.
$^c$\,Evaluated via QA or TEDS; not data-point extraction.
$^d$\,6\% enterprise (financial reports); 62\% Chinese.
$^e$\,42\% arXiv math papers.
$^f$\,Cell adjacency checks only; no merged-cell or header hierarchy evaluation.\par
}
\caption{Comparison of document parsing benchmarks. \textit{Subtask} benchmarks evaluate a single capability on a narrow corpus; \textit{end-to-end} benchmarks target multiple capabilities jointly. ParseBench covers all five capability dimensions on enterprise documents. Scale reflects the primary annotation unit (table/chart images for subtask benchmarks, document pages for end-to-end benchmarks).}
\label{tab:benchmark-comparison}
\end{table*}

AI agents are changing the requirements for document parsing.
Earlier systems were largely built for narrower goals: making PDFs searchable for human readers, or extracting predefined fields from familiar document templates.
Enterprise agent workflows are broader in scope and less forgiving.
To automate work over financial filings, contracts, insurance documents, and regulatory submissions, an agent must parse diverse, previously unseen documents with enough fidelity to support downstream reasoning and action.
The relevant standard has therefore shifted from ``good enough to read'' to ``reliable enough to act on.''

In this setting, small parsing errors become decision errors.
An agent approving a claim may read the wrong value if a table header is misaligned.
An agent analyzing a financial report may fail if a chart is reduced to raw text rather than structured data.
Meaning-bearing formatting such as strikethrough, superscripts, or indentation is often dropped, and missing or hallucinated text can silently corrupt the context on which an agent reasons.
What matters is not whether a parser produces text that looks similar to a reference, but whether it preserves the structure and meaning needed for correct downstream decisions.
We refer to this requirement as \emph{semantic correctness}.

Existing benchmarks do not fully capture this setting for two reasons (\Cref{tab:benchmark-comparison}).
First, they underrepresent the enterprise document distributions that matter most for automation.
Benchmarks such as PubTabNet~\cite{zhong2020pubtabnet}, FinTabNet~\cite{zheng2020gte}, DocLayNet~\cite{pfitzmann2022doclaynet}, and ChartQA~\cite{masry2022chartqa} focus on narrow document types or individual subtasks.
Even broader efforts such as OmniDocBench~\cite{ouyang2024omnidocbench} and OCRBench~v2~\cite{fu2025ocrbenchv2} do not center the financial, legal, regulatory, and similarly high-stakes documents that drive many agent workflows.
Second, their metrics often overemphasize surface-level text similarity rather than semantic correctness of parsed output, penalizing minor formatting or representation differences while missing the substantive failures that matter for agents, such as broken table structure, incorrect chart values, dropped content, or missing grounding.
Recent work such as olmOCR~\cite{poznanski2025olmocr} makes important progress toward more robust evaluation, but the gap remains.
These limitations motivate ParseBench, a benchmark designed for agent-facing document parsing that evaluates both the right documents and the right notion of correctness.

ParseBench evaluates semantic correctness on ${\sim}2{,}000$ human-verified pages drawn from enterprise domains such as insurance, finance, and government.
ParseBench evaluates five capability dimensions/cmark: tables, charts, content faithfulness, semantic formatting, and visual grounding.
Its central design choice is to evaluate semantic correctness directly rather than relying on surface-level text overlap: table extraction is scored through structural record matching, chart extraction through exact data-point verification, and text- and layout-related dimensions through rule-based binary tests.
We publicly release the dataset on \href{https://huggingface.co/datasets/llamaindex/ParseBench}{HuggingFace} and the evaluation framework on \href{https://github.com/run-llama/ParseBench}{GitHub} to support reproducible and extensible benchmarking.

We evaluate 14 methods spanning vision-language models, specialized document parsers, and LlamaParse.
The benchmark reveals a fragmented capability landscape: no method is consistently strong across all five dimensions.
VLMs are competitive on content extraction, but remain weak on chart recovery and visual grounding; specialized parsers provide stronger grounding support, but often underperform on charts and semantic formatting.
Even on content faithfulness, the strongest systems reach only around 90\%, leaving non-trivial error rates for workflows in which parsing errors propagate into agent decisions.
LlamaParse Agentic achieves the highest overall score at \agenticoverall\%, and the benchmark highlights the remaining capability gaps across current systems.

\section{ParseBench}
\label{sec:parsebench}

ParseBench comprises ${\sim}2{,}000$ human-verified, annotated pages drawn from publicly available enterprise documents spanning insurance, finance, government, and other domains.
The benchmark spans five evaluation dimensions: tables, charts, content faithfulness, semantic formatting, and visual grounding.
Each dimension defines its own task-specific metrics and ground-truth format, detailed in \Cref{sec:deep-dive}.
Four principles guide the benchmark design:
\begin{enumerate}
    \item \textbf{Real-world, high-value enterprise documents.} We sample from real enterprise documents, capturing the diversity of formats, layouts, and domain-specific conventions found in the document types behind real-world automation use cases for AI agents.
    \item \textbf{Clear stratification.} We organize the benchmark along these five dimensions, each spanning from table-stakes easy to adversarially hard, giving diagnostic power over precisely where a parser breaks down.
    \item \textbf{Human-verified, yet scalable.} We verify all annotations with human reviewers. The annotation pipeline auto-labels with frontier VLMs first, then routes to human reviewers for targeted correction, minimizing manual effort while maintaining quality.
    \item \textbf{Reproducible, open, and extensible.} The dataset and metrics are fully open and easy to reproduce. The benchmark is designed to be extended along all three axes: data annotation, evaluation metrics, and provider integrations. We encourage independent replication and extension; stronger external results would validate the benchmark's usefulness, not undermine it.
\end{enumerate}

\subsection{Capability Dimensions}
\label{sec:dimensions}

\newcommand{\capabilitydim}[2]{\par\vspace{0.45em}\noindent\textbf{#1.} #2\par}

Each dimension targets a distinct failure mode that consistently breaks production agentic workflows.

\capabilitydim{Tables}{This dimension measures structural fidelity for merged cells, hierarchical headers, and cross-page continuity. A single shifted header or merged-cell error can cause the agent to extract the wrong value while failing silently in financial workflows.}

\capabilitydim{Charts}{This dimension measures exact data-point extraction with correct labels from bar, line, pie, and complex chart types. Agents need precise values rather than descriptive summaries.}

\capabilitydim{Content Faithfulness}{This dimension captures omissions, hallucinations, and reading-order mistakes. Dropped or fabricated content means the agent acts on the wrong context.}

\capabilitydim{Semantic Formatting}{This dimension measures preservation of text formatting that carries semantic significance, including strikethrough, superscript/subscript, bold, and hyperlinks.}

\capabilitydim{Visual Grounding}{This dimension measures whether each extracted element can be traced back to its precise source location, which is required for auditability in regulated workflows.}

\subsection{Dataset Curation \& Annotation}
\label{sec:curation}

\Cref{tab:dataset-stats} summarizes the benchmark. ParseBench covers over 2,000 pages from over 1,100 documents to maximize source diversity. Evaluation is dense, with over 169K test rules across the rule-based datasets.

\begin{table}[t]
\centering
\small
\begin{tabular}{@{}lp{3cm}rrr@{}}
\toprule
\textbf{Dataset} & \textbf{Metric} & \textbf{Pages} & \textbf{Docs} & \textbf{Rules} \\
\midrule
Tables & GTRM                                         & 503 & 284 & --- \\
Charts & Data Point Match Rate                       & 568 &  99 & 4,864 \\
Text   & Content Faithfulness, Semantic Formatting & 507 & 507 & 147,319 \\
Layout & Element Pass Rate                         & 500 & 321 & 16,325 \\
\midrule
\textbf{Total} &                                   & \textbf{2,078} & \textbf{1,180} & \textbf{169,011} \\
\bottomrule
\end{tabular}
\caption{ParseBench dataset statistics. The Text dataset serves both the Content Faithfulness and Semantic Formatting dimensions with different evaluation rules. Tables uses \textsc{GTRM}, a continuous metric (no discrete rules).}
\label{tab:dataset-stats}
\end{table}

\paragraph{Data sourcing \& selection.}
The curation pipeline proceeds in four stages:
\begin{enumerate}
    \item \textbf{Crawl.} Collect candidate documents from public online sources (insurance filings, financial reports, government documents, industry publications) using targeted search queries.
    \item \textbf{Detect.} Apply ML-based detection to identify pages containing relevant content per dimension (e.g., DocLayout models for chart/table detection).
    \item \textbf{Categorize.} Tag each page along key difficulty axes specific to its dimension (e.g., structural complexity for tables, value explicitness for charts, script diversity for text).
    \item \textbf{Sample.} Stratified sampling along difficulty axes to ensure coverage from straightforward to adversarially hard cases.
\end{enumerate}
We prioritize publicly available enterprise documents collected directly from the web, but we also incorporate a limited number of pages from existing public datasets when they provide relevant enterprise examples or fill coverage gaps. Source distributions are reported per dimension in the following sections.
Each dimension targets ${\sim}500$ annotated pages. Since content faithfulness and semantic formatting share the same underlying corpus (evaluated with different rule sets), the benchmark totals ${\sim}2{,}000$ unique annotated pages.
The sampling prioritizes documents with production-level complexity, such as tables with merged cells and hierarchical headers, charts requiring value estimation rather than explicit labels, text pages with dense layouts, handwriting, and multi-column structures, and pages with complex multi-element layouts, while limiting over representation from any single source or template family.

\paragraph{Ground truth generation.}

Each dimension uses a task-specific ground-truth format: full HTML tables for the table dimension, structured data points (value, labels, tolerance) for charts, Markdown transcriptions for content faithfulness and semantic formatting, and bounding box annotations with class labels for visual grounding.

All annotations are produced through a two-pass pipeline. In the first pass, frontier VLMs generate initial annotations from source PDF pages. In the second pass, human annotators verify and correct the outputs, with the review workflow tailored to each format's editability. Per-dimension annotation details and the iterative review process are described in \Cref{sec:appendix-annotation}.

\section{Dimensions Deep-Dive}
\label{sec:deep-dive}

\begin{table}[t!]
\centering
\resizebox{\columnwidth}{!}{%
\begin{tabular}{@{}lrrr@{}}
\toprule
\textbf{Category} & \textbf{Docs} & \textbf{Pages} & \textbf{\% of Pages} \\
\midrule
SERFF                          & 199 & 274 & 54.5\% \\
Public Financial Docs          &  26 & 103 & 20.5\% \\
Government Docs                &  12 &  54 & 10.7\% \\
Other Public Docs              &   9 &  24 &  4.8\% \\
FinTabNet                      &  19 &  19 &  3.8\% \\
Academic                       &   6 &  13 &  2.6\% \\
Synthetic                      &   6 &   8 &  1.6\% \\
Public Insurance (non-SERFF)   &   4 &   5 &  1.0\% \\
VRDU                           &   3 &   3 &  0.6\% \\
\bottomrule
\end{tabular}}
\caption{Table dimension: data source distribution}
\label{tab:table-sources}
\end{table}

\subsection{Tables}
\label{sec:tables}

In modern production pipelines, parsed tables are consumed primarily by automated systems and agents. Most systems we have seen share two properties.
First, they understand tables as \textit{collections of records}---where each row is a set of values keyed by their column headers. And second, they expect tables to be near-perfect, because modern systems have far fewer humans in the loop, and even a small textual difference between a source document and the parsed result can lead to a substantial semantic difference. For example, mistranscribing a 2.0\% interest rate as 0.2\% can cause a downstream financial model to be wildly inaccurate.

\paragraph{Data curation.}
Most existing table parsing benchmarks rely on artificially clean inputs: cropped table screenshots, PDF pages containing only a single table, or tables presented in isolation.
Such data do not reflect real-world documents, where tables constantly appear alongside other content.
In practice, accurate table parsing is a compound task---it requires detecting the table, semantically separating it from surrounding elements, determining whether to split or merge adjacent tables, all in addition to correctly identifying table structure and content.

ParseBench is designed to reflect real-world conditions.
Every table remains embedded in its original PDF page, preserving the full visual and structural context a production parser must navigate.
The dataset skews toward insurance and financial tables, which are among the most commonly parsed in high-value enterprise automation / agentic use cases. (\Cref{tab:table-sources}).

\begin{figure}[t!]
\centering
\includegraphics[width=\columnwidth]{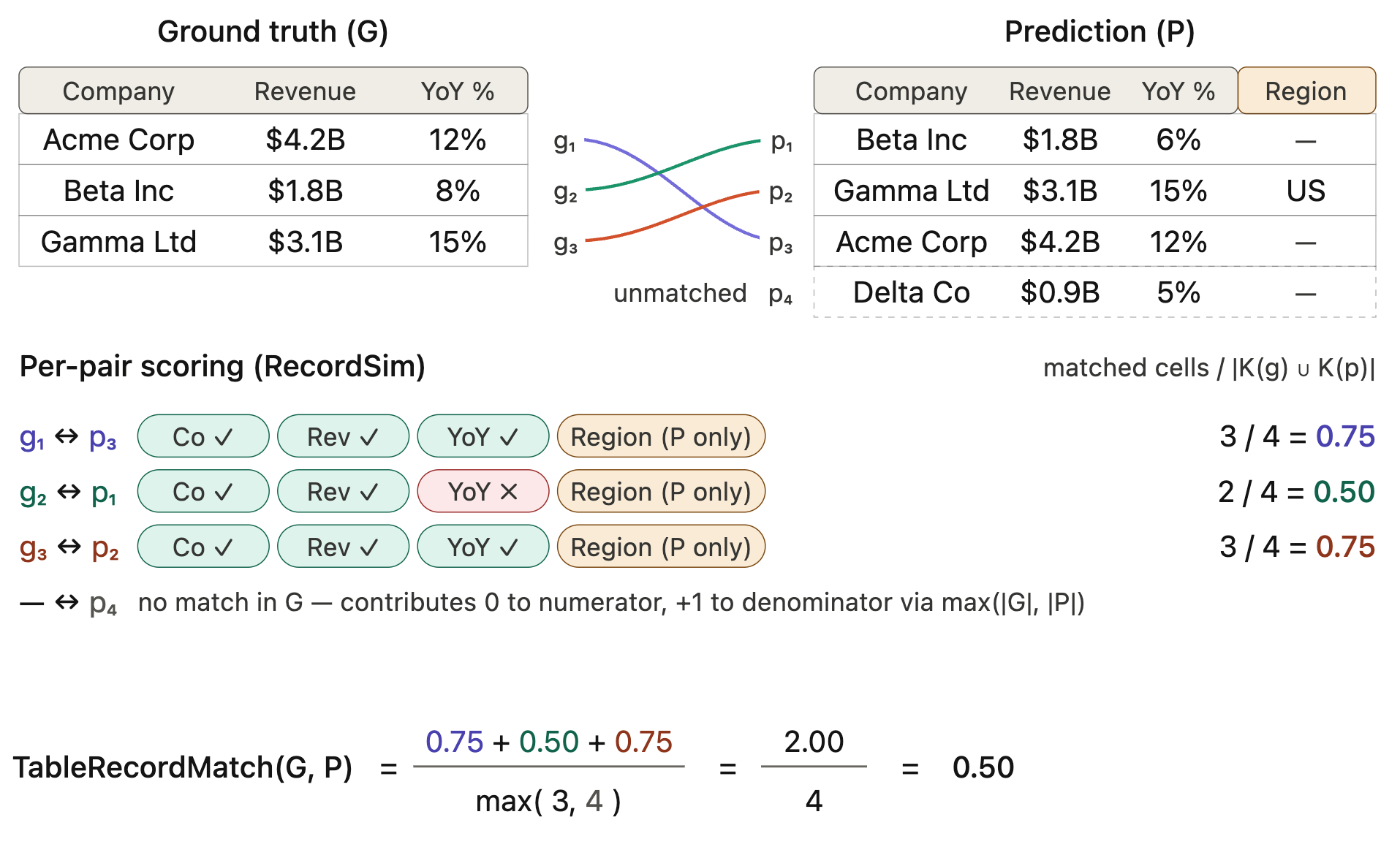}
\caption{Illustration of the \textsc{TableRecordMatch} metric. Predicted records and columns are matched to those in the ground truth. Each matched record is scored by binary cell-level agreement.}
\label{fig:table-metric}
\end{figure}

\paragraph{Metric definition.}
Existing table accuracy metrics, like GriTS~\cite{smock2023gritsgridtablesimilarity} and TEDS~\cite{zhong2020imagebasedtablerecognitiondata}, naturally generalize across arbitrarily complex tables. However, they are insufficient for measuring parsing quality for modern production use cases, because they place excessive weight on table structure, and underweight the importance of header cells. For example, they heavily penalize swapping column and row order---a transformation that preserves every key--value relationship---while assigning mild penalties to dropping all column headers or transposing headers across columns, either of which renders table data essentially useless to downstream systems.

To address these deficiencies, we define the \textsc{TableRecordMatch} metric (\Cref{fig:table-metric}).

\textsc{TableRecordMatch} treats a table as a bag of records: each row is a record whose cell values are keyed by the corresponding column header(s). Ground-truth records are matched to predicted records, and each matched pair is scored by binary cell-level agreement (\Cref{eq:recsim}). \textsc{TableRecordMatch} is insensitive to differences in column and row order, because such differences don't alter the bag of records a table corresponds to. Meanwhile, dropped and transposed headers cause enormous mismatches in record keys-value pairs, and are penalized accordingly.

Let $G$ and $P$ denote the bags of ground-truth and predicted records, and let $\mathcal{M}$ be the optimal matching between them. For a record $r$, let $K(r)$ denote its set of column keys and let $r[k]$ be its value at key $k$. Then:
\begin{gather}
\text{TableRecordMatch}(G, P) = \frac{\sum_{(g,p) \in \mathcal{M}} \text{RecordSim}(g, p)}{\max(|G|, |P|)} \label{eq:trm} \\
\text{RecordSim}(g, p) = \frac{\sum_{k \in K(g) \cap K(p)} \mathbf{1}\big[g[k] = p[k]\big]}{|K(g) \cup K(p)|} \label{eq:recsim}
\end{gather}
That is, two records are compared by counting cells that share a key and have equal values, normalized by the union of their keys, so that unmatched columns on either side are penalized. The overall metric averages these per-pair scores over the larger of the two record bags, so unmatched records are also penalized.

We combine both perspectives into a single score. We define \textsc{GTRM} as the unweighted average of GriTS and \textsc{TableRecordMatch}:
\begin{equation}
\textsc{GTRM} = \frac{\text{GriTS} + \textsc{TableRecordMatch}}{2}
\label{eq:gtrm}
\end{equation}

We note that some tables, such as those with multiple discontiguous header rows or headers on both the top and side of the table, have semantics that are not captured well by \textsc{TableRecordMatch}. In such cases, we report \text{GriTS} instead of \textsc{GTRM} for all providers. These cases are identified by manual review and marked in the publicly released dataset as \texttt{trm\_unsupported} .

\begin{figure}[t!]
    \centering
    \includegraphics[width=\columnwidth]{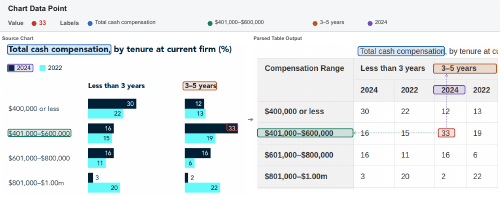}
    \captionof{figure}{Illustration of the \textsc{ChartDataPointMatch} evaluation: a chart is parsed into a table, and annotated data points are verified by matching expected values and labels against the table's rows and columns.}
    \label{fig:charts-labels}
\end{figure}

\subsection{Charts}
\label{sec:charts}

Effective chart extraction requires converting visuals into structured tables with precise numerical data and labels. This serves two critical needs: providing machine-readable data for AI agents and indexing, and offering human-ready spreadsheets for practitioners in data-heavy fields (i.e.\ finance, energy, healthcare, and marketing), replacing manual transcription with instant, actionable insights.
Converting a chart to a table is inherently imprecise, both in value and granularity: should a continuous line chart representing a stock value be represented daily, weekly, or monthly?
While datasets like ChartQA~\cite{masry2022chartqa} advance the field, they lack the visual diversity and richness of axis labels and legends found in ``in-the-wild'' charts.

\paragraph{Data curation.}
Existing academic benchmarks fall short in two ways.
First, \textbf{visual homogeneity}: they rely on uniform plotting libraries like Matplotlib, whereas real-world PDFs exhibit extreme heterogeneity---custom BI palettes, stylized corporate infographics, low-resolution, and scanning artifacts.
Second, \textbf{compositional complexity}: they typically isolate one chart per image, while real-world documents feature multi-chart panels, shared legends, inset axes, and plots tightly interwoven with dense text.

We focus on the four chart types most prevalent in data-heavy fields: bar, line, pie, and compound (mixed of bar, point and line elements). These are the only types present in our dataset; rather than chasing breadth across rarer chart types, we maximize diversity \emph{within} these four to capture the visual heterogeneity and compositional complexity described above.
Since the surrounding page context provides additional information (titles, captions, legends), we parse entire PDF pages rather than cropped chart images.
Concretely, we ensure a mix of charts with:
\begin{itemize}
    \item Charts with explicitly written values and charts without any.
    \item Discrete series and continuous series.
    \item Few values (e.g., 2~bars) and high data density (e.g., 10~lines over 20~months). %
    \item Clear markers or inflection points and smooth continuous appearances.
    \item Pages with a single simple chart and pages with multiple related charts sharing axes.
\end{itemize}
See \Cref{sec:appendix-chart-dimensions} for the resulting distribution across these dimensions.

\paragraph{Data annotation.}
Rather than annotating full ground-truth tables (which would be both costly and brittle due to the imprecision inherent in chart reading), we annotate a set of \emph{spot-check data points} per chart (up to 10).
Each data point specifies a numerical value and one or more labels (e.g., series name, x-axis category, chart title) that should be locatable in the parser's table output.
Each data point contains a tolerance: charts with explicit value annotations (e.g., bar labels) admit an exact match, while charts without them require a relative tolerance (default 1\%), since recovering an exact value purely from visual estimation is virtually impossible.
Annotations were initially generated by a Gemini~3.0 Flash agent and verified by human annotators.

\paragraph{Metric definition.}
We define \textsc{ChartDataPointMatch} as the proportion of annotated data points successfully verified in the parser's table output.
A data point is verified if its annotated value and all associated labels can be located in the table.
This design makes the metric resilient in several ways: it is insensitive to the output table orientation (rows and columns can be swapped), and tolerant of numeric formatting differences (currency symbols, unit suffixes, thousands separators, decimal separators). The evaluation process follows four steps:
\\
\begin{enumerate}
    \item \textbf{Parse:} Extract Markdown/HTML tables from the parser output.
    \item \textbf{Locate context:} Find the relevant table output using label-matching on surrounding context (bold text, headings, captions).
    \item \textbf{Match values:} Find cells matching the expected value using fuzzy string matching combined with numeric normalization (currency symbols, suffixes like k/M/B, thousands separators, decimal separators, percentages) within a configurable relative tolerance.
    \item \textbf{Verify labels:} For each matched value, verify that all remaining labels are correctly associated with the proper row or column.
\end{enumerate}

\subsection{Content Faithfulness}
\label{sec:content-faithfulness}

Content faithfulness measures whether parsed output preserves \emph{all and only} the content present in the source document: no dropped text, no hallucinated text, and correct reading order.
This is the most fundamental requirement for agentic workflows. If the agent's context is incomplete or contains fabricated content, every downstream decision is compromised.
Unlike text-similarity metrics that tolerate minor deviations, content faithfulness demands that every sentence, every number, and every structural relationship survives parsing intact.

\paragraph{Score definition.}
To produce document-length-invariant scores, we aggregate individual rule results through two levels of averaging.
Let $R_{d,t}$ denote the set of rule results of type $t$ for document $d$, where each rule result carries a score in $[0,1]$.
The \textbf{per-type score} averages over all rules of a given type:
\begin{equation}
s_{d,t} = \frac{1}{|R_{d,t}|} \sum_{r \in R_{d,t}} \text{score}(r)
\label{eq:per_type}
\end{equation}
This ensures that a document with many rules of one type does not dominate the category score.
Each category $C$ groups related rule types (e.g., the \emph{text correctness} category includes \texttt{missing\_sentence}, \texttt{unexpected\_sentence}, \texttt{present}, and \texttt{absent} rules).
The \textbf{per-document category score} averages over the rule types present in that document:
\begin{equation}
S_{d,C} = \frac{1}{|\{t \in C : R_{d,t} \neq \emptyset\}|} \sum_{\substack{t \in C \\ R_{d,t} \neq \emptyset}} s_{d,t}
\label{eq:category}
\end{equation}
The benchmark-level category score is then the mean of $S_{d,C}$ across all documents.

The composite \textbf{Content Faithfulness Score} combines text correctness and reading order with fixed weights $w_{\text{text}} = 1.0$ and $w_{\text{order}} = 0.5$:
\begin{equation}
\text{CFS}_d = \frac{w_{\text{text}} \cdot S_{d,\text{text}} + w_{\text{order}} \cdot S_{d,\text{order}}}{\sum_{C \in \mathcal{P}_d} w_C}
\label{eq:cfs}
\end{equation}
where $\mathcal{P}_d \subseteq \{\text{text}, \text{order}\}$ is the set of categories for which document $d$ has at least one rule, and the denominator sums only over those present categories.
We down-weight reading order because text correctness is the more critical failure mode.
The remainder of this section details how data is curated, annotated, and how individual rule types are defined.

\paragraph{Data curation.}

We randomly sample 500 PDF documents from the internet, using Google to source them.
We select one page per document, removing pages with significant non-textual elements (tables, charts) to avoid biasing the dataset.
Finally, we categorize documents as shown in \Cref{tab:text-categories}.

\begin{table}[t]
\centering
\resizebox{\columnwidth}{!}{%
\begin{tabular}{@{}llr@{}}
\toprule
\textbf{Category} & \textbf{Description} & \textbf{Docs} \\
\midrule
text\_simple       & Simple text with some styling       & 170 \\
text\_ocr          & Scanned/image docs, various quality & 129 \\
text\_multicolumns & 1--8 columns, different layouts     & 117 \\
text\_multilang    & 20+ languages, all major scripts    &  47 \\
text\_misc         & Unusual content/layout/reading order &  33 \\
text\_dense        & Dense, large docs (e.g., newspapers) & 24 \\
text\_handwriting  & Significant handwritten text         &  23 \\
text\_formatting   & Heavy text styling                   &  10 \\
\bottomrule
\end{tabular}}
\caption{Content faithfulness and semantic formatting: shared document categories. The same corpus serves both dimensions, with different evaluation rules applied.}
\label{tab:text-categories}
\end{table}

\paragraph{Data annotation.}
Rather than writing evaluation rules directly, we first transcribe each document into a complete Markdown ground truth, from which rules are automatically generated (see \emph{Metric definition} below).
This approach optimizes for annotation efficiency: Markdown is easy for humans to read and edit while remaining parseable, and it naturally captures formatting, title hierarchy, and reading order.
We use a VLM-assisted iterative pipeline in which a human annotator reviews and corrects machine-generated transcriptions until convergence (details in \Cref{sec:appendix-annotation}).

\paragraph{Metric definition.}
Content faithfulness is evaluated through rule-based metrics that detect omissions, hallucinations, and reading-order violations, following the same general evaluation philosophy as olmOCR-Bench~\cite{poznanski2025olmocr} while extending it to the document-parsing setting studied here.
Each document is annotated with a set of test rules derived from its ground-truth content.
A rule produces a score in $[0, 1]$: binary rules yield $1.0$ (pass) or $0.0$ (fail), while percentage rules yield a continuous value.
All text comparisons are performed after a normalization pipeline that strips Markdown formatting, canonicalizes Unicode, and collapses whitespace.

\subparagraph{Text correctness.}
Rules measure whether the parser faithfully reproduces textual content, detecting both \emph{omissions} (dropped content) and \emph{hallucinations} (fabricated content).
We apply recall, precision, and duplication checks at word, sentence, and digit granularities.
Word- and sentence-level rules capture missing or unexpected content at different scales, while digit-level rules (\texttt{bag\_of\_digit\_percent}) compare digit frequency distributions to catch common OCR errors (e.g., \texttt{6}~$\to$~\texttt{8}).
Note that frequency-based comparison cannot detect swaps between equally frequent digits (e.g., if every \texttt{6} becomes \texttt{8} and vice versa, the distribution is unchanged); we accept this limitation as digit-level checks remain effective for the more common case of isolated substitutions and insertions.

\subparagraph{Reading order.}
Document parsing systems must linearize two-dimensional page layouts into a one-dimensional text stream, making reading-order preservation critical.
We evaluate reading order through pairwise precedence assertions: each \texttt{order} rule specifies two text fragments, \emph{before} and \emph{after}, and passes if and only if the first occurrence of \emph{before} precedes the last occurrence of \emph{after} in the linearized output.
We deliberately use first/last rather than first/first matching: when a fragment is repeated (e.g., a heading that reappears in a table of contents), the lenient criterion avoids penalising correct body-text order for duplications outside the parser's control.
To ensure full coverage, we annotate the full document content (excluding tables), yielding $N_{\text{sentences}} - 1$ order rules per document.

\subsection{Semantic Formatting}
\label{sec:semantic-formatting}

Semantic formatting measures whether the parser preserves inline formatting that \emph{carries meaning}, as opposed to purely presentational styling.
We focus on formatting classes whose omission changes the semantics of the document:
\begin{itemize}
    \item \textbf{Strikethrough} marks superseded or deleted content; losing it causes the agent to treat invalidated text as current.
    \item \textbf{Superscript/subscript} denote footnote references, chemical formulae, and mathematical notation; flattening them loses critical distinctions (e.g., $x^2$ vs.\ $x2$).
    \item \textbf{Bold} often signals defined terms, section labels, or key values in structured documents (e.g., totals in financial reports).
\end{itemize}
Most existing benchmarks ignore formatting entirely or treat it as cosmetic.
In agentic workflows, however, formatting directly influences the agent's interpretation: a strikethrough price is not the current price, and a superscript ``1'' is a footnote reference rather than a quantity (\Cref{fig:formatting-matters}).

\begin{figure}[t!]
\centering
\small
\setlength{\tabcolsep}{4pt}
\renewcommand{\arraystretch}{1.4}
\begin{tabular}{@{}p{0.28\columnwidth}p{0.28\columnwidth}p{0.34\columnwidth}@{}}
\toprule
\textbf{Source PDF} & \textbf{Stripped output} & \textbf{Agent misreading} \\
\midrule
\sout{\$49.99} \textbf{\$39.99}
  & \$49.99 \$39.99
  & Two valid prices; may quote the old one \\
H\textsubscript{2}O
  & H2O
  & Treats ``H2O'' as an opaque identifier \\
See note\textsuperscript{1}
  & See note1
  & Reads ``note1'' as a single token, footnote lost \\
\textbf{Total: \$12{,}450}
  & Total: \$12{,}450
  & Misses that this is a key aggregate value \\
\bottomrule
\end{tabular}
\caption{Formatting loss changes document semantics. Each row shows a source snippet, what a formatting-unaware parser emits, and the resulting misinterpretation by a downstream agent.}
\label{fig:formatting-matters}
\end{figure}

\paragraph{Score definition.}
The \textbf{Semantic Formatting Score} combines four sub-metrics --- text styling, title accuracy, LaTeX, and code blocks --- using a frequency-adjusted weighted average:
\begin{equation}
\text{SFS} = \frac{1.0 \cdot S_{\text{style}} + 1.0 \cdot S_{\text{title}} + \frac{1}{5} S_{\text{latex}} + \frac{1}{5} S_{\text{code}}}{W}
\label{eq:sfs}
\end{equation}
where $W$ is the sum of weights over categories actually present in the document:
\begin{gather}
W = \sum_{c \in \mathcal{C}} w_c \cdot \mathbf{1}[c \text{ present}], \notag \\
w_{\text{style}} = w_{\text{title}} = 1.0,\quad w_{\text{latex}} = w_{\text{code}} = \tfrac{1}{5}
\label{eq:sfs-w}
\end{gather}
LaTeX and code blocks are down-weighted because they appear in only a small fraction of documents; without adjustment, their binary pass/fail outcomes would disproportionately influence the aggregate.
The following paragraphs detail each sub-metric.
The semantic formatting dimension shares the same document corpus as content faithfulness (\Cref{tab:text-categories}), with a different set of evaluation rules applied to each document.

\subparagraph{Text styling.}
Styling rules verify that semantically meaningful inline formatting is correctly preserved in the Markdown output.
We score four formatting classes: strikethrough, superscript, subscript, and bold.
Each class has both positive (e.g., \texttt{is\_strikeout}) and negative (e.g., \texttt{is\_not\_strikeout}) tests, catching both stripped and falsely applied styling.
We also evaluate additional formatting classes (italic, underline, highlight); see \Cref{app:styling-classes} for full details.

The per-document styling score combines positive and negative pass rates using a weighted harmonic mean (analogous to the $F_\beta$-score):
\begin{equation}
S_{\text{style}}^{(d)} = \frac{(1 + \beta^2)\;\bar{s}^{+}_{d}\;\bar{s}^{-}_{d}}{\beta^2\;\bar{s}^{+}_{d} + \bar{s}^{-}_{d}}
\label{eq:styling}
\end{equation}
where $\bar{s}^{+}_{d}$ is the mean pass rate over positive rules, $\bar{s}^{-}_{d}$ over negative rules, and $\beta$ controls the relative importance of the two.
We set $\beta = 0.5$ to penalise false styling more heavily than missed styling: the score remains low whenever either rate is low, but negative-rule failures carry greater weight.

\subparagraph{Title accuracy.}
\texttt{is\_title} checks that a text fragment appears as a heading; \texttt{title\_hierarchy\_percent} scores whether detected titles respect the expected parent--child hierarchy.

\subparagraph{LaTeX and code blocks.}
\texttt{is\_latex} checks that mathematical formulae appear in LaTeX notation; \texttt{is\_code\_block} checks for fenced code blocks with correct language annotations.

\subsection{Visual Grounding}
\label{sec:grounding}

Visual grounding measures whether a system can connect generated document content back to the correct region on the page.
A parser can produce readable Markdown while still failing at grounding if it assigns the right words to the wrong visual region.
We evaluate visual grounding as a joint problem over \textbf{localization}, \textbf{classification}, and \textbf{attribution}.
This matters for both agents and human reviewers: extracted claims, values, and tables remain auditable only when they can be traced back to the correct source region.

\paragraph{Data curation.}
\begin{itemize}
    \item \textbf{Common label space.} The benchmark uses a compact visual grounding label set composed of \texttt{Text}, \texttt{Table}, \texttt{Picture}, \texttt{Page-Header}, and \texttt{Page-Footer}. Provider-specific labels are collapsed into this shared set so cross-model comparison remains fair.
    \item \textbf{Content-eligible elements.} Attribution is evaluated only for elements with a human-verified content association whose expected literal text is well-defined enough to grade directly. In the current dataset, elements marked \texttt{explicit} are chart-like picture regions where some visible values or labels are expected but the ground truth does not exhaust every valid description, so attribution uses recall-oriented scoring. Elements whose attributes include \texttt{caption=true} are primarily picture- and chart-like regions whose associated text is descriptive rather than literal, so attribution is skipped; this is distinct from the layout label \texttt{Caption}. Formulas are excluded because equivalent LaTeX can be rendered in many valid forms, and elements marked \texttt{ignore} are auxiliary non-gradeable regions excluded from evaluation entirely.
    \item \textbf{Layout-level reading order.} The dataset stores reading order at the layout level, enabling analysis of whether extracted blocks are correctly sequenced.
\end{itemize}

\paragraph{Metric definitions.}
The primary metric is the \textbf{Element Pass Rate}: a ground-truth element passes only if localization, classification, and, when applicable, attribution all pass for the same element. We use this joint pass rate because visual grounding is only correct when the system finds the right region, assigns the right label, and attaches the right content. Reporting the components separately is still useful, but averaging them would over-credit partial successes that still break provenance. The three component checks are:
\begin{itemize}
  \item \textbf{Localization.} Evaluates whether the parser identifies the correct region for the target element. We use asymmetric intersection-over-area, defined as $\mathrm{IoA}(A, B) = |A \cap B| / |A|$, rather than IoU, because IoA better tolerates split and merge behavior while still requiring meaningful spatial alignment. A ground-truth element is considered localized when its best-matching prediction satisfies $\mathrm{IoA}(\mathrm{GT},\mathrm{Pred}) \ge 0.50$ and $\mathrm{IoA}(\mathrm{Pred},\mathrm{GT}) \ge 0.20$. The first threshold requires substantial coverage of the ground-truth region; the second rejects very large loose boxes while still tolerating split and merge behavior. For \texttt{Page-Header} and \texttt{Page-Footer}, we apply the same principle to a grouped top/bottom furniture band so wide ground-truth bands are not unfairly penalized when a system emits multiple smaller fragments, or the reverse.

  \item \textbf{Classification.} Evaluates whether the best localized prediction is assigned the correct semantic label after provider-specific labels are collapsed into the shared set.

    \item \textbf{Attribution.} Evaluates whether the content assigned to that region matches the expected content when such a human-verified literal content association exists. Candidate predictions are gathered using $\mathrm{IoA}(\mathrm{GT},\mathrm{Pred}) \ge 0.30$, and text is normalized before comparison. Standard content-bearing elements pass when token-level $\mathrm{F1} \ge 0.80$, while the chart-like \texttt{explicit} cases in the current dataset use recall at the same threshold. For \texttt{Page-Header} and \texttt{Page-Footer}, attribution is matched against the best ordered grouped span inside the recovered furniture band rather than a single fragment. Details on \texttt{explicit}, the \texttt{caption} attribute, \texttt{ignore}, formula-specific skips, merge-aware filtering, and auxiliary attribution diagnostics are provided in \Cref{app:grounding-aux-metrics}.
\end{itemize}
Let $g_i$ denote the $i$-th ground-truth element, let $L_i$, $C_i$, and $A_i$ denote binary localization, classification, and attribution outcomes, and let $E_i$ indicate whether attribution is applicable. Then:
\begin{equation}
\begin{aligned}
\mathrm{Pass}(g_i) &= L_i \cdot C_i \cdot \bigl((1 - E_i) + E_i A_i\bigr), \\
\mathrm{EPR} &= \frac{1}{N}\sum_{i=1}^{N}\mathrm{Pass}(g_i),
\end{aligned}
\label{eq:grounding-epr}
\end{equation}
When $E_i = 0$, the element-level pass reduces to localization and classification alone.

We also report standard COCO-style detection metrics~\cite{lin2014coco} and define auxiliary attribution diagnostics (\textbf{LAP}, \textbf{LAR}, \textbf{AF1}) for further analysis; details are provided in \Cref{app:grounding-aux-metrics}. These diagnostics isolate attribution behavior under the same grounding policy, but they are not averaged into the main Element Pass Rate reported in \Cref{tab:main-results}.

\section{Experiments}
\label{sec:results}

\subsection{Experimental Setup}
\label{sec:setup}
We evaluate three categories of methods:
(1)~VLMs prompted directly with document pages, both proprietary and open-weight, including general-purpose and OCR-specialized models;
(2)~specialized document parsers, including commercial products and open-source pipelines;
and (3)~LlamaParse, our system. \
Full per-provider configurations are detailed in \Cref{tab:provider-configs} (Appendix).

\paragraph{VLMs.}
For the main comparison, we configure proprietary VLMs near a common low-cost operating point of roughly one cent per page when possible, so the results reflect capability under a practically relevant budget rather than unconstrained inference spend.
We use minimal reasoning/thinking settings for Gemini~3 Flash~\cite{gemini3flash2025} and GPT-5 Mini~\cite{openai2025gpt5}, as their defaults exceed this target. We use default settings for Haiku~4.5~\cite{anthropic2025haiku45}, as its cost already exceeds the target without thinking.
Full cost analysis in \Cref{app:cost-analysis}.
We run open-weight VLMs (Qwen~3~VL 8B~\cite{bai2025qwen3vl}, Dots~OCR~1.5~\cite{dots_ocr}) on NVIDIA H100 GPUs via vLLM~\cite{kwon2023vllm} on Modal~\cite{modal}.
Full details in \Cref{app:infra}.

All VLMs are prompted to produce both parsed content and layout annotations for evaluation across all five dimensions. Proprietary VLMs use a single combined prompt; Dots~OCR~1.5 uses its officially recommended prompt~\cite{dots_ocr}; Qwen~3~VL uses separate parse and layout pipelines, as combining both tasks in one prompt significantly degrades its performance. Full prompts are in \Cref{app:vlm-prompts}.

\paragraph{Specialized document parsers.}
We evaluate commercial APIs with their default or recommended configurations.
We run Docling with default configuration on an H100.
Where multiple modes are available (e.g., Azure Layout vs.\ Read mode), we select the one best suited for structured extraction.
Full details in \Cref{app:provider-configs}.

\paragraph{LlamaParse.}
We evaluate two configurations: \emph{Cost Effective}, a single-pass pipeline optimized for throughput and cost, and \emph{Agentic}, a multi-step pipeline that orchestrates VLMs with specialized tools and iterative refinement for higher accuracy.

\begin{table*}[t!]
\centering
\small
\begin{tabular*}{\textwidth}{@{\extracolsep{\fill}}lcccccc@{}}
\toprule
\textbf{Provider} & \textbf{Overall} & \textbf{Tables} & \textbf{Charts} & \makecell{\textbf{Content}\\\textbf{Faithful.}} & \makecell{\textbf{Semantic}\\\textbf{Format.}} & \makecell{\textbf{Visual}\\\textbf{Ground.}} \\
\midrule
\multicolumn{7}{@{}l}{\textit{VLMs}} \\
\quad OpenAI GPT-5 Mini      & 46.8 & 69.8 & 30.1 & 82.3 & 45.8 & 6.2 \\
\quad Anthropic Haiku 4.5    & 45.2 & 77.2 & 13.8 & 78.7 & 49.4 & 6.7 \\
\quad Google Gemini 3 Flash  & 71.0  & \underline{89.9} & 64.8 & 86.2 & 58.4 & 56.0 \\
\quad Qwen 3 VL              & 62.0 & 74.7 & 28.2 & 87.6 & 64.2 & 55.2\textsuperscript{*} \\
\quad Dots OCR 1.5            & 55.8 & 85.2 &  0.9 & \textbf{90.0} & 47.0 & 55.8 \\
\midrule
\multicolumn{7}{@{}l}{\textit{Specialized Document Parsers}} \\
\quad Docling (OSS)           & 50.6  & 66.4 & 52.8 & 66.9 &  1.0 & 66.1\textsuperscript{$\dagger$} \\
\quad AWS Textract            & 47.9  & 84.6 &  6.0 & 74.8 &  3.7 & 70.4 \\
\quad Google Cloud Doc AI     & 50.4 & 55.1 &  1.4 & 83.7 & 50.5 & 61.3 \\
\quad Azure Doc Intelligence  & 59.6  & 86.0 &  1.6 & 84.9 & 51.9 & \underline{73.8} \\
\quad \reductoname                 & 67.8  & 70.3 & 57.0 & 86.4 & 56.8 & 68.7 \\
\quad \extendname                  & 55.8 & 85.1 &  1.6 & 84.1 & 47.4 & 60.7 \\
\quad \landingainame               & 45.2 & 73.7 & 10.9 & 88.6 & 27.9 & 25.1 \\
\midrule
\multicolumn{7}{@{}l}{\textit{LlamaParse (Ours)}} \\
\quad Cost Effective  & \underline{\costeffectiveoverall} & \costeffectivetables & \underline{\costeffectivecharts} & \costeffectivecontent & \underline{\costeffectivesemantic} & \costeffectivegrounding \\
\quad Agentic  & \textbf{\agenticoverall} & \textbf{\agentictables} & \textbf{\agenticcharts} & \underline{\agenticcontent} & \textbf{\agenticsemantic} & \textbf{\agenticgrounding} \\
\bottomrule
\end{tabular*}
\caption{Main ParseBench results (\%). Overall is the unweighted mean across all five dimensions. \textbf{Bold} marks the best score across all providers in each column; \underline{underlined values} mark the second-best. \textsuperscript{*}For visual grounding, Qwen uses a separate layout-only pipeline; 4 pages where that pipeline failed to return usable output are excluded. See Appendix~\ref{app:vlm-prompts}. \textsuperscript{$\dagger$}Docling's visual-grounding score excludes 13 pages with pipeline failures.}
\label{tab:main-results}
\end{table*}

\subsection{Main Results}
\label{sec:main-results}

\Cref{tab:main-results} presents the main benchmark results across all five capability dimensions.
At the top line, LlamaParse leads overall, with Agentic at \agenticoverall\% and Cost Effective at \costeffectiveoverall\%. The strongest external baselines are Gemini~3 Flash (71.0\%) and \reductoname{} (67.8\%).
The per-dimension breakdown reveals distinct capability profiles:
\begin{enumerate}
    \item \textbf{Tables} are competitive at the top end, and the main differentiation comes from the long tail of adversarially hard documents (deeply nested headers, merged cells, cross-page continuity). LlamaParse Agentic (\agentictables\%) and Gemini (89.9\%) lead, while the gap widens on hard cases.
  \item \textbf{Charts} are the most polarizing dimension. LlamaParse Agentic leads at \agenticcharts\%, and only five providers exceed 50\%, while most specialized parsers score below 6\%. Most existing parsers either skip charts entirely or output raw OCR text rather than the structured data tables that downstream agents require. %

    \item \textbf{Content faithfulness} is the most table-stakes dimension. Strong performance here is critical for any downstream agent workflow, since omissions, hallucinations, and ordering errors directly corrupt the agent's context. The top scores come from Dots~OCR~1.5 (90.0\%) and LlamaParse Agentic (\agenticcontent\%).

    \item \textbf{Semantic formatting} is one of the clearest separators in the benchmark. Most parsers treat formatting as cosmetic, stripping strikethrough, superscripts, and hyperlinks that carry semantic meaning in business documents. LlamaParse Agentic leads (\agenticsemantic\%), followed by LlamaParse Cost Effective (\costeffectivesemantic\%) and Qwen~3~VL (64.2\%).

    \item \textbf{Visual grounding} is difficult for proprietary single-pass VLMs: GPT-5 Mini and Haiku~4.5 both score below 10\%. Layout-aware systems and agentic pipelines perform much better, with LlamaParse Agentic leading (\agenticgrounding\%), followed by Azure (73.8\%) and Textract (70.4\%). Reported Qwen and Docling grounding scores omit pages where their layout pipelines failed to produce usable grounding output: 4 pages for Qwen's separate layout-only pipeline and 13 for Docling. Appendix~\ref{app:layout-metric-breakdown} breaks these layout results down into localization, classification, attribution, reading order, and standalone detection/attribution diagnostics.
\end{enumerate}

Taken together, the benchmark reveals a consistent pattern: VLMs tend to excel at content-level understanding, while specialized parsers tend to excel at structure- and layout-aware extraction. The difficulty is that downstream document agents need both, and no external baseline is consistently strong across all five dimensions. LlamaParse Agentic comes closest to that combined profile.

\begin{figure*}[t!]
\centering
\includegraphics[width=\textwidth]{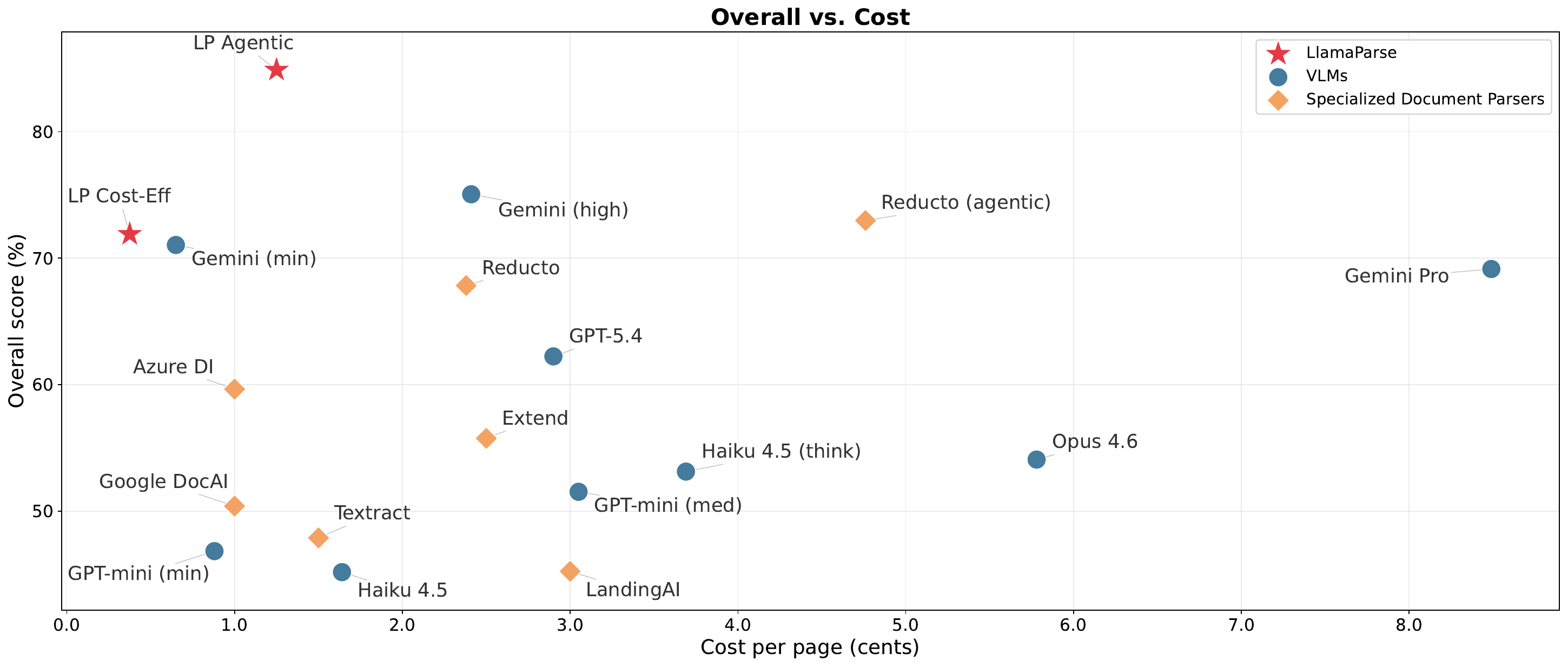}
\caption{Quality vs.\ cost for \qualitycostpricingproviders{} (\Cref{sec:cost}). Per-page costs use publicly listed pay-as-you-go prices~\qualitycostpricingcites.}
\label{fig:quality-cost}
\end{figure*}

\subsection{Quality--Cost Frontier}
\label{sec:cost}

\Cref{fig:quality-cost} plots overall score (mean of the five dimension scores) against average per-page cost (averaged across the four dataset categories) for all evaluated providers.
Open-weight providers are excluded, as their costs depend on hardware and deployment choices and are not directly comparable.
In addition to the configurations used in the main evaluation, we include higher-reasoning variants for comparison: GPT-5 Mini at its default medium reasoning, Gemini~3 Flash at its default high thinking, and Haiku~4.5 with thinking enabled. We also include three premium frontier VLMs at provider-default settings: Opus~4.6, Gemini~3.1 Pro, and GPT-5.4.

Two patterns stand out. First, increasing compute budget yields diminishing returns: Gemini~3 Flash gains ${\sim}$4 points moving from minimal to high thinking at ${\sim}$4$\times$ the cost, and upgrading to Gemini~3.1 Pro (8.5\,\textcent/page) does not help further, scoring below Flash~(minimal) on overall at 13$\times$ the cost; GPT-5.4 (2.9\,\textcent/page) and Opus~4.6 (5.8\,\textcent/page) at their default non-reasoning settings similarly fall short. GPT-5 Mini and Haiku~4.5 follow the same pattern, gaining ${\sim}$5 and ${\sim}$8 points respectively when enabling reasoning at 2--4$\times$ the cost. Among specialized parsers, \reductoname{}'s agentic mode (${\sim}$5\,\textcent/page) yields a ${\sim}$5 point overall improvement, with gains in both tables (+10 points) and charts (+16 points). This supports our choice to standardize on minimal reasoning. Second, LlamaParse Agentic sits on the Pareto frontier (${\sim}$1.2\,\textcent/page, \agenticoverall\% overall), outperforming all providers, including premium frontier VLMs, at any cost level. LlamaParse Cost Effective offers the lowest cost ($<$0.4\,\textcent/page) while remaining competitive with Gemini~(minimal) at \costeffectiveoverall\%.

Per-dimension quality-cost plots are provided in \Cref{app:per-dim-cost}.

\section{Related Work}
\label{sec:related-work}

\subsection{Benchmarks for Document Parsing}

\paragraph{Subtask benchmarks.}
Early benchmarks each target a single parsing capability on a narrow document domain (\Cref{tab:benchmark-comparison}).
PubTabNet~\cite{zhong2020pubtabnet} and FinTabNet~\cite{zheng2020gte} evaluate table structure recognition on academic and financial documents;
DocLayNet~\cite{pfitzmann2022doclaynet} evaluates layout detection but not content extraction;
ChartQA~\cite{masry2022chartqa} tests chart reasoning through question answering rather than structured data-point extraction.

\paragraph{End-to-end benchmarks.}
Recent benchmarks broaden scope to multi-task evaluation but leave significant gaps in document coverage and capability breadth.
OmniDocBench~\cite{ouyang2024omnidocbench} jointly evaluates layout, tables, formulas, and text across nine document types, but only 6\% of its pages are enterprise-related, and it does not evaluate charts, formatting, or visual grounding.
OCRBench~v2~\cite{fu2025ocrbenchv2} scales to 10K human-verified samples across 23 tasks and 31 scenarios, providing broad coverage of LMM OCR capabilities including element parsing evaluated with TEDS; however, it draws from academic datasets rather than enterprise documents, relies on text-similarity metrics for structured elements, and does not assess semantic formatting or visual grounding.
olmOCR-Bench~\cite{poznanski2025olmocr} introduces rule-based binary tests---a meaningful step over fuzzy matching---but skews toward academic content, uses coarse cell-adjacency checks for tables, and strips formatting during evaluation.

\paragraph{Gaps addressed by ParseBench.}
These benchmarks collectively leave three gaps: enterprise documents remain underrepresented in multi-capability benchmarks; no existing benchmark evaluates chart data-point extraction or semantic formatting; and visual grounding with content attribution is not evaluated.
ParseBench addresses these gaps with enterprise-focused documents, five jointly evaluated capability dimensions, and metrics designed for semantic correctness.

\subsection{Models for Document Parsing}

\paragraph{Vision-language models.}
General-purpose VLMs such as GPT~\cite{openai2025gpt5}, Gemini~\cite{gemini3flash2025}, and Claude~\cite{anthropic2025haiku45} can extract structured content from document images in a single pass.
A separate class of OCR-specialized VLMs, including Qwen-VL~\cite{bai2025qwen3vl} and Dots OCR~\cite{dots_ocr}, are fine-tuned specifically for document transcription and offer competitive quality at lower cost.
Both categories generalize well across document types, languages, and layouts without task-specific engineering.
Visual grounding remains a weak point for most VLMs, though recent models such as Gemini~3 Flash show meaningful improvement.
Many VLMs offer configurable thinking budgets, but it is not yet clear whether increased compute reliably improves document transcription quality.

\paragraph{Specialized document parsers.}
Specialized document parsers combine layout detection, OCR, table recognition, and other task-specific modules in a pipeline.
Commercial platforms~\cite{azure_doc_intel_pricing, textract_pricing, google_docai_pricing} have been deployed at scale for field extraction and digitization, while open-source pipelines such as Docling~\cite{docling2024}, MinerU~\cite{wang2024mineru}, and PaddleOCR~\cite{cui2025paddleocr} provide configurable alternatives.
These systems excel at layout detection and spatial grounding but face challenges adapting to diverse document formats beyond their training distribution.
Most were built for digitization and field-extraction workflows rather than the open-ended understanding that agents require, and as a result often lack support for capabilities like chart data extraction and semantically meaningful formatting.

\section{Conclusion}
\label{sec:conclusion}

We introduced ParseBench, a benchmark of ${\sim}2{,}000$ human-verified enterprise document pages that evaluates parsing quality across five capability dimensions with semantic correctness metrics.
Our results show that document parsing remains a fragmented problem, with current systems exhibiting clear tradeoffs rather than consistent strength across all dimensions.
Within this landscape, LlamaParse Agentic delivers the strongest overall profile, combining robust performance across dimensions rather than excelling narrowly on a subset.
At the same time, larger models and more expensive inference do not reliably close the remaining gaps, indicating that future progress will depend on better system design and evaluation, not compute alone.

ParseBench is a first step toward benchmarking document processing for AI agents in enterprise automation. One of the central gaps this work highlights is not only model capability, but evaluation itself: existing benchmarks do not yet capture the full range of documents, tasks, and failure modes that matter in real agentic workflows. We view ParseBench as an initial foundation for closing that gap.

There are several clear directions for extending this benchmark. A first is greater scale and broader enterprise domain coverage, so that the benchmark captures more of the domain-specific parsing challenges that arise in practice. A second is broader task scope: extending evaluation beyond parsing quality to downstream tasks such as structured extraction and document classification-and-splitting workflows. A third is harder evaluation settings: ultra-high-resolution pages, visually dense technical documents, and more adversarial enterprise cases that remain challenging even for the strongest current systems. We hope ParseBench helps establish a path toward larger and more realistic benchmarks for agent-facing document understanding, where progress is measured not only by text recovery, but by reliable structure, semantics, and grounding.

\newpage
\bibliographystyle{plainnat}
\bibliography{references}

\begin{thebibliography}{32}
\providecommand{\natexlab}[1]{#1}
\providecommand{\url}[1]{\texttt{#1}}
\expandafter\ifx\csname urlstyle\endcsname\relax
  \providecommand{\doi}[1]{doi: #1}\else
  \providecommand{\doi}{doi: \begingroup \urlstyle{rm}\Url}\fi

\bibitem[{Amazon Web Services}(2026)]{textract_pricing}
{Amazon Web Services}.
\newblock Amazon {Textract} pricing.
\newblock \url{https://aws.amazon.com/textract/pricing/}, 2026.
\newblock Accessed: 2026-04-01.

\bibitem[{Anthropic}(2025{\natexlab{a}})]{anthropic2025haiku45}
{Anthropic}.
\newblock {Claude Haiku 4.5 System Card}.
\newblock Technical report, October 2025{\natexlab{a}}.
\newblock URL
  \url{https://www-cdn.anthropic.com/7aad69bf12627d42234e01ee7c36305dc2f6a970.pdf}.

\bibitem[{Anthropic}(2025{\natexlab{b}})]{anthropic2025opus45}
{Anthropic}.
\newblock {Claude Opus 4.5 System Card}.
\newblock Technical report, November 2025{\natexlab{b}}.
\newblock URL
  \url{https://www-cdn.anthropic.com/bf10f64990cfda0ba858290be7b8cc6317685f47.pdf}.

\bibitem[{Anthropic}(2026)]{anthropic_pricing}
{Anthropic}.
\newblock Claude pricing.
\newblock \url{https://platform.claude.com/docs/en/about-claude/pricing}, 2026.
\newblock Accessed: 2026-04-01.

\bibitem[Bai et~al.(2025)Bai, Cai, Chen, Chen, Chen, Cheng, Deng, Ding, Gao,
  Ge, Ge, Guo, Huang, Huang, Huang, Hui, Jiang, Li, Li, Li, Li, Lin, Lin, Liu,
  Liu, Liu, Liu, Liu, Liu, Lu, Luo, Lv, Men, Meng, Ren, Ren, Song, Sun, Tang,
  Tu, Wan, Wang, Wang, Wang, Wang, Xie, Xu, Xu, Xu, Yang, Yang, Yang, Yang, Yu,
  Zhang, Zhang, Zhang, Zheng, Zhong, Zhou, Zhou, Zhou, Zhu, and
  Zhu]{bai2025qwen3vl}
Shuai Bai, Yuxuan Cai, Ruizhe Chen, Keqin Chen, Xionghui Chen, Zesen Cheng,
  Lianghao Deng, Wei Ding, Chang Gao, Chunjiang Ge, Wenbin Ge, Zhifang Guo,
  Qidong Huang, Jie Huang, Fei Huang, Binyuan Hui, Shutong Jiang, Zhaohai Li,
  Mingsheng Li, Mei Li, Kaixin Li, Zicheng Lin, Junyang Lin, Xuejing Liu,
  Jiawei Liu, Chenglong Liu, Yang Liu, Dayiheng Liu, Shixuan Liu, Dunjie Lu,
  Ruilin Luo, Chenxu Lv, Rui Men, Lingchen Meng, Xuancheng Ren, Xingzhang Ren,
  Sibo Song, Yuchong Sun, Jun Tang, Jianhong Tu, Jianqiang Wan, Peng Wang,
  Pengfei Wang, Qiuyue Wang, Yuxuan Wang, Tianbao Xie, Yiheng Xu, Haiyang Xu,
  Jin Xu, Zhibo Yang, Mingkun Yang, Jianxin Yang, An~Yang, Bowen Yu, Fei Zhang,
  Hang Zhang, Xi~Zhang, Bo~Zheng, Humen Zhong, Jingren Zhou, Fan Zhou, Jing
  Zhou, Yuanzhi Zhu, and Ke~Zhu.
\newblock Qwen3-vl technical report, 2025.
\newblock URL \url{https://arxiv.org/abs/2511.21631}.

\bibitem[Cui et~al.(2025)Cui, Sun, Lin, Gao, Zhang, Liu, Wang, Zhang, Zhou,
  Liu, Zhang, Lv, Huang, Zhang, Zhang, Zhang, Liu, Yu, and
  Ma]{cui2025paddleocr}
Cheng Cui, Ting Sun, Manhui Lin, Tingquan Gao, Yubo Zhang, Jiaxuan Liu, Xueqing
  Wang, Zelun Zhang, Changda Zhou, Hongen Liu, Yue Zhang, Wenyu Lv, Kui Huang,
  Yichao Zhang, Jing Zhang, Jun Zhang, Yi~Liu, Dianhai Yu, and Yanjun Ma.
\newblock Paddleocr 3.0 technical report, 2025.
\newblock URL \url{https://arxiv.org/abs/2507.05595}.

\bibitem[{Extend AI}(2026)]{extend_pricing}
{Extend AI}.
\newblock How credits work -- {Extend AI}.
\newblock
  \url{https://docs.extend.ai/2026-02-09/product/general/how-credits-work},
  2026.
\newblock Accessed: 2026-04-01.

\bibitem[Fu et~al.(2025)Fu, Kuang, Song, Huang, Yang, Li, Zhu, Luo, Wang, Lu,
  Li, Tang, Shan, Lin, Liu, Wu, Feng, Liu, Huang, Tang, Chen, Jin, Liu, and
  Bai]{fu2025ocrbenchv2}
Ling Fu, Zhebin Kuang, Jiajun Song, Mingxin Huang, Biao Yang, Yuzhe Li, Linghao
  Zhu, Qidi Luo, Xinyu Wang, Hao Lu, Zhang Li, Guozhi Tang, Bin Shan, Chunhui
  Lin, Qi~Liu, Binghong Wu, Hao Feng, Hao Liu, Can Huang, Jingqun Tang, Wei
  Chen, Lianwen Jin, Yuliang Liu, and Xiang Bai.
\newblock {OCRBench} v2: An improved benchmark for evaluating large multimodal
  models on visual text localization and reasoning.
\newblock \emph{arXiv preprint arXiv:2501.00321}, 2025.

\bibitem[{Google}(2026)]{gemini_pricing}
{Google}.
\newblock Gemini {API} pricing.
\newblock \url{https://ai.google.dev/gemini-api/docs/pricing}, 2026.
\newblock Accessed: 2026-04-01.

\bibitem[{Google Cloud}(2026)]{google_docai_pricing}
{Google Cloud}.
\newblock Google cloud {Document AI} pricing.
\newblock \url{https://cloud.google.com/document-ai/pricing}, 2026.
\newblock Accessed: 2026-04-01.

\bibitem[{Google DeepMind}(2025)]{gemini3flash2025}
{Google DeepMind}.
\newblock Gemini 3 flash model card.
\newblock Technical report, Google DeepMind, December 2025.
\newblock URL
  \url{https://storage.googleapis.com/deepmind-media/Model-Cards/Gemini-3-Flash-Model-Card.pdf}.

\bibitem[{Granite Vision Team} et~al.(2025){Granite Vision Team}, Karlinsky,
  Arbelle, Daniels, Nassar, et~al.]{granitevision2025}
{Granite Vision Team}, Leonid Karlinsky, Assaf Arbelle, Abraham Daniels, Ahmed
  Nassar, et~al.
\newblock Granite vision: A lightweight, open-source multimodal model for
  enterprise intelligence, 2025.
\newblock URL \url{https://arxiv.org/abs/2502.09927}.

\bibitem[Kwon et~al.(2023)Kwon, Li, Zhuang, Sheng, Zheng, Yu, Gonzalez, Zhang,
  and Stoica]{kwon2023vllm}
Woosuk Kwon, Zhuohan Li, Siyuan Zhuang, Ying Sheng, Lianmin Zheng, Cody~Hao Yu,
  Joseph Gonzalez, Hao Zhang, and Ion Stoica.
\newblock Efficient memory management for large language model serving with
  pagedattention.
\newblock In \emph{Proceedings of the 29th Symposium on Operating Systems
  Principles (SOSP)}, pages 611--626. ACM, 2023.
\newblock \doi{10.1145/3600006.3613165}.

\bibitem[{LandingAI}(2026)]{landingai_pricing}
{LandingAI}.
\newblock Landingai {Document Extraction} pricing.
\newblock \url{https://docs.landing.ai/ade/ade-pricing}, 2026.
\newblock Accessed: 2026-04-01.

\bibitem[Li et~al.(2025)Li, Yang, Liu, Wang, and Zhang]{dots_ocr}
Yumeng Li, Guang Yang, Hao Liu, Bowen Wang, and Colin Zhang.
\newblock dots.ocr: Multilingual document layout parsing in a single
  vision-language model, 2025.
\newblock URL \url{https://arxiv.org/abs/2512.02498}.

\bibitem[Lin et~al.(2014)Lin, Maire, Belongie, Hays, Perona, Ramanan,
  Doll{\'a}r, and Zitnick]{lin2014coco}
Tsung-Yi Lin, Michael Maire, Serge Belongie, James Hays, Pietro Perona, Deva
  Ramanan, Piotr Doll{\'a}r, and C.~Lawrence Zitnick.
\newblock Microsoft {COCO}: Common objects in context.
\newblock In \emph{European Conference on Computer Vision (ECCV)}, pages
  740--755. Springer, 2014.

\bibitem[Livathinos et~al.(2025)Livathinos, Auer, Lysak, Nassar, Dolfi,
  Vagenas, Ramis, Omenetti, Dinkla, Kim, Gupta, de~Lima, Weber, Morin, Meijer,
  Kuropiatnyk, and Staar]{docling2024}
Nikolaos Livathinos, Christoph Auer, Maksym Lysak, Ahmed Nassar, Michele Dolfi,
  Panos Vagenas, Cesar~Berrospi Ramis, Matteo Omenetti, Kasper Dinkla, Yusik
  Kim, Shubham Gupta, Rafael~Teixeira de~Lima, Valery Weber, Lucas Morin,
  Ingmar Meijer, Viktor Kuropiatnyk, and Peter W.~J. Staar.
\newblock Docling: An efficient open-source toolkit for ai-driven document
  conversion, 2025.
\newblock URL \url{https://arxiv.org/abs/2501.17887}.

\bibitem[{LlamaIndex}(2026)]{llamaparse_pricing}
{LlamaIndex}.
\newblock Llamacloud pricing.
\newblock \url{https://developers.llamaindex.ai/python/cloud/general/pricing/},
  2026.
\newblock Accessed: 2026-04-01.

\bibitem[Masry et~al.(2022)Masry, Long, Tan, Joty, and Hoque]{masry2022chartqa}
Ahmed Masry, Do~Xuan Long, Jia~Qing Tan, Shafiq Joty, and Enamul Hoque.
\newblock {ChartQA}: A benchmark for question answering about charts with
  visual and logical reasoning.
\newblock In \emph{Findings of the Association for Computational Linguistics:
  ACL 2022}, pages 2263--2279. Association for Computational Linguistics, 2022.

\bibitem[{Microsoft}(2026)]{azure_doc_intel_pricing}
{Microsoft}.
\newblock Azure {AI} {Document Intelligence} pricing.
\newblock
  \url{https://azure.microsoft.com/en-us/pricing/details/document-intelligence/},
  2026.
\newblock Accessed: 2026-04-01.

\bibitem[{Modal Labs}(2026)]{modal}
{Modal Labs}.
\newblock Modal: {AI} infrastructure that developers love.
\newblock \url{https://modal.com}, 2026.
\newblock Accessed: 2026-04-01.

\bibitem[{OpenAI}(2025)]{openai2025gpt5}
{OpenAI}.
\newblock {OpenAI GPT-5 System Card}.
\newblock \emph{arXiv preprint arXiv:2601.03267}, 2025.

\bibitem[{OpenAI}(2026)]{openai_pricing}
{OpenAI}.
\newblock {OpenAI} {API} pricing.
\newblock \url{https://developers.openai.com/api/docs/pricing}, 2026.
\newblock Accessed: 2026-04-01.

\bibitem[Ouyang et~al.(2025)Ouyang, Qu, Zhou, Zhu, Zhang, Lin, Wang, Zhao,
  Jiang, Zhao, Shi, Wu, Chu, Liu, Li, Xu, Zhang, Shi, Tu, and
  He]{ouyang2024omnidocbench}
Linke Ouyang, Yuan Qu, Hongbin Zhou, Jiawei Zhu, Rui Zhang, Qunshu Lin, Bin
  Wang, Zhiyuan Zhao, Man Jiang, Xiaomeng Zhao, Jin Shi, Fan Wu, Pei Chu,
  Minghao Liu, Zhenxiang Li, Chao Xu, Bo~Zhang, Botian Shi, Zhongying Tu, and
  Conghui He.
\newblock {OmniDocBench}: Benchmarking diverse {PDF} document parsing with
  comprehensive annotations.
\newblock In \emph{Proceedings of the IEEE/CVF Conference on Computer Vision
  and Pattern Recognition (CVPR)}, pages 24838--24848, 2025.

\bibitem[Pfitzmann et~al.(2022)Pfitzmann, Auer, Dolfi, Nassar, and
  Staar]{pfitzmann2022doclaynet}
Birgit Pfitzmann, Christoph Auer, Michele Dolfi, Ahmed~S. Nassar, and Peter
  Staar.
\newblock Doclaynet: A large human-annotated dataset for document-layout
  segmentation.
\newblock In \emph{Proceedings of the 28th ACM SIGKDD Conference on Knowledge
  Discovery and Data Mining}, pages 3743--3751. ACM, August 2022.
\newblock \doi{10.1145/3534678.3539043}.
\newblock URL \url{http://dx.doi.org/10.1145/3534678.3539043}.

\bibitem[Poznanski et~al.(2025)Poznanski, Rangapur, Borchardt, Dunkelberger,
  Huff, Lin, Rangapur, Wilhelm, Lo, and Soldaini]{poznanski2025olmocr}
Jake Poznanski, Aman Rangapur, Jon Borchardt, Jason Dunkelberger, Regan Huff,
  Daniel Lin, Aman Rangapur, Christopher Wilhelm, Kyle Lo, and Luca Soldaini.
\newblock olmocr: Unlocking trillions of tokens in pdfs with vision language
  models, 2025.
\newblock URL \url{https://arxiv.org/abs/2502.18443}.

\bibitem[{Reducto}(2026)]{reducto_pricing}
{Reducto}.
\newblock Credit usage -- {Reducto}.
\newblock \url{https://docs.reducto.ai/reference/credit-usage}, 2026.
\newblock Accessed: 2026-04-01.

\bibitem[Smock et~al.(2023)Smock, Pesala, and
  Abraham]{smock2023gritsgridtablesimilarity}
Brandon Smock, Rohith Pesala, and Robin Abraham.
\newblock Grits: Grid table similarity metric for table structure recognition,
  2023.
\newblock URL \url{https://arxiv.org/abs/2203.12555}.

\bibitem[Wang et~al.(2024)Wang, Xu, Zhao, Ouyang, Wu, Zhao, Xu, Liu, Qu, Shang,
  Zhang, Wei, Sui, Li, Shi, Qiao, Lin, and He]{wang2024mineru}
Bin Wang, Chao Xu, Xiaomeng Zhao, Linke Ouyang, Fan Wu, Zhiyuan Zhao, Rui Xu,
  Kaiwen Liu, Yuan Qu, Fukai Shang, Bo~Zhang, Liqun Wei, Zhihao Sui, Wei Li,
  Botian Shi, Yu~Qiao, Dahua Lin, and Conghui He.
\newblock Mineru: An open-source solution for precise document content
  extraction, 2024.
\newblock URL \url{https://arxiv.org/abs/2409.18839}.

\bibitem[Zheng et~al.(2021)Zheng, Burdick, Popa, Zhong, and Wang]{zheng2020gte}
Xinyi Zheng, Douglas Burdick, Lucian Popa, Xu~Zhong, and Nancy Xin~Ru Wang.
\newblock Global table extractor ({GTE}): A framework for joint table
  identification and cell structure recognition using visual context.
\newblock In \emph{Proceedings of the IEEE/CVF Winter Conference on
  Applications of Computer Vision (WACV)}, pages 697--706, 2021.

\bibitem[Zhong et~al.(2020{\natexlab{a}})Zhong, ShafieiBavani, and
  Jimeno~Yepes]{zhong2020imagebasedtablerecognitiondata}
Xu~Zhong, Elaheh ShafieiBavani, and Antonio Jimeno~Yepes.
\newblock Image-based table recognition: Data, model, and evaluation.
\newblock In \emph{European Conference on Computer Vision (ECCV)}, pages
  564--580. Springer, 2020{\natexlab{a}}.

\bibitem[Zhong et~al.(2020{\natexlab{b}})Zhong, ShafieiBavani, and
  Jimeno~Yepes]{zhong2020pubtabnet}
Xu~Zhong, Elaheh ShafieiBavani, and Antonio Jimeno~Yepes.
\newblock Image-based table recognition: Data, model, and evaluation.
\newblock In \emph{European Conference on Computer Vision (ECCV)}, pages
  564--580. Springer, 2020{\natexlab{b}}.

\end{thebibliography}

\newpage

\onecolumn
\appendix

\etocsetnexttocdepth{subsection}
\etocsettocstyle{\section*{Appendix Contents}\small}{}
\etocsetlocaltop.toc{part}
\localtableofcontents

\section{Benchmark Construction Details}
\label{app:benchmark}
\subsection{Annotation Pipeline}
\label{sec:appendix-annotation}
This section details the two-pass annotation pipeline used across all benchmark dimensions.

\paragraph{Pass~1: Auto-labeling with frontier VLMs.}
Frontier VLMs generate initial pseudo-ground-truth from source PDF pages:
\begin{itemize}
    \item \textbf{Tables:} Claude Opus~\cite{anthropic2025opus45} generates full HTML table structure.
    \item \textbf{Charts:} Gemini Flash extracts data points, each a numerical value with associated labels and a tolerance (default 1\%).
    \item \textbf{Content faithfulness \& semantic formatting:} Frontier VLMs transcribe each page to Markdown; a second VLM pass reviews the transcription against the source PDF and produces an analysis report flagging suspected errors in text, formatting, and reading order.
    \item \textbf{Visual grounding:} Frontier VLMs generate layout element annotations (bounding boxes, class labels, content associations, reading order).
\end{itemize}

\paragraph{Pass~2: Human verification \& correction.}
The review workflow is tailored to how ``editable'' each ground-truth format is:
\begin{itemize}
    \item \textbf{Tables:} HTML structure is too complex for annotators to manually edit reliably. Instead, annotators flag cells and write fix suggestions in natural language; an LLM then applies targeted fixes, preserving overall structure. The annotator then confirms the changes are correct, or the process loops until convergence.
    \item \textbf{Charts:} Data points are simple (value + labels), so annotators verify each data point directly and fix incorrect values or labels in place.
    \item \textbf{Content faithfulness \& semantic formatting:} Annotators review Markdown transcriptions guided by AI-flagged issues from Pass~1, correcting missed text, hallucinations, formatting errors, and reading-order problems directly.
    \item \textbf{Visual grounding:} Annotators verify all element annotations through a human-in-the-loop review process.
\end{itemize}

\paragraph{Content faithfulness \& semantic formatting: iterative pipeline.}
The annotation process for Markdown ground truth proceeds as follows:
\begin{enumerate}
    \item A VLM (Gemini~3.0 Flash) transcribes the source document into Markdown.
    \item A second VLM pass analyzes the Markdown against the original document, generating a quality report with suggested patches for any errors or omissions.
    \item A human annotator manually reviews the Markdown and approves or rejects patches in a custom-built annotation interface.
    \item If modifications were made in step~3, we return to step~2 and re-generate patches for human review. The loop terminates when the annotator accepts the transcription without further edits.
\end{enumerate}

This VLM-in-the-loop approach substantially reduces annotation time compared to fully manual transcription, while the human review step ensures ground-truth quality.

\subsection{Chart Dataset Composition}
\label{sec:appendix-chart-dimensions}
\begin{figure*}[h]
    \centering
    \includegraphics[width=0.95\linewidth]{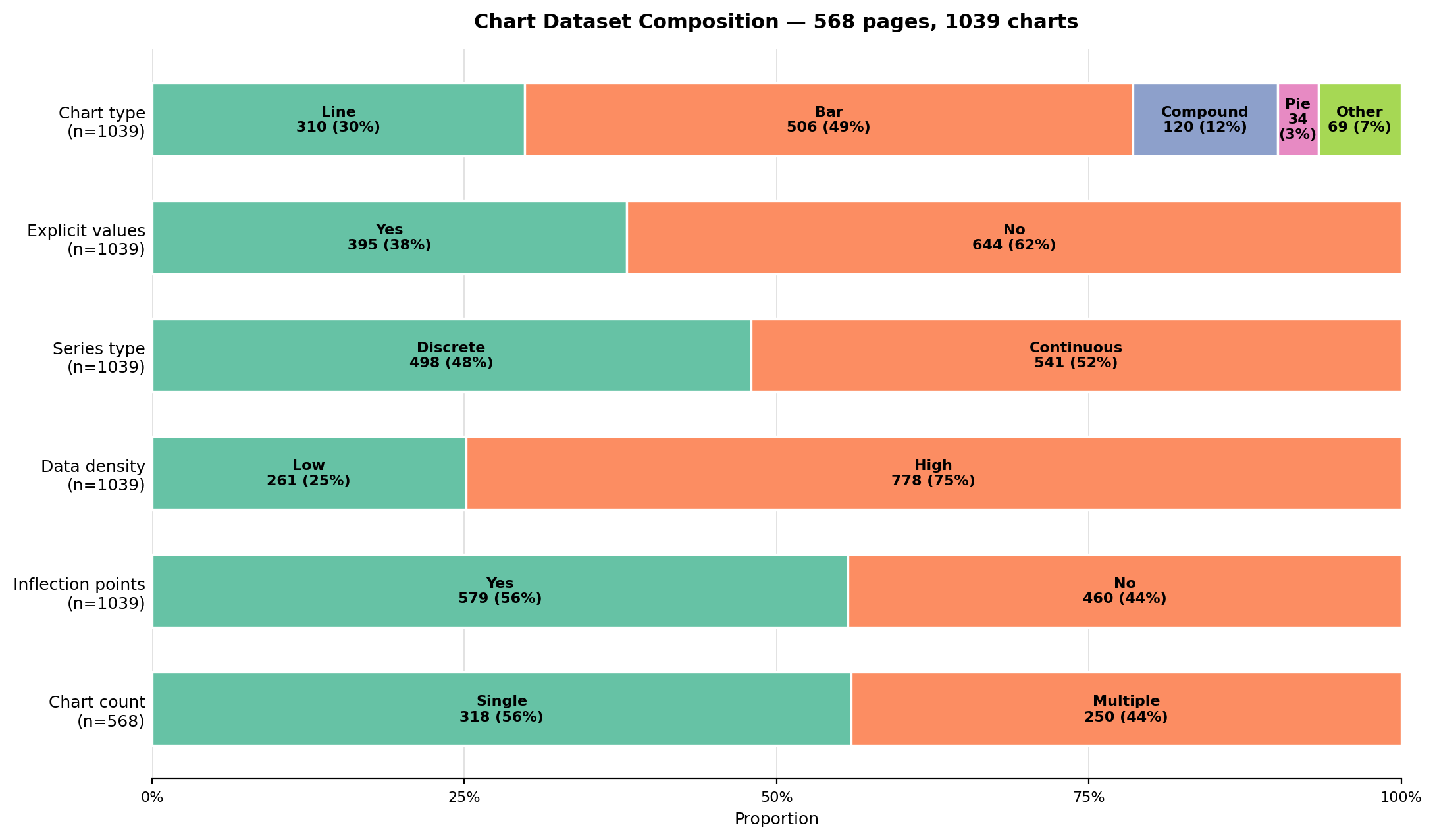}
    \caption{Distribution of chart characteristics in the benchmark dataset. Per-chart dimensions are computed over all $1{,}039$ charts; the \emph{chart count} dimension is computed per page.}
    \label{fig:chart-dimensions}
\end{figure*}

To validate the diversity of the chart benchmark, we classified every page along five per-chart dimensions (chart type, presence of explicit value labels, series type, data density, and inflection points) and one per-page dimension (number of distinct charts).
Here, \emph{data density} is a coarse low/high label reflecting how many plotted values a chart presents: low-density charts have only a few marks (e.g., 2--4 bars or a pie with 3 slices), whereas high-density charts contain many plotted values (e.g., 8+ bars or multiple lines over many time periods).
\Cref{fig:chart-dimensions} reports the resulting distribution across all $568$ pages and $1{,}039$ individual charts.

\section{Metric Details}
\label{app:metrics}
\subsection{Visual Grounding Auxiliary Metrics}
\label{app:grounding-aux-metrics}

In addition to the headline \textbf{Element Pass Rate}, we report auxiliary attribution metrics that isolate the attribution component of the joint grounding criterion. They are diagnostic summaries rather than separate terms averaged into the main score. All of these metrics use normalized tokenized text and the same attribution overlap threshold, $\mathrm{IoA}(\mathrm{GT},\mathrm{Pred}) \ge 0.30$.

\paragraph{Element-level attribution policies.}
\begin{itemize}
    \item \textbf{Standard attribution.} For standard content-bearing elements, we select the best overlapping prediction and mark attribution as passing when token-level $\mathrm{F1} \ge 0.80$.
    \item \textbf{Page furniture handling.} For \texttt{Page-Header} and \texttt{Page-Footer}, we use a grouped furniture-band policy for localization and attribution, so split/merge differences between a wide ground-truth band and multiple predicted fragments are handled more robustly. Classification remains strict on the matched representative prediction.
    \item \textbf{Explicit attribution.} In the current dataset, \texttt{explicit} is used on chart-like \texttt{Picture} elements. These regions contain some visible values or labels that should be recovered, but the ground truth is not intended to enumerate every valid description of the chart region. We therefore use recall-only attribution with the same $0.80$ threshold. Extra descriptive or inferred content is ignored as long as the expected explicit content is recovered.
    \item \textbf{Caption-attribute handling.} In the current dataset, the \texttt{caption} attribute is used primarily on picture- and chart-like regions. It marks associated text as descriptive text for that region rather than as a literal target string, so attribution is not graded. This attribute is distinct from the layout label \texttt{Caption}. The element still contributes to localization and classification.
    \item \textbf{Formula handling.} Attribution is skipped for formulas because equivalent mathematical content can be serialized into many valid LaTeX strings, making literal token-level grading unreliable.
    \item \textbf{Ignored elements.} In the current dataset, \texttt{ignore} is used for auxiliary non-gradeable regions, mostly chart-related descriptions and a small number of text fragments. When \texttt{ignore=true}, the entire layout element is excluded from evaluation.
    \item \textbf{Other skipped attribution.} Attribution is also skipped for elements without a human-verified content association whose expected literal text is well-defined enough to grade directly.
    \item \textbf{Merge-aware filtering.} When scoring element-level attribution, tokens attributable to nearby overlapping ground-truth elements are filtered out so merged predictions are not over-credited.
\end{itemize}

Let $T(\cdot)$ denote the multiset of normalized tokens associated with an element or predicted block.

\paragraph{LAP (Local Attribution Precision).}
For each predicted block $p$, we gather all overlapping ground-truth elements and take the union of their tokens, denoted $G(p)$. LAP measures the fraction of predicted tokens that can be attributed to some overlapping ground-truth content:
$$
\mathrm{LAP}(p)=\frac{|T(p)\cap T(G(p))|}{|T(p)|}.
$$
The reported score is a token-weighted micro-average over predicted blocks.

\paragraph{LAR (Local Attribution Recall).}
For each ground-truth element $g$, we gather all overlapping predicted blocks and take the union of their tokens, denoted $P(g)$. LAR measures the fraction of ground-truth content recovered by the prediction:
$$
\mathrm{LAR}(g)=\frac{|T(g)\cap T(P(g))|}{|T(g)|}.
$$
The reported score is a token-weighted micro-average over ground-truth elements.

\paragraph{AF1.}
AF1 is the harmonic mean of the global LAP and LAR scores:
$$
\mathrm{AF1}=\frac{2\,\mathrm{LAP}\,\mathrm{LAR}}{\mathrm{LAP}+\mathrm{LAR}}.
$$

\paragraph{Eligibility in the global attribution diagnostics.}
\begin{itemize}
    \item In the standalone attribution diagnostics, formulas and elements with the \texttt{caption} attribute are excluded from grading, and elements marked \texttt{ignore} are removed entirely.
    \item Elements marked \texttt{explicit} remain in recall-oriented metrics, but they are excluded from LAP, and predictions overlapping only explicit elements are removed from the LAP denominator. This avoids penalizing additional descriptive or inferred content when the benchmark only requires recovery of the explicit target content.
\end{itemize}

\subsection{Semantic Formatting: Full Styling Classes}
\label{app:styling-classes}

The main text reports results on the four formatting classes with the highest semantic impact: strikethrough, superscript, subscript, and bold.
Here we describe the full set of styling classes evaluated in our framework.
Each class has both positive and negative test rules: a positive rule (e.g., \texttt{is\_strikeout}) checks that expected formatting is present, while its negative counterpart (e.g., \texttt{is\_not\_strikeout}) checks that formatting is not falsely applied.
Rules check for the appropriate Markdown markers (\texttt{**...**} for bold, \texttt{\textasciitilde\textasciitilde...\textasciitilde\textasciitilde} for strikethrough, \texttt{<sup>...</sup>} for superscript, etc.).

\begin{itemize}
    \item \textbf{Strikethrough} (\texttt{is\_strikeout} / \texttt{is\_not\_strikeout}): Detects text marked as deleted or superseded.
    \item \textbf{Superscript} (\texttt{is\_sup} / \texttt{is\_not\_sup}): Detects superscript notation used in footnote references, exponents, and ordinals.
    \item \textbf{Subscript} (\texttt{is\_sub} / \texttt{is\_not\_sub}): Detects subscript notation used in chemical formulae and mathematical indices.
    \item \textbf{Bold} (\texttt{is\_bold} / \texttt{is\_not\_bold}): Detects bold text used for defined terms, section labels, or key values.
    \item \textbf{Italic} (\texttt{is\_italic} / \texttt{is\_not\_italic}): Detects italic text used for emphasis, titles, or foreign terms.
    \item \textbf{Underline} (\texttt{is\_underline} / \texttt{is\_not\_underline}): Detects underlined text.
    \item \textbf{Highlight} (\texttt{is\_highlight} / \texttt{is\_not\_highlight}): Detects highlighted or background-colored text.
\end{itemize}

Only strikethrough, superscript, subscript, and bold contribute to the headline semantic formatting score; the remaining classes are reported for completeness and may be promoted in future benchmark revisions.

\section{Evaluation Protocol}
\label{app:protocol}
\subsection{Provider Configurations}
\label{app:provider-configs}

\Cref{tab:provider-configs} details the exact configuration used for each evaluated method.

\begin{table*}[h]
\centering
\small
\begin{tabular*}{\textwidth}{@{\extracolsep{\fill}}lllp{5cm}@{}}
\toprule
\textbf{Provider} & \textbf{Category} & \textbf{Model / Mode} & \textbf{Notes} \\
\midrule
OpenAI GPT-5 Mini      & VLM (commercial)        & gpt-5-mini        & Reasoning: minimal (default exceeds cost target) \\
Anthropic Haiku 4.5    & VLM (commercial)        & claude-haiku-4-5   & Thinking: disabled (default) \\
Google Gemini 3 Flash  & VLM (commercial)        & gemini-3-flash     & Thinking: minimal (default exceeds cost target) \\
Qwen 3 VL              & VLM (open) & Qwen3-VL-8B-Instruct & H100, vLLM 0.11.2, BF16 \\
Dots OCR 1.5           & VLM (open) & dots-ocr-1.5       & H100, vLLM 0.16.0, BF16 \\
\midrule
Docling                & OSS Pipeline        & Default pipeline   & v2.73.1, H100 \\
AWS Textract           & Commercial API        & AnalyzeDocument     & TABLES feature; FORMS disabled \\
Google Cloud Doc AI    & Commercial API        & Layout parser       & Layout Parser: semantic labels, no bboxes. OCR: bboxes, no semantic labels. \\
Azure Doc Intelligence & Commercial API        & prebuilt-layout     & Markdown output \\
\reductoname                & Commercial API        & Agentic API         & Agentic=True (text scope), HTML tables \\
\extendname                 & Commercial API        & Default API         & Markdown target; HTML tables \\
\landingainame              & Commercial API        & dpt-2-latest        & Default model \\
\midrule
LlamaParse Cost Eff.   & Ours       & Cost Effective      & Optimized for throughput \\
LlamaParse Agentic     & Ours       & Agentic             & Multi-step pipeline \\
\bottomrule
\end{tabular*}
\caption{Per-provider evaluation configurations. All VLMs use the same standardized prompt per benchmark dimension.}
\label{tab:provider-configs}
\end{table*}

\subsection{VLM Evaluation Prompts}
\label{app:vlm-prompts}

We use three prompt configurations across the evaluated VLMs. All produce both parsed content and layout annotations (bounding boxes, class labels), enabling evaluation across all five benchmark dimensions. Each VLM receives one document page image per request.

\paragraph{Proprietary VLMs (OpenAI, Anthropic, Gemini).}
All proprietary VLMs receive the same combined prompt that requests markdown content with inline layout annotations. GPT-5 Mini, GPT-5.4, Haiku~4.5, Opus~4.6 share one prompt; Gemini~3 Flash and Gemini~3.1 Pro use the same prompt structure with its \texttt{data-bbox} coordinate order adapted to Gemini's native format \texttt{[y\_min, x\_min, y\_max, x\_max]} rather than \texttt{[x1, y1, x2, y2]}. The output uses HTML \texttt{<div>} wrappers with \texttt{data-bbox} and \texttt{data-label} attributes; Gemini outputs are converted back to \texttt{[x1, y1, x2, y2]} in post-processing before evaluation.

\subparagraph{Shared system prompt (GPT-5 Mini, Haiku~4.5, Opus~4.6, GPT-5.4).}
\begin{quote}
\small\ttfamily
You are a document parser. Your task is to convert document images to clean, well-structured markdown.\\[4pt]
Guidelines:\\
- Preserve the document structure (headings, paragraphs, lists, tables)\\
- Convert tables to HTML format (<table>, <tr>, <th>, <td>)\\
- For existing tables in the document: use colspan and rowspan attributes to preserve merged cells and hierarchical headers\\
- For charts/graphs being converted to tables: use flat combined column headers (e.g., "Primary 2015" not separate rows) so each data cell's row contains all its labels\\
- Describe images/figures briefly in square brackets like [Figure: description]\\
- Preserve any code blocks with appropriate syntax highlighting\\
- Maintain reading order (left-to-right, top-to-bottom for Western documents)\\
- Do not add commentary or explanations - only output the parsed content\\[4pt]
Additionally, wrap each layout element in a <div> tag with:\\
- data-bbox="[x1, y1, x2, y2]" -- bounding box in normalized 0-1000 coordinates where x is horizontal (left edge = 0, right edge = 1000) and y is vertical (top = 0, bottom = 1000). x1,y1 is the top-left corner and x2,y2 is the bottom-right corner.\\
- data-label="<category>" -- one of: Caption, Footnote, Formula, List-item, Page-footer, Page-header, Picture, Section-header, Table, Text, Title\\[4pt]
Place elements in reading order. Every piece of content must be inside exactly one <div> wrapper.
\end{quote}

\subparagraph{Shared user prompt (GPT-5 Mini, GPT-5.4, Haiku~4.5, Opus~4.6).}
\begin{quote}
\small\ttfamily
Parse this document page and output its content as clean markdown, with each layout element wrapped in a <div data-bbox="[x1,y1,x2,y2]" data-label="Category"> tag. Use HTML tables for any tabular data. For charts/graphs, use flat combined column headers. Output ONLY the parsed content with div wrappers, no explanations.
\end{quote}

\subparagraph{Gemini bbox adaptation (Gemini~3 Flash, Gemini~3.1 Pro).}
Gemini models use the same prompts with only the \texttt{data-bbox} format changed to their native coordinate order.

\begin{quote}
\small\ttfamily
Additionally, wrap each layout element in a <div> tag with:\\
- data-bbox="[y\_min, x\_min, y\_max, x\_max]" -- bounding box in normalized 0-1000 coordinates where x is horizontal (left edge = 0, right edge = 1000) and y is vertical (top = 0, bottom = 1000). The order is [y\_min, x\_min, y\_max, x\_max].\\
\end{quote}

\begin{quote}
\small\ttfamily
Parse this document page and output its content as clean markdown, with each layout element wrapped in a <div data-bbox="[y\_min,x\_min,y\_max,x\_max]" data-label="Category"> tag. Use HTML tables for any tabular data. For charts/graphs, use flat combined column headers. Output ONLY the parsed content with div wrappers, no explanations.
\end{quote}

\paragraph{Dots~OCR~1.5.}
We use the recommended prompt from the official Dots~OCR documentation~\cite{dots_ocr}, which requests structured JSON output with bounding boxes, categories, and text content in a single call.

\begin{quote}
\small\ttfamily
Please output the layout information from the PDF image, including each layout element's bbox, its category, and the corresponding text content within the bbox.\\[4pt]
1. Bbox format: [x1, y1, x2, y2]\\
2. Layout Categories: The possible categories are ['Caption', 'Footnote', 'Formula', 'List-item', 'Page-footer', 'Page-header', 'Picture', 'Section-header', 'Table', 'Text', 'Title'].\\
3. Text Extraction \& Formatting Rules:\\
\quad - Picture: If the picture is a chart or graph, extract all data points and format as an HTML table with flat combined column headers. For non-chart pictures, the text field should be omitted.\\
\quad - Formula: Format its text as LaTeX.\\
\quad - Table: Format its text as HTML.\\
\quad - All Others: Format their text as Markdown.\\
4. Constraints: The output text must be the original text from the image, with no translation. All layout elements must be sorted according to human reading order.\\
5. Final Output: The entire output must be a single JSON object.
\end{quote}

\paragraph{Qwen~3~VL (8B).}
As a general-purpose VLM not specifically trained for document parsing or layout detection, Qwen~3~VL performs significantly worse when asked to produce both parsed content and layout annotations in a single prompt. We therefore use two separate pipelines: one for content parsing (markdown output) and one for layout detection (structured JSON with bounding boxes).

\subparagraph{Parse prompt.}
\begin{quote}
\small\ttfamily
Parse this document image and output its content as clean markdown.\\
- Preserve document structure (headings, paragraphs, lists, tables)\\
- Convert tables to HTML format (<table>, <tr>, <th>, <td>) with colspan/rowspan for merged cells\\
- Format formulas as LaTeX\\
- Describe images/figures briefly in square brackets like [Figure: description]\\
- Maintain reading order\\
- Output the original text with no translation\\
- Do not add commentary - only output the parsed content
\end{quote}

\subparagraph{Layout prompt.}
\begin{quote}
\small\ttfamily
Please output the layout information from the PDF image, including each layout element's bbox, its category, and the corresponding text content within the bbox.\\[4pt]
1. Bbox format: [x1, y1, x2, y2] using normalized 0-1000 coordinates.\\
2. Layout Categories: ['Caption', 'Footnote', 'Formula', 'List-item', 'Page-footer', 'Page-header', 'Picture', 'Section-header', 'Table', 'Text', 'Title'].\\
3. Text Extraction \& Formatting Rules: Picture charts as HTML tables, Formula as LaTeX, Table as HTML, all others as Markdown.\\
4. Constraints: Original text only, no translation. Reading order.\\
5. Final Output: Return ONLY a JSON array.
\end{quote}

\subsection{Infrastructure Details}
\label{app:infra}

Open-weight VLMs and the open-source Docling pipeline are deployed on Modal's~\cite{modal} serverless GPU infrastructure. \Cref{tab:infra} summarizes the hardware and software stack for each self-hosted model.

\begin{table*}[h]
\centering
\small
\begin{tabular}{lllll}
\toprule
\textbf{Model} & \textbf{GPU} & \textbf{Serving Framework} & \textbf{Version} & \textbf{Notes} \\
\midrule
Qwen 3 VL (8B)  & NVIDIA H100 & vLLM & 0.11.2 & Full precision (BF16) \\
Dots OCR 1.5    & NVIDIA H100 & vLLM & 0.16.0 & Full precision (BF16) \\
Docling         & NVIDIA H100 & Native pipeline & v2.73.1 & Chart extraction via Granite Vision 3.3 2B~\cite{granitevision2025} \\
\bottomrule
\end{tabular}
\caption{Hardware and software configurations for self-hosted models. All deployments use Modal serverless GPU containers with on-demand scaling.}
\label{tab:infra}
\end{table*}

Commercial APIs (AWS Textract, Azure Document Intelligence, Google Cloud Document AI, \reductoname, \extendname, \landingainame) are called through their official REST APIs with default or recommended configurations for document parsing.

\section{Additional Results}
\label{app:results}
\subsection{Cost Analysis}
\label{app:cost-analysis}

\Cref{tab:vlm-costs} reports per-page costs across dataset categories for each VLM configuration.

For the main comparison, we configure proprietary VLMs near a common low-cost operating point of roughly one cent per page when possible. All configurations include layout element extraction. At default reasoning settings, Gemini~3 Flash (thinking: high) averages 2.41\textcent/page, GPT-5 Mini (reasoning: medium) averages 3.05\textcent/page, and Anthropic Haiku~4.5 (thinking) averages 3.69\textcent/page, all exceeding this target. Reducing reasoning to minimal brings Gemini~3 Flash to 0.65\textcent/page and GPT-5 Mini to 0.88\textcent/page. Disabling thinking on Anthropic Haiku~4.5 brings it to 1.64\textcent/page.

We additionally report costs for three premium frontier VLMs at provider-default settings: Opus~4.6 (5.78\textcent/page, no thinking), Gemini~3.1 Pro (8.49\textcent/page, default high thinking), and GPT-5.4 (2.90\textcent/page, no reasoning). These represent a 3--13$\times$ cost increase over the minimal-reasoning configurations of their smaller counterparts.

\begin{table*}[h]
\centering
\small
\begin{tabular}{llccccc}
\toprule
\textbf{Provider} & \textbf{Reasoning} & \textbf{Charts} & \textbf{Tables} & \textbf{Text} & \textbf{Layout} & \textbf{Average} \\
 & & \multicolumn{5}{c}{\textit{(\textcent/page)}} \\
\midrule
\multirow{2}{*}{Anthropic Haiku 4.5}
  & disabled$^\dagger$ & 0.97 & 2.46 & 1.53 & 1.60 & 1.64 \\
  & thinking & 3.49 & 4.26 & 3.16 & 3.83 & 3.69 \\
\midrule
\multirow{2}{*}{Gemini 3 Flash}
  & minimal         & 0.49 & 0.84 & 0.56 & 0.69 & 0.65 \\
  & high$^\dagger$  & 2.38 & 2.29 & 2.25 & 2.70 & 2.41 \\
\midrule
\multirow{2}{*}{GPT-5 Mini}
  & minimal         & 0.70 & 1.22 & 0.67 & 0.91 & 0.88 \\
  & medium$^\dagger$ & 2.93 & 3.43 & 2.67 & 3.15 & 3.05 \\
\midrule
\multicolumn{7}{@{}l}{\textit{Premium frontier VLMs (provider-default settings)}} \\
Opus 4.6          & default$^\dagger$ (no thinking)   & 4.02 & 8.16 & 5.19 & 6.40 & 5.78 \\
Gemini 3.1 Pro    & default$^\dagger$ (high thinking) & 6.69 & 9.69 & 7.73 & 10.54 & 8.49 \\
GPT-5.4           & default$^\dagger$ (no reasoning)  & 2.55 & 4.04 & 2.13 & 3.00 & 2.90 \\
\bottomrule
\end{tabular}
\caption{Per-page cost (\textcent/page) across dataset categories for proprietary VLM configurations. All configurations include layout extraction. $\dagger$ marks default reasoning settings.}
\label{tab:vlm-costs}
\end{table*}

\subsection{Per-Dimension Quality vs.\ Cost}
\label{app:per-dim-cost}

\begin{figure*}[!htb]
\centering
\includegraphics[width=\textwidth]{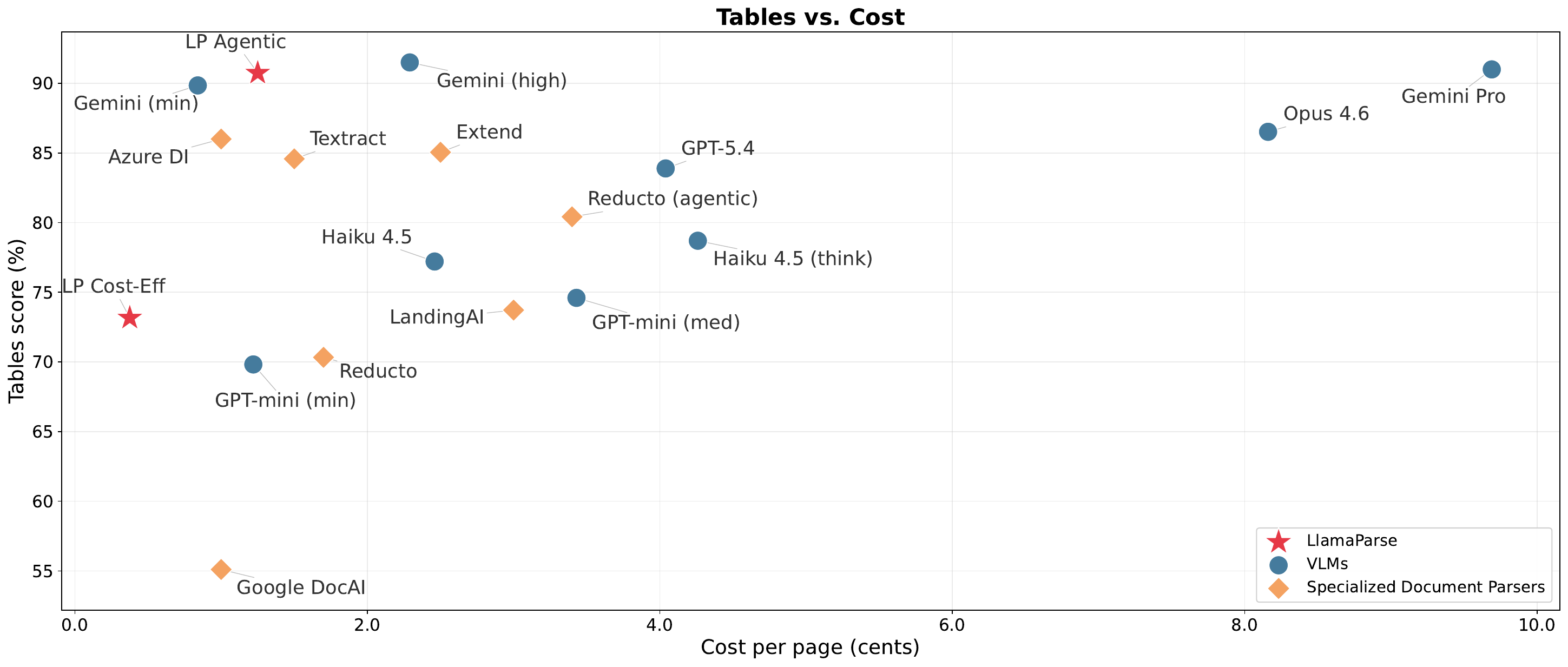}
\caption{Tables: quality vs.\ cost.}
\label{fig:cost-tables}
\end{figure*}

\begin{figure*}[!htb]
\centering
\includegraphics[width=\textwidth]{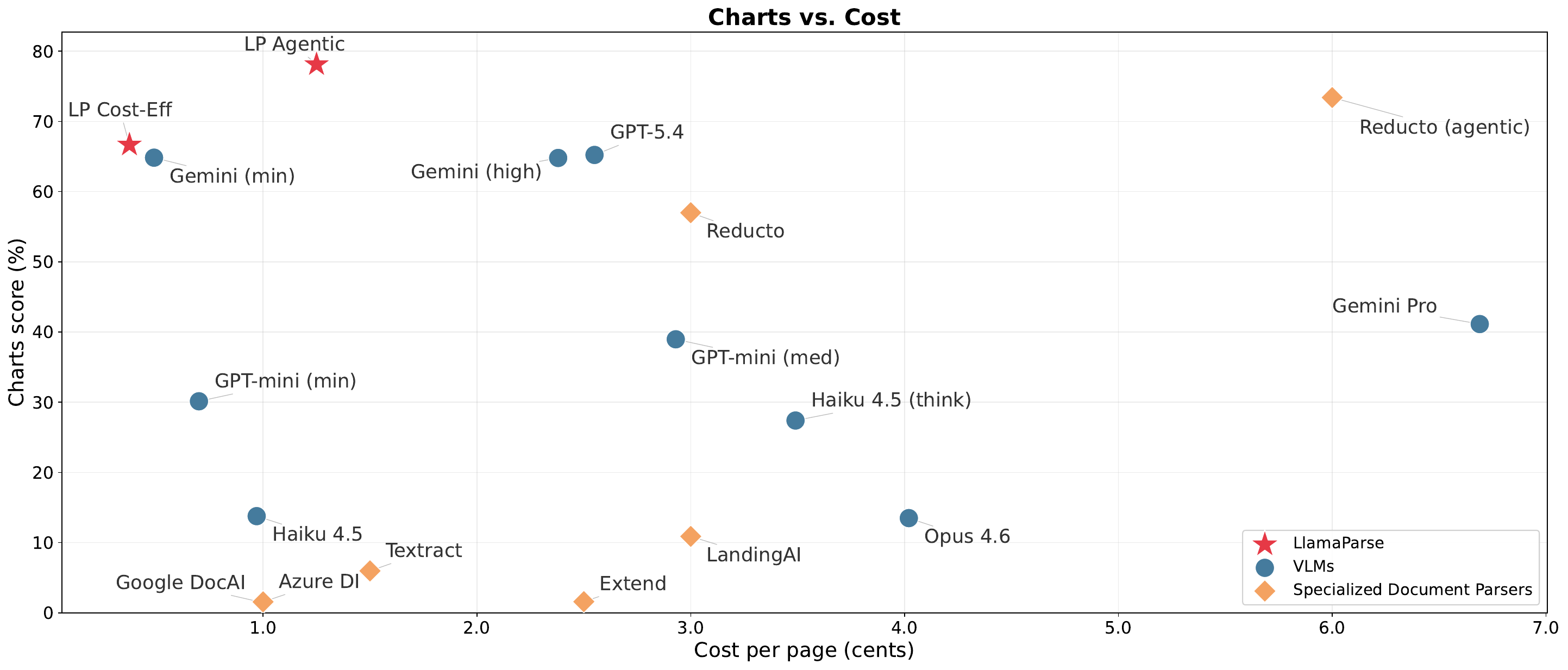}
\caption{Charts: quality vs.\ cost.}
\label{fig:cost-charts}
\end{figure*}

\begin{figure*}[!htb]
\centering
\includegraphics[width=\textwidth]{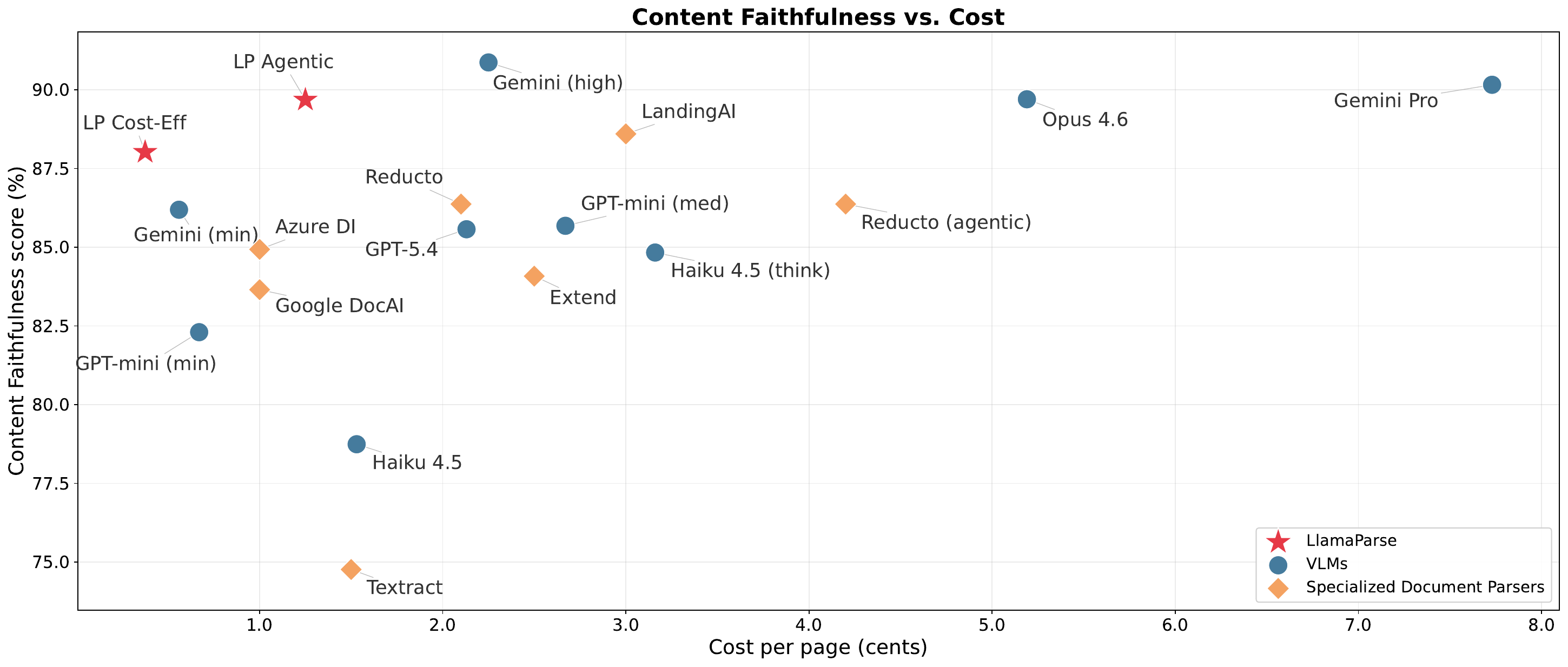}
\caption{Content Faithfulness: quality vs.\ cost.}
\label{fig:cost-content}
\end{figure*}

\begin{figure*}[!htb]
\centering
\includegraphics[width=\textwidth]{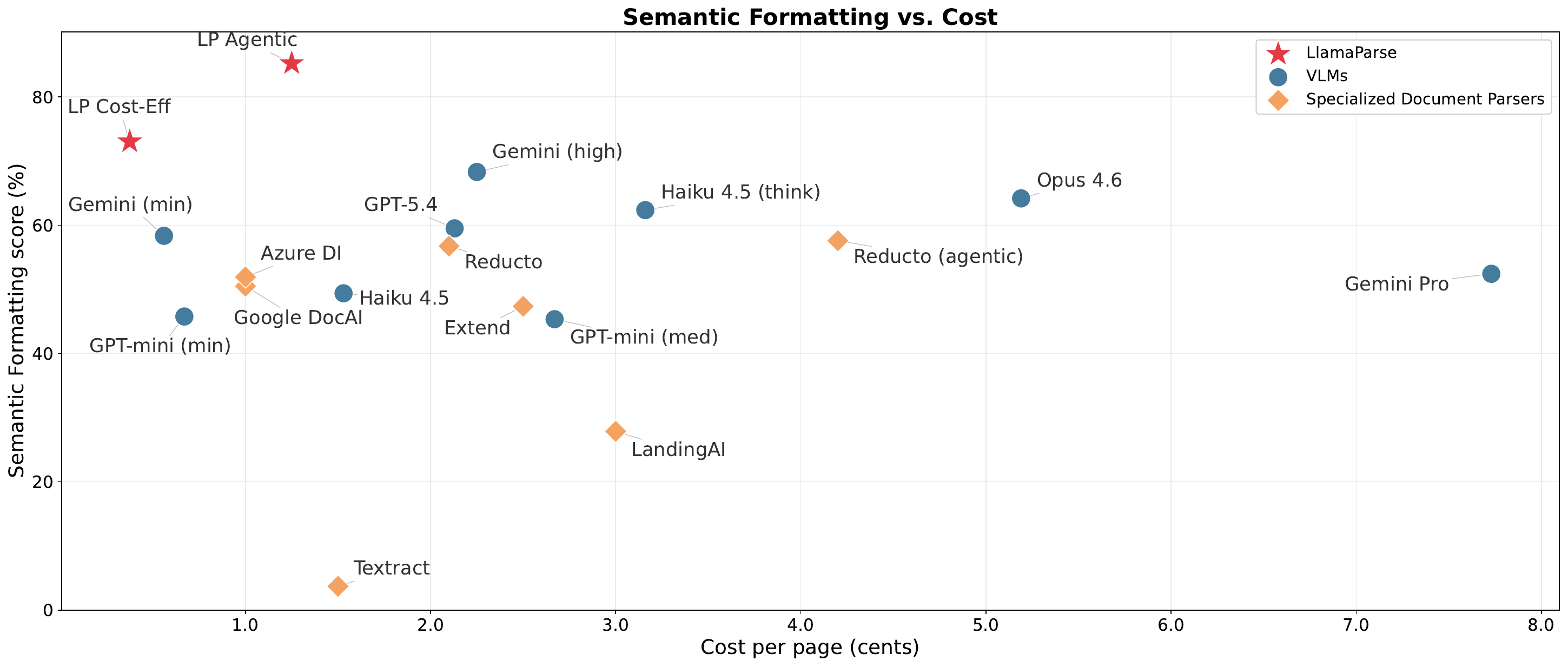}
\caption{Semantic Formatting: quality vs.\ cost.}
\label{fig:cost-formatting}
\end{figure*}

\begin{figure*}[!htb]
\centering
\includegraphics[width=\textwidth]{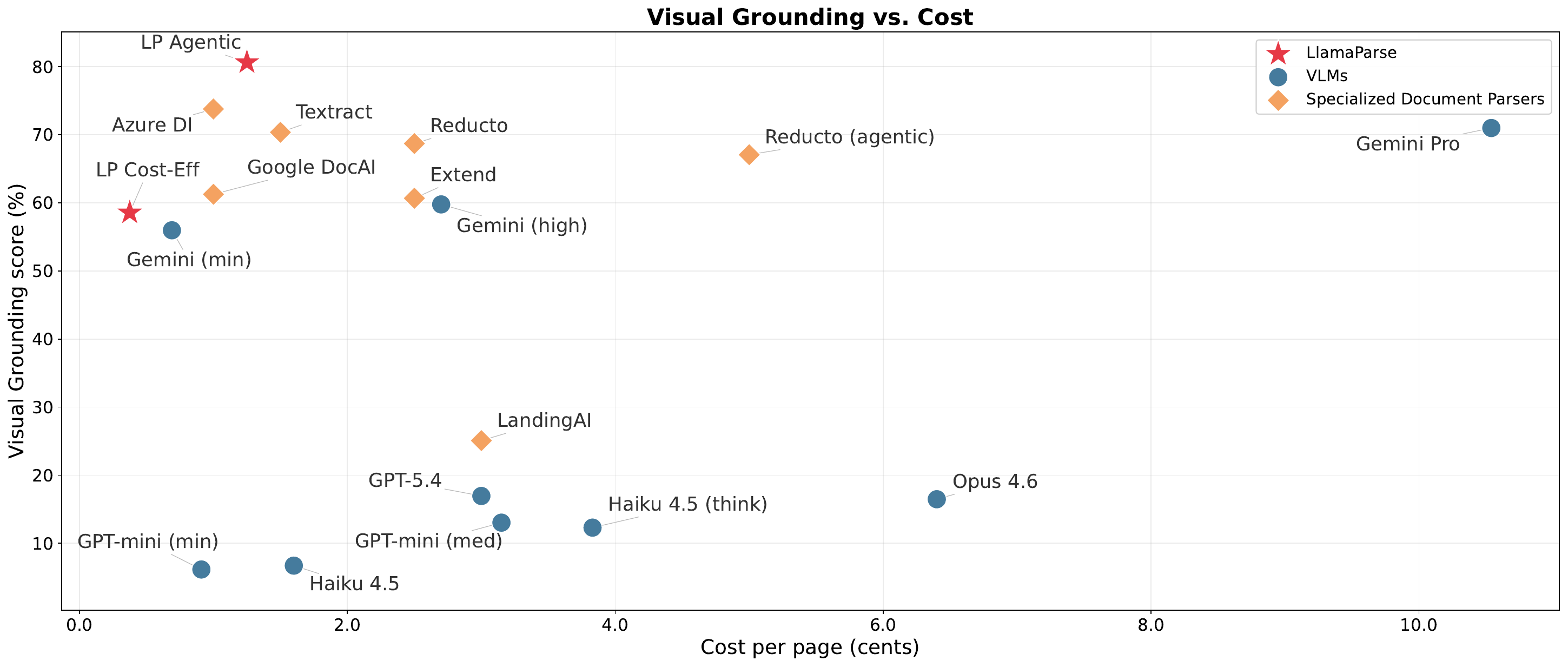}
\caption{Visual Grounding: quality vs.\ cost.}
\label{fig:cost-grounding}
\end{figure*}

\Cref{fig:cost-tables,fig:cost-charts,fig:cost-content,fig:cost-formatting,fig:cost-grounding} break down the overall quality-cost frontier (\Cref{fig:quality-cost}) into individual dimensions. Rather than a single global trade-off, the plots reveal five different frontier shapes: some dimensions have crowded high-quality regions where cost becomes decisive, while others have sparse frontiers where paying more is the only way to obtain the capability at all.

\paragraph{Tables} (\Cref{fig:cost-tables}): Tables show a relatively crowded frontier. Several providers already operate in the high-quality region, so the main question is not whether strong table parsing is attainable, but how much one should pay for the last few points. Gemini~(minimal), LlamaParse Cost Effective, and LlamaParse Agentic all sit near the top of the plot, while higher-reasoning variants buy only small additional gains. Gemini~3.1 Pro achieves the highest table score among VLMs (91.0\%) but at 13$\times$ the cost of Flash~(minimal). In this dimension, marginal quality improvements are real but relatively expensive, making lower-cost high-performing configurations attractive.

\paragraph{Charts} (\Cref{fig:cost-charts}): Charts have the sparsest frontier and the clearest quality-cost separation. Many low-cost and traditional IDP systems are simply non-competitive here, with scores near zero because they do not recover structured chart data at all. The meaningful trade-off is therefore among the few viable systems: LlamaParse Agentic leads, LlamaParse Cost Effective offers a much cheaper point that remains strong, and Gemini provides a competitive VLM alternative. Among the frontier models, GPT-5.4 is notably strong on charts (65.2\%), competitive with Gemini~Flash, while Opus~4.6 and Gemini~3.1 Pro score much lower. Unlike tables, this is a dimension where paying more can buy a qualitatively different level of capability rather than just a marginal improvement.

\paragraph{Content Faithfulness} (\Cref{fig:cost-content}): This is the most table-stakes dimension: every practical parser must clear a high bar here because omissions, hallucinations, and ordering errors directly corrupt the agent's context. The frontier is comparatively compressed, so higher spend does not radically separate providers, but the remaining gaps are still operationally meaningful. The cost story is therefore not that this dimension is ``solved,'' but that it imposes a minimum acceptable quality threshold below which a parser is unusable regardless of its strengths elsewhere.

\paragraph{Semantic Formatting} (\Cref{fig:cost-formatting}): Semantic formatting has a sparse frontier with large quality gaps. Many low-cost or traditional pipelines effectively opt out of the dimension, preserving text while discarding the formatting cues that carry meaning. LlamaParse Agentic clearly defines the top frontier point, while the rest of the field trails far behind. Here the relevant cost question is not how much to pay for a few extra points, but whether a system supports the capability at all.

\paragraph{Visual Grounding} (\Cref{fig:cost-grounding}): Visual grounding exposes a different frontier shape from the text-centric dimensions. VLM-only approaches such as GPT-5 Mini and Haiku~4.5 remain far from competitive even at higher spend, whereas layout-aware systems and agentic pipelines occupy the useful part of the curve. LlamaParse Agentic leads, with Azure DI and Textract forming the next tier. Among the frontier models, Gemini~3.1 Pro is the exception, achieving competitive grounding (71.0\%) likely due to its default high thinking budget, while GPT-5.4 and Opus~4.6 remain weak. This dimension shows that extra spend on generic reasoning is not enough by itself; the frontier favors systems with explicit layout and attribution capabilities.

\subsection{Fine-Grained Layout Metric Breakdown}
\label{app:layout-metric-breakdown}

\Cref{tab:layout-pass-breakdown,tab:layout-diagnostic-breakdown} unpack the headline visual-grounding score into finer-grained layout metrics for the same comparison set used in the main results.
The \texttt{Pages} column reports how many pages each provider returns usable grounding output for out of the nominal 500-page layout slice.
Metric values are aggregated over the pages counted in the \texttt{Pages} column, while the page count itself makes coverage differences explicit.
Qwen uses the same separate layout-only pipeline noted in \Cref{tab:main-results}; lower counts for several other providers reflect pages where the pipeline failed to return usable grounding output.

\paragraph{Rule-based pass rates.}
\textbf{Element Pass Rate} is the main joint metric: a ground-truth element counts as passing when the prediction satisfies localization, classification, and, when applicable, content attribution (see \Cref{app:grounding-aux-metrics}).
\textbf{Classification}, \textbf{Localization}, and \textbf{Attribution} report those stages individually.
\textbf{Reading Order} measures ordering rules among eligible layout elements.
\textbf{Layout Rule Pass Rate} is the broader average over all layout rules, so it mixes element-level checks with ordering and related layout constraints.

\paragraph{Standalone diagnostics.}
\textbf{mAP@[.50:.95]} is the standard COCO-style detection metric averaged over IoU thresholds from 0.50 to 0.95.
\textbf{LAP}, \textbf{LAR}, and \textbf{AF1} isolate local attribution quality without requiring the full element to pass the joint layout evaluation; formal definitions are given in \Cref{app:grounding-aux-metrics}.

\paragraph{Takeaway.}
The breakdown clarifies the pattern behind the main visual-grounding results.
LlamaParse Agentic leads the joint layout metrics and the main detection diagnostic, while LlamaParse Cost Effective is strongest on reading order.
Among specialized parsers, Azure is the strongest overall rule-based competitor and leads AF1 and LAR, while Textract leads LAP.
This separation is useful because strong local attribution alone does not guarantee that the system recovers the right element type, location, and page structure together.

\begin{table*}[t!]
\centering
\scriptsize
\begin{tabular*}{\textwidth}{@{\extracolsep{\fill}}p{0.24\textwidth}ccccccc@{}}
\toprule
\textbf{Provider} & \textbf{Pages} & \makecell{\textbf{Element}\\\textbf{Pass}} & \textbf{Cls.} & \textbf{Loc.} & \textbf{Attr.} & \makecell{\textbf{Reading}\\\textbf{Order}} & \makecell{\textbf{All Layout}\\\textbf{Rules}} \\
\midrule
\multicolumn{8}{@{}l}{\textit{VLMs}} \\
\quad OpenAI GPT-5 Mini & 500/500 & 6.1 & 11.5 & 17.8 & 16.6 & 32.3 & 15.2 \\
\quad Anthropic Haiku 4.5 & 499/500 & 6.7 & 13.4 & 18.8 & 11.0 & 39.3 & 14.5 \\
\quad Google Gemini 3 Flash & 497/500 & 56.0 & 60.0 & 68.5 & 71.9 & 82.2 & 66.5 \\
\quad Qwen 3 VL & 496/500 & 55.2 & 60.7 & 68.0 & 70.4 & 79.9 & 66.1 \\
\quad Dots OCR 1.5 & 492/500 & 55.8 & 60.0 & 64.6 & 65.3 & 81.0 & 63.2 \\
\midrule
\multicolumn{8}{@{}l}{\textit{Specialized Document Parsers}} \\
\quad Docling (OSS) & 487/500 & 66.1 & 73.2 & 77.1 & 68.5 & 77.6 & 73.1 \\
\quad AWS Textract & 500/500 & 70.4 & 75.7 & 85.5 & 84.2 & 80.4 & 81.7 \\
\quad Google Cloud Doc AI & 500/500 & 61.3 & 65.3 & 76.0 & 78.1 & 73.4 & 72.8 \\
\quad Azure Doc Intelligence & 499/500 & \underline{73.8} & \underline{78.6} & \underline{87.8} & \underline{88.7} & 73.6 & \underline{84.7} \\
\quad \reductoname & 500/500 & 68.7 & 73.8 & 83.7 & 82.9 & 78.8 & 80.0 \\
\quad \extendname & 499/500 & 60.7 & 64.8 & 78.8 & 79.0 & 69.3 & 73.9 \\
\quad \landingainame & 500/500 & 25.1 & 38.2 & 50.4 & 24.3 & 78.5 & 38.2 \\
\midrule
\multicolumn{8}{@{}l}{\textit{LlamaParse (Ours)}} \\
\quad LlamaParse Cost Effective & 498/500 & 58.6 & 71.3 & 74.9 & 68.3 & \textbf{84.6} & 71.6 \\
\quad LlamaParse Agentic & 500/500 & \textbf{80.6} & \textbf{85.0} & \textbf{90.0} & \textbf{89.0} & \underline{82.8} & \textbf{87.9} \\
\bottomrule
\end{tabular*}
\caption{Fine-grained rule-based layout metrics (\%). \textbf{Bold} marks the best score in each column; \underline{underlined values} mark the second-best.}
\label{tab:layout-pass-breakdown}
\end{table*}

\begin{table*}[t!]
\centering
\scriptsize
\begin{tabular*}{\textwidth}{@{\extracolsep{\fill}}p{0.28\textwidth}ccccc@{}}
\toprule
\textbf{Provider} & \textbf{Pages} & \textbf{mAP@[.50:.95]} & \textbf{AF1} & \textbf{LAP} & \textbf{LAR} \\
\midrule
\multicolumn{6}{@{}l}{\textit{VLMs}} \\
\quad OpenAI GPT-5 Mini & 500/500 & 2.1 & 40.8 & 39.5 & 44.2 \\
\quad Anthropic Haiku 4.5 & 499/500 & 1.9 & 42.4 & 40.8 & 47.2 \\
\quad Google Gemini 3 Flash & 497/500 & 42.8 & 83.8 & 84.0 & 85.4 \\
\quad Qwen 3 VL & 496/500 & 32.1 & 86.7 & 91.4 & 85.1 \\
\quad Dots OCR 1.5 & 492/500 & 39.4 & 77.8 & 84.9 & 76.0 \\
\midrule
\multicolumn{6}{@{}l}{\textit{Specialized Document Parsers}} \\
\quad Docling (OSS) & 487/500 & 42.8 & 86.2 & 94.9 & 83.5 \\
\quad AWS Textract & 500/500 & \underline{60.2} & 93.3 & \textbf{96.8} & 91.7 \\
\quad Google Cloud Doc AI & 500/500 & 41.5 & 92.0 & 95.9 & 91.3 \\
\quad Azure Doc Intelligence & 499/500 & 48.8 & \textbf{94.2} & 92.5 & \textbf{97.4} \\
\quad \reductoname & 500/500 & 48.3 & \underline{93.4} & 95.9 & 92.8 \\
\quad \extendname & 499/500 & 42.0 & 90.2 & \underline{96.1} & 88.4 \\
\quad \landingainame & 500/500 & 19.9 & 76.4 & 76.2 & 80.3 \\
\midrule
\multicolumn{6}{@{}l}{\textit{LlamaParse (Ours)}} \\
\quad LlamaParse Cost Effective & 498/500 & 56.4 & 80.9 & 81.8 & 83.4 \\
\quad LlamaParse Agentic & 500/500 & \textbf{64.4} & 91.2 & 88.2 & \underline{96.7} \\
\bottomrule
\end{tabular*}
\caption{Detection and attribution-only layout diagnostics (\%). \textbf{Bold} marks the best score in each column; \underline{underlined values} mark the second-best.}
\label{tab:layout-diagnostic-breakdown}
\end{table*}

\subsection{Difficulty-Stratified Layout Performance}
\label{app:layout-difficulty}

\begin{figure*}[t!]
\centering
\includegraphics[width=0.95\textwidth]{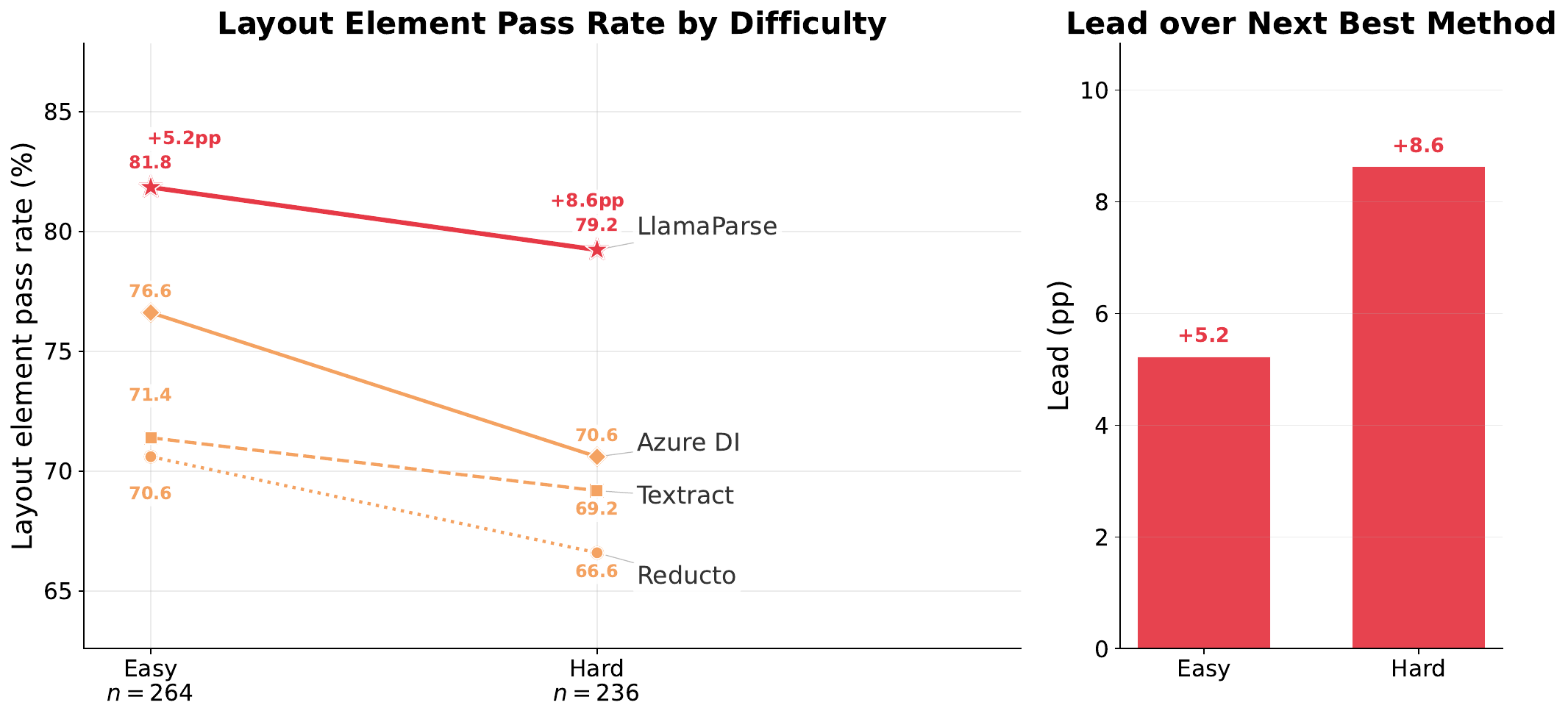}
\caption{Difficulty-stratified layout element pass rate on two reported buckets, \texttt{Easy} and \texttt{Hard}, for LlamaParse and the three strongest specialized parsers. The right panel reports LlamaParse's absolute lead over the next-best method in each bucket.}
\label{fig:layout-difficulty}
\end{figure*}

Aggregate layout scores average over pages with very different structural complexity.
\Cref{fig:layout-difficulty} therefore contrasts \texttt{Easy} and \texttt{Hard} pages directly. LlamaParse leads the next-best method in both: +5.2 points on easy pages and +8.6 on the harder subset.
This makes the appendix point cleaner: the advantage is not confined to routine pages, and it remains visible when evaluation is concentrated on the more structurally complex half of the benchmark.

\paragraph{Difficulty stratification heuristic.}
Difficulty is assigned \emph{per page}, since the layout benchmark and grounding evaluation are page-based.
We use a deterministic, hand-tuned heuristic implemented in \texttt{layout\_difficulty.py}, rather than a learned classifier.
The score combines annotation-derived page complexity with a small PDF native-text corruption signal. For the appendix comparison above, pages with scores $\leq 3$ are reported as \texttt{Easy}, and all higher-scoring pages are reported as \texttt{Hard}.
Threshold contributions are cumulative: for example, \texttt{rule\_count >= 45, 60, 80, 110} means a page with 87 layout elements receives +3 from that feature alone.

The heuristic is designed to capture several failure modes that correlate with difficult grounding pages: dense layouts with many annotated elements; structurally fragmented reading order; large pages or pages with high annotation coverage; scanned or image-backed assets; and PDFs whose native extracted text appears corrupted.
Concretely, \texttt{rule\_count} counts only \texttt{layout} annotations, \texttt{fragmented\_row\_count} counts text-like rows containing many separate boxes, \texttt{horizontal\_band\_count} estimates the number of distinct text bands or columns, \texttt{upward\_reset\_ratio} measures how often reading order jumps upward on the page, and \texttt{bbox\_coverage} is the total normalized area covered by layout boxes.

\begin{table*}[t!]
\centering
\small
\begin{tabular}{p{0.18\textwidth}p{0.38\textwidth}p{0.16\textwidth}}
\toprule
\textbf{Signal family} & \textbf{Feature and thresholds} & \textbf{Contribution} \\
\midrule
Layout density & \texttt{rule\_count >= 45, 60, 80, 110} & +1 each \\
Layout density & \texttt{picture\_count >= 4, 10} & +1 each \\
Layout density & \texttt{table\_count >= 1, 3} & +1 each \\
Structural fragmentation & \texttt{fragmented\_row\_count >= 5, 9} & +1 each \\
Structural fragmentation & \texttt{horizontal\_band\_count >= 3, 4} & +1 each \\
Structural fragmentation & \texttt{upward\_reset\_ratio >= 0.12} & +1 \\
Page scale and coverage & \texttt{page\_area >= 750k, 1.5M} & +1 each \\
Page scale and coverage & \texttt{bbox\_coverage >= 0.55} & +1 \\
Asset type & asset is not a PDF & +1 \\
Native-text quality & native PDF text flagged as buggy & +3 \\
Native-text quality & \makecell[l]{\texttt{native\_text\_unusual\_}\\\texttt{punctuation\_count >= 100}} & +1 \\
\bottomrule
\end{tabular}
\caption{Deterministic page-level heuristic used to score page difficulty for the appendix \texttt{Easy}/\texttt{Hard} analysis. Pages with scores $\leq 3$ are reported as \texttt{Easy}; all higher-scoring pages are reported as \texttt{Hard}.}
\label{tab:layout-difficulty-heuristic}
\end{table*}

\section{Qualitative Examples}
\label{app:qualitative}
\subsection{Tables: \textsc{TableRecordMatch} vs. GriTS and TEDS}
\label{app:qual-tables}
This section presents examples where GriTS and TEDS are poorly correlated with a table's usefulness for downstream applications, due to their symmetric treatment of header and data cells.

Both metrics consider a table spatially, as a grid or tree of cells, and compare cell location and content without taking into account how header and data cells have different semantics. So, a small edit to header text barely moves the scores even when it completely changes a table's semantics, while a semantics-preserving column reordering is heavily penalized. \textsc{TableRecordMatch} (TRM) instead converts each table to a bag of column-keyed records, which makes it sensitive to header semantics and invariant to column order.

We note that by default, TEDS is completely blind to header content. So, before computing TEDS on the examples below, we demote \texttt{<thead>}/\texttt{<th>} to data cells. TEDS would be even more permissive of header differences without this adjustment.

\paragraph{Results overview.}

\begin{figure}[!ht]
\centering
\begin{subfigure}[t]{\textwidth}
    \centering
    \includegraphics[width=0.85\textwidth]{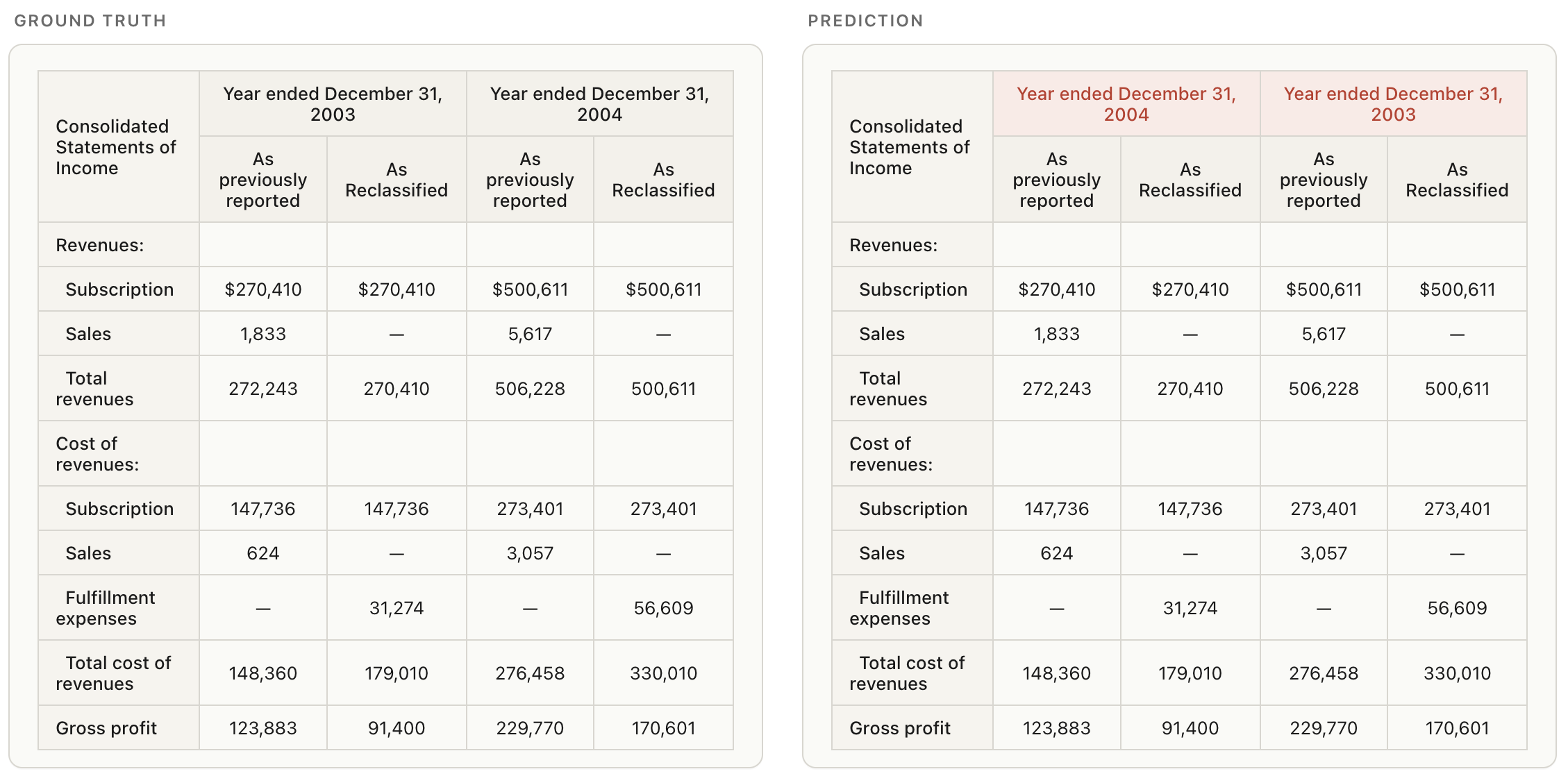}
    \caption{\textbf{Headers swapped, semantics destroyed.} The predicted table is identical to the ground truth except for the two highlighted top-level header cells, which are swapped. Every revenue figure now sits under the wrong fiscal year, yet GriTS${}=0.998$ and TEDS${}=0.999$. In contrast, TRM${}=0.480$, correctly penalizing the misattribution of all the numeric values in the table.}
    \label{fig:trm-year-swap}
\end{subfigure}

\vspace{0.8em}

\begin{subfigure}[t]{\textwidth}
    \centering
    \includegraphics[width=0.85\textwidth]{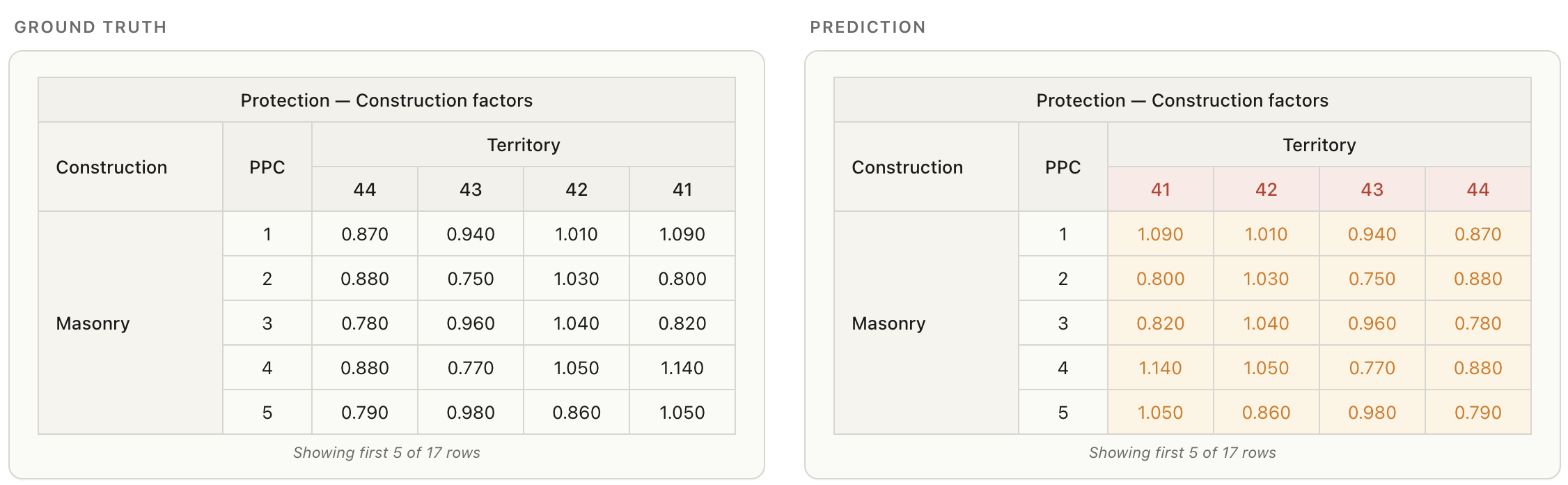}
    \caption{\textbf{Columns reordered, semantics preserved.} The prediction reverses the \texttt{Territory} column order and permutes the body to match, so every (Construction, PPC, Territory) $\rightarrow$ factor record is preserved. Highlighted cells differ positionally from the ground truth but encode the same semantic content. GriTS${}=0.740$ and TEDS${}=0.716$, heavily penalizing the content difference between positionally-matched cells. TRM${}=1.000$, correctly recognizing that the two tables represent the same bag of records.}
    \label{fig:trm-territory-reverse}
\end{subfigure}
\caption{Two contrasting predictions where GriTS and TEDS disagree with the table's downstream usefulness. (a)~A semantically wrong prediction that GriTS and TEDS round to $\approx 1.0$. (b)~A semantically equivalent rewrite that GriTS and TEDS heavily penalize. TRM ranks both correctly.}
\label{fig:trm-vs-grits-teds}
\end{figure}

\Cref{fig:trm-year-swap} and \Cref{fig:trm-territory-reverse} demonstrate cases where GriTS and TEDS ignore semantic differences and over-penalize cosmetic differences, while TRM is correctly captures table semantics.

\paragraph{Takeaway.}
These two examples highlight the failure modes of position-based table metrics: GriTS and TEDS can simultaneously under-penalize a structurally correct but semantically inverted prediction (\Cref{fig:trm-year-swap})  and over-penalize a semantically correct but structurally rearranged one (\Cref{fig:trm-territory-reverse}). TRM's record-set view has neither of these failure modes, providing a semantics-based score that aligns with a table's utility to downstream consumers, who understand tables as structured data rather than as grids of pixels.

\subsection{Charts: 3D Grouped Bar Chart}
\label{app:qual-charts}

This section presents selected case studies that illustrate how the benchmark behaves on representative failure modes across four dimensions.

\begin{figure}[!ht]
\centering
\begin{subfigure}[t]{0.48\textwidth}
    \centering
    \includegraphics[width=\textwidth]{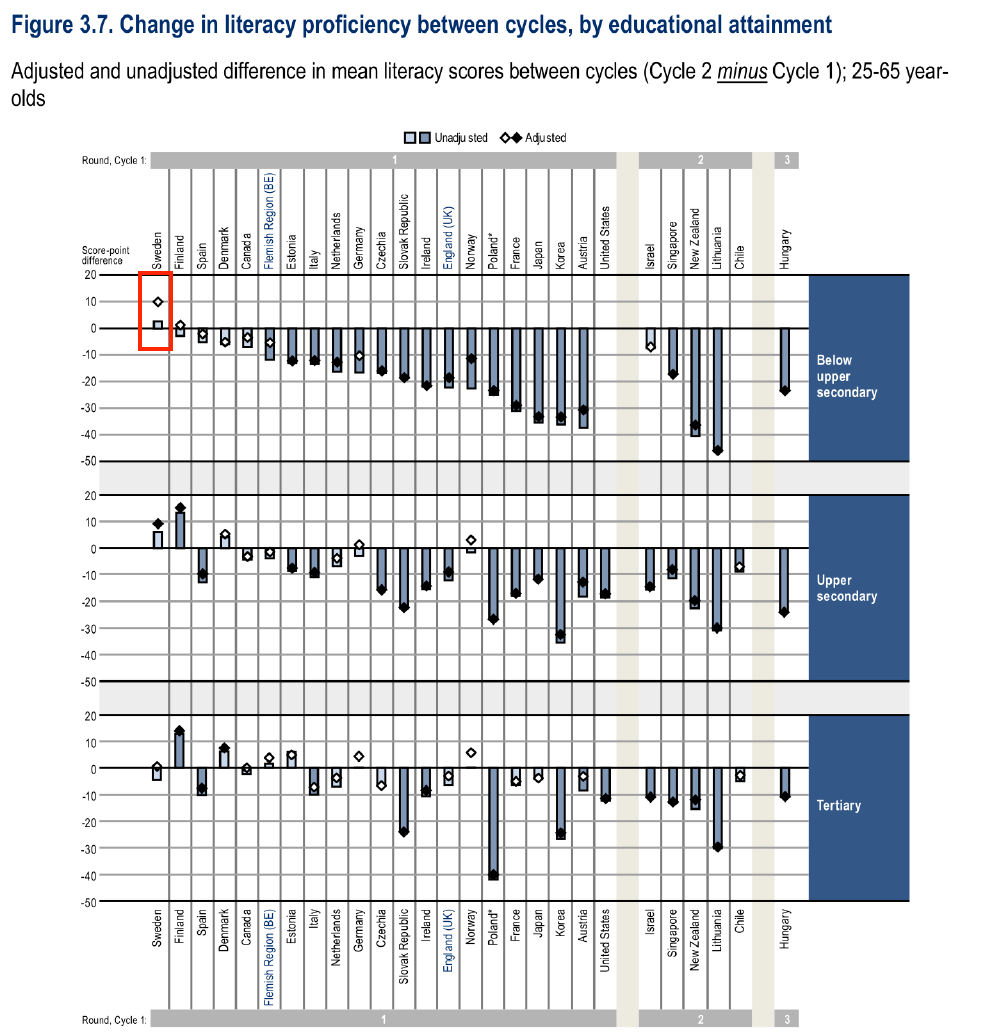}
    \caption{Source chart (OECD Figure~3.7). Red box marks Sweden / Below upper secondary.}
    \label{fig:chart-source}
\end{subfigure}
\hfill
\begin{subfigure}[t]{0.48\textwidth}
    \centering
    \includegraphics[width=0.70\textwidth,page=1]{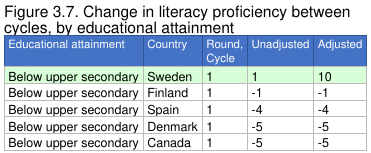}
    \caption{LlamaParse Agentic (8/10).}
    \label{fig:chart-ours}
\end{subfigure}

\vspace{0.6em}
\begin{subfigure}[t]{0.31\textwidth}
    \centering
    \includegraphics[width=\textwidth,page=1]{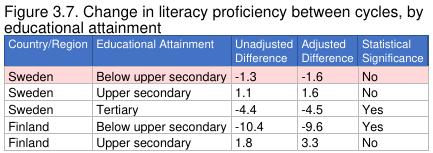}
    \caption{\reductoname{} (3/10).}
    \label{fig:chart-reducto}
\end{subfigure}
\hfill
\begin{subfigure}[t]{0.31\textwidth}
    \centering
    \includegraphics[width=\textwidth,page=1]{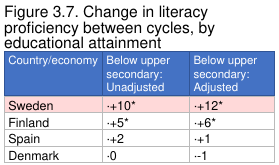}
    \caption{GPT-5 Mini, minimal reasoning (0/10).}
    \label{fig:chart-gpt5}
\end{subfigure}
\hfill
\begin{subfigure}[t]{0.31\textwidth}
    \centering
    \includegraphics[width=\textwidth,page=1]{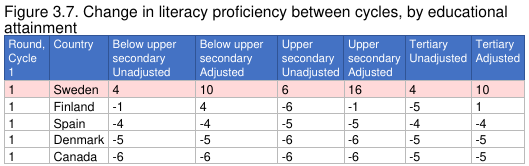}
    \caption{Gemini 3 Flash, high thinking (3/10).}
    \label{fig:chart-gemini-high}
\end{subfigure}
\caption{Source chart and provider outputs for the OECD chart example. (a)~Source chart with Sweden / Below upper secondary marked in red. (b)--(e)~Rendered table outputs (top portion shown); Sweden row highlighted in green (pass) or pink (fail). Gemini~3 Flash with minimal thinking omitted (no table, 0/10).}
\label{fig:chart-outputs}
\end{figure}

We illustrate the \textsc{ChartDataPointMatch} evaluation on a challenging example (\texttt{b263dc5d-en\_p119}): an OECD 3D grouped bar chart (\Cref{fig:chart-source}) where each data point requires \emph{three} labels to locate (education level, country, adjustment type).
The axes lack explicit group separators and no values are printed on the chart, so a VLM must fully comprehend the chart structure, legend, and 3D perspective before it can correctly read any single data point.

\paragraph{Test rules and tolerance.}
From the 10 annotated data points for this chart, we highlight two representative rules for Sweden / Below upper secondary in \Cref{tab:chart-rules-example}.
Each rule specifies a set of \emph{labels} that must co-occur with a \emph{value} in the parser's output table. The \texttt{relative\_tolerance} defines an acceptance range around the expected value: for example, a value of $10$ with a relative tolerance of $0.1$ (i.e., $\pm 10\%$) accepts any value in $[10 - 10 \times 0.1,\; 10 + 10 \times 0.1] = [9, 11]$.
These tolerances are intentionally generous, set during human verification based on how precisely each value can be read from the chart, to account for the inherent imprecision of visual estimation from charts without printed data labels.

\begin{table}[h]
\centering
\small
\begin{tabular}{@{}lrrr@{}}
\toprule
\textbf{Labels} & \textbf{Value} & \textbf{Relative tolerance} & \textbf{Accepted range} \\
\midrule
Below upper secondary, Sweden, Adjusted & 10 & 0.1 & $[9, 11]$ \\
Below upper secondary, Sweden, Unadjusted & 3 & 0.5 & $[1.5, 4.5]$ \\
\bottomrule
\end{tabular}
\caption{Two representative test rules for the OECD chart example (Sweden / Below upper secondary). The wide 0.5 relative tolerance on the Unadjusted rule reflects that the bar is very short ($\approx$3 score points) and difficult to read precisely from the 3D chart; any estimate between 1.5 and 4.5 is accepted.}
\label{tab:chart-rules-example}
\end{table}

\paragraph{Results overview.}
\Cref{tab:chart-scores} shows each provider's pass rate on this example (10 rules total). LlamaParse Agentic passes 8 of 10 rules; the best competing method passes only 3 of 10.

\begin{table}[h]
\centering
\small
\begin{tabular}{@{}lrl@{}}
\toprule
\textbf{Provider} & \textbf{Passed} & \textbf{Primary failure mode} \\
\midrule
LlamaParse Agentic & \textbf{8/10} & Minor label association gaps \\
Gemini 3 Flash (high thinking) & 3/10 & Labels not associated with values \\
\reductoname & 3/10 & Incorrect values \\
GPT-5 Mini (minimal reasoning) & 0/10 & Incorrect values \\
Gemini 3 Flash (minimal thinking) & 0/10 & No table extracted \\
\bottomrule
\end{tabular}
\caption{Per-provider results on the OECD chart example (10 test rules).}
\label{tab:chart-scores}
\end{table}

\paragraph{Provider output comparison.}
\Cref{fig:chart-outputs} shows the rendered table outputs from four providers. In each output, the Sweden row is highlighted in \textcolor{green!50!black}{green} if the test rules pass or \textcolor{red!70!black}{pink} if they fail. Every provider uses a different table structure, yet the evaluation is agnostic to these differences: it matches labels and values, not table layout.

\noindent Focusing on the two Sweden / Below upper secondary rules from \Cref{tab:chart-rules-example}:

\begin{itemize}
    \item \textbf{LlamaParse Agentic} (\Cref{fig:chart-ours}, 8/10): Both Sweden rules pass. Adjusted${}=10$ is within $[9,11]$, and the Unadjusted value is matched within $[1.5,4.5]$ with all three labels correctly associated.

    \item \textbf{\reductoname} (\Cref{fig:chart-reducto}, 3/10): Uses a completely different Markdown table layout, yet the metric still evaluates it fairly. Sweden values: Unadjusted${}=-1.3$, Adjusted${}=-1.6$, both wrong (the chart clearly shows Sweden as the only country with positive scores). Still passes 3/10 rules where its values happen to be correct, showing the evaluation credits correct values regardless of table format.

    \item \textbf{GPT-5 Mini, minimal reasoning} (\Cref{fig:chart-gpt5}, 0/10): Sweden values: Unadjusted${}=10$ (expected range $[1.5,4.5]$), Adjusted${}=12$ (expected range $[9,11]$), both outside tolerance. The values are simply wrong; the model lacks the visual precision to read this 3D chart. All 10 rules fail.

    \item \textbf{Gemini 3 Flash, high thinking} (\Cref{fig:chart-gemini-high}, 3/10): Uses a wide column-major layout, yet the metric can still match values. Sweden: Adjusted${}=10$ (pass), Unadjusted${}=4$ (within $[1.5,4.5]$, pass). Fails on 7 other rules.

    \item \textbf{Gemini 3 Flash, minimal thinking} (0/10, not shown): Produces only a figure description with no table or value at all. Without an extracted table, no data points can be verified.
\end{itemize}

\paragraph{Takeaway.}
This example illustrates three key properties of the evaluation:
\begin{enumerate}
    \item \textbf{Structure-agnostic.} All four table-producing providers use different formats, yet the evaluation handles them equivalently by searching for label-value co-occurrences, not specific table layouts. \reductoname{} and Gemini both score 3/10 despite structures entirely different from LlamaParse.
    \item \textbf{Generous tolerances, fundamental failures.} Tolerances of 5--50\% accommodate visual estimation imprecision, yet most providers still fail because their errors are not small estimation differences but fundamentally wrong values (e.g., $-1.3$ for a clearly positive data point).
    \item \textbf{3D chart parsing remains challenging for VLMs.} GPT-5 Mini (minimal reasoning) scores 0/10 with systematically wrong values; Gemini (high thinking) achieves only 3/10. Accurate chart-to-table conversion requires both strong visual reasoning and sufficient compute budget.
\end{enumerate}

\subsection{Content Faithfulness: Multi-Column Federal Register}
\label{app:qual-text}

\begin{figure}[!ht]
\centering
\begin{subfigure}[t]{0.48\textwidth}
    \centering
    \includegraphics[width=0.82\textwidth]{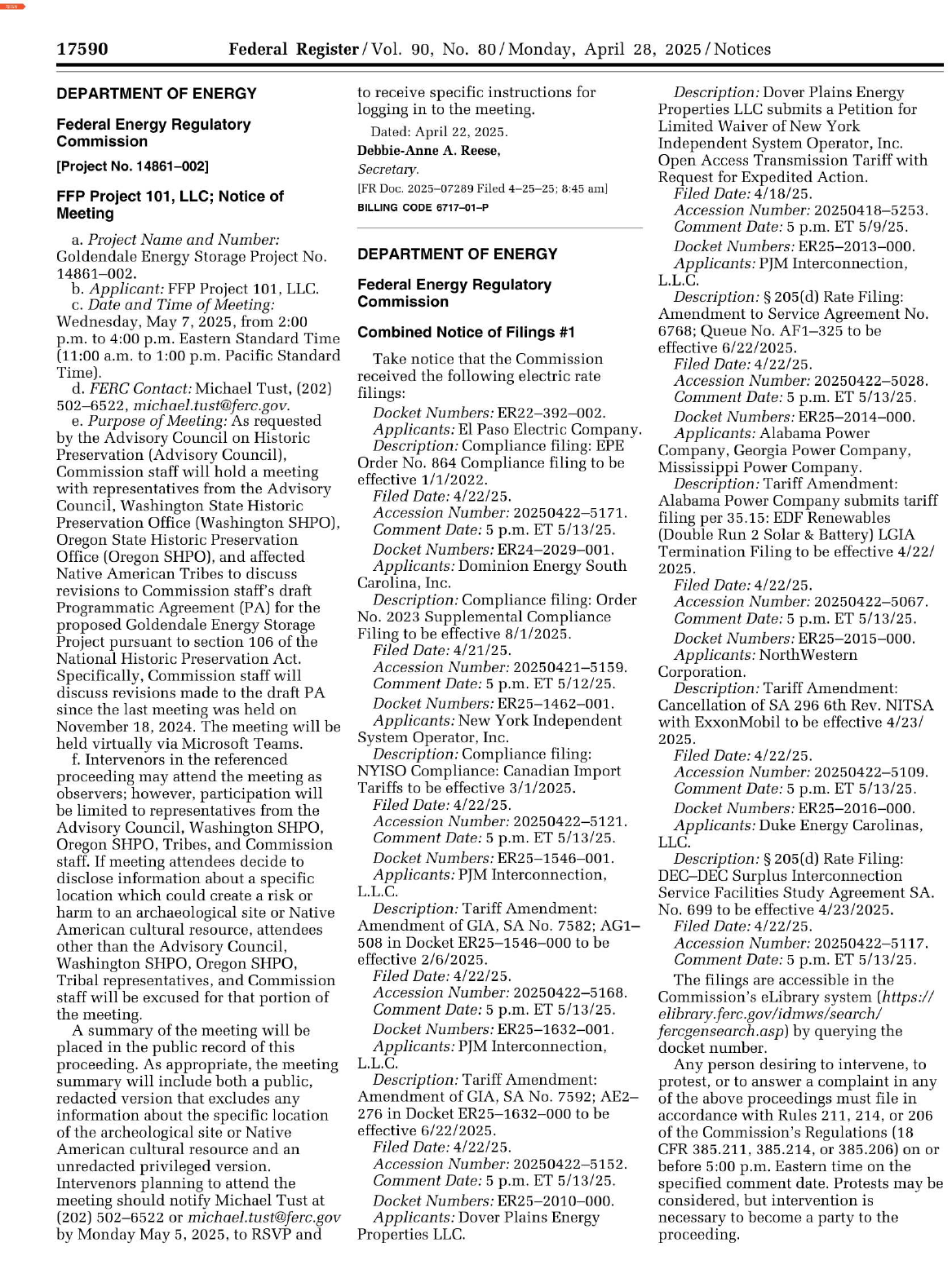}
    \caption{Source page (Federal Register Vol.~90, No.~80). 3-column layout with DOE notice and docket entries.}
    \label{fig:text-source}
\end{subfigure}
\hfill
\begin{subfigure}[t]{0.48\textwidth}
    \centering
    \includegraphics[width=\textwidth,page=1]{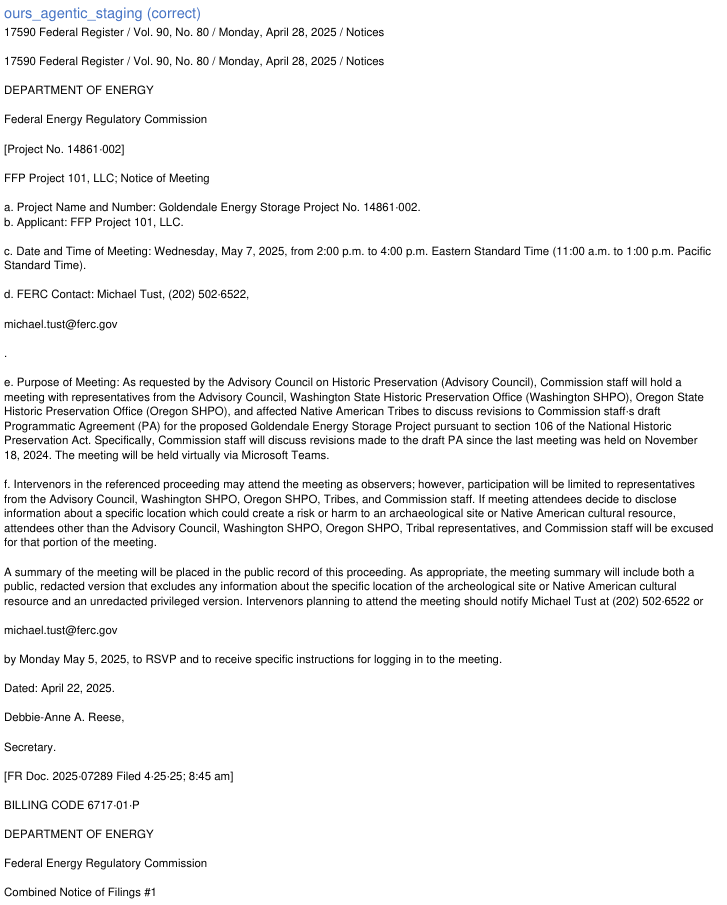}
    \caption{LlamaParse Agentic (0.966).}
    \label{fig:text-ours}
\end{subfigure}

\vspace{0.6em}
\begin{subfigure}[t]{0.31\textwidth}
    \centering
    \includegraphics[width=\textwidth,page=1]{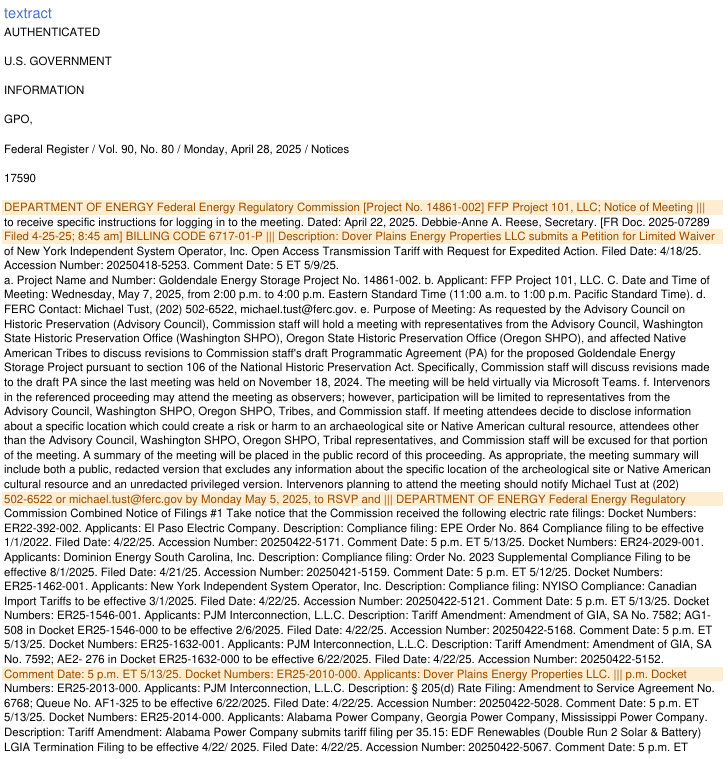}
    \caption{Textract (0.780).}
    \label{fig:text-textract}
\end{subfigure}
\hfill
\begin{subfigure}[t]{0.31\textwidth}
    \centering
    \includegraphics[width=\textwidth,page=1]{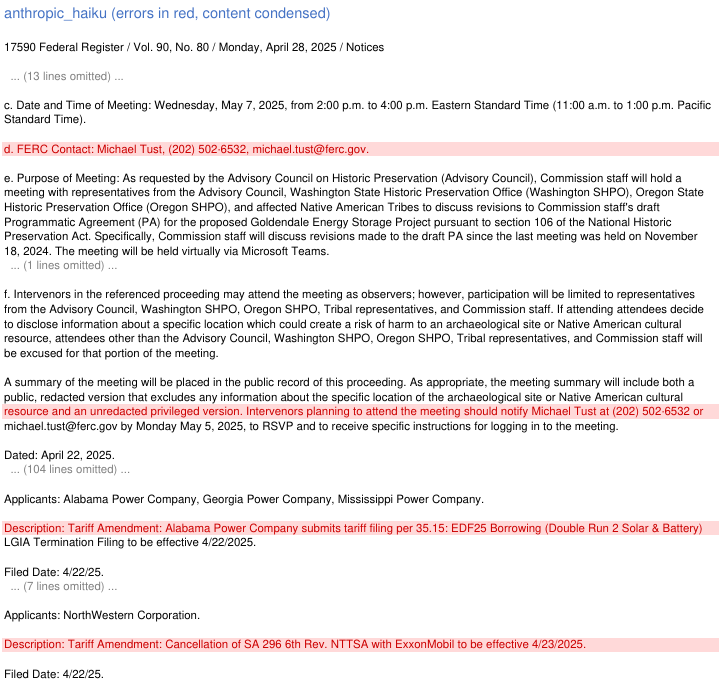}
    \caption{Haiku~4.5 (0.659).}
    \label{fig:text-haiku}
\end{subfigure}
\hfill
\begin{subfigure}[t]{0.31\textwidth}
    \centering
    \includegraphics[width=\textwidth,page=1]{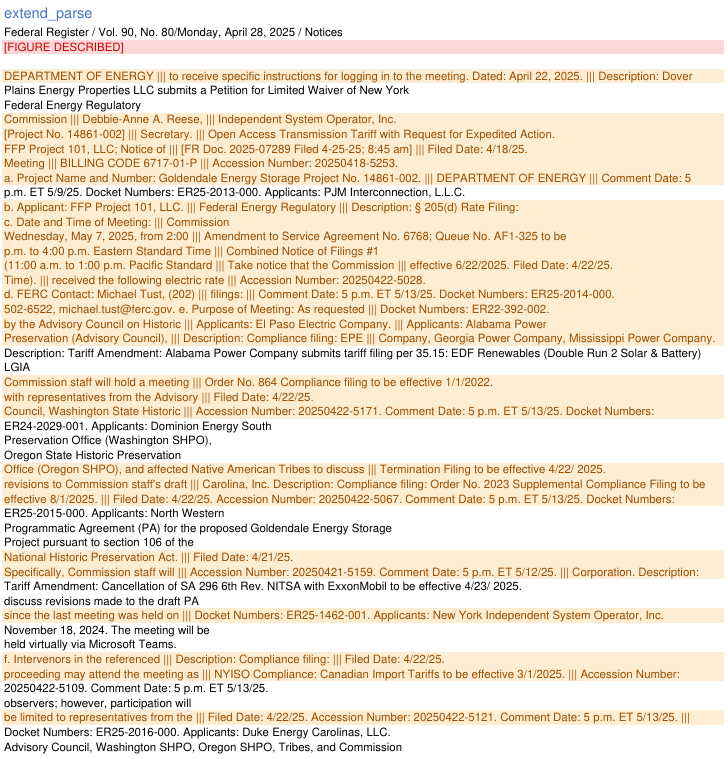}
    \caption{\extendname{} (0.577).}
    \label{fig:text-extend}
\end{subfigure}
\caption{Source page and provider outputs for the Federal Register example. (a)~Source 3-column page. (b)--(e)~Rendered outputs. LlamaParse Agentic correctly linearizes columns; Textract wraps them into an HTML table; \extendname{} reads row-wise across columns; Haiku~4.5 hallucinates words and jumbles column order.}
\label{fig:text-outputs}
\end{figure}

We illustrate the Content Faithfulness evaluation on a 3-column Federal Register page (\texttt{text\_multicolumns/energy}, \Cref{fig:text-source}).
Multi-column documents require parsers to linearize a 2D layout into the correct 1D reading sequence: each column must be read top-to-bottom before moving to the next.
Errors in column detection produce jumbled text, duplicated content, or hallucinated artifacts.

\paragraph{Metric overview.}
Content Faithfulness (\Cref{sec:content-faithfulness}) is a weighted average:
$\text{CFS} = (1.0 \cdot S_{\text{text}} + 0.5 \cdot S_{\text{order}}) / 1.5$,
where \emph{text correctness} checks for omissions, hallucinations, and duplications, and \emph{order} verifies reading sequence through pairwise precedence assertions.

\paragraph{Results overview.}
\Cref{tab:text-scores} shows per-provider scores. LlamaParse Agentic achieves 0.966 with perfect reading order; each competitor fails in a different way.

\begin{table}[h]
\centering
\small
\begin{tabular}{@{}lrrr@{}}
\toprule
\textbf{Provider} & \textbf{Content Faithfulness} & \textbf{Text Correctness} & \textbf{Order} \\
\midrule
LlamaParse Agentic & \textbf{0.966} & \textbf{0.949} & \textbf{1.000} \\
Textract            & 0.780 & 0.720 & 0.901 \\
Haiku~4.5           & 0.659 & 0.876 & 0.225 \\
\extendname              & 0.577 & 0.648 & 0.437 \\
\bottomrule
\end{tabular}
\caption{Per-provider Content Faithfulness scores on the Federal Register example. Each competitor exhibits a distinct failure mode: Textract duplicates content, Haiku~4.5 hallucinates text and jumbles column order, and \extendname{} reads across columns row-wise.}
\label{tab:text-scores}
\end{table}

\paragraph{Provider output comparison.}
\Cref{fig:text-outputs} shows the rendered outputs. The 3-column layout causes qualitatively different failures across providers.

\begin{itemize}
    \item \textbf{LlamaParse Agentic} (\Cref{fig:text-ours}, 0.966): Correctly reads all three columns top-to-bottom in sequence. Perfect reading order (1.000) and near-perfect text correctness (0.949).

    \item \textbf{Textract} (\Cref{fig:text-textract}, 0.780): Wraps the 3-column page into a 2-row HTML table, splitting each column at a horizontal cutoff. Reading order is reasonable (0.901), but content duplication drags text correctness to 0.720: only 34\% of words pass the duplication check, indicating substantial repeated content.

    \item \textbf{Haiku~4.5} (\Cref{fig:text-haiku}, 0.659): Achieves decent text correctness (0.876) but catastrophic order (0.225). The model confuses the column~2/3 boundary where docket entries span both columns, interleaving entries from the two columns. It also hallucinates specific content: phone number ``502--6522'' becomes ``502--6532,'' and company name ``EDF Renewables'' becomes ``EDF25 Borrowing.''

    \item \textbf{\extendname} (\Cref{fig:text-extend}, 0.577): Treats the 3-column layout as a single HTML table, reading \emph{across} columns at the same vertical position. The first table row merges ``DEPARTMENT OF ENERGY'' (column~1 top), ``to receive specific instructions'' (column~2 top), and ``Description: Dover Plains Energy'' (column~3 top), three completely unrelated text fragments. This produces both low text correctness (0.648) and low order (0.437).
\end{itemize}

\paragraph{Takeaway.}
This example illustrates two key properties of Content Faithfulness evaluation:
\begin{enumerate}
    \item \textbf{Orthogonal failure modes.} The text correctness and order sub-scores reveal qualitatively different failures. Haiku~4.5 extracts most words correctly (0.876) but destroys reading order (0.225); Textract preserves order (0.901) but duplicates content (0.720). A single aggregate score would mask these distinct error profiles.
    \item \textbf{Multi-column layout remains challenging.} Three of four providers score below 0.80 on a standard government document. The failures are not OCR errors but structural: incorrect column detection, row-wise reading, and column boundary confusion. These are precisely the errors that break downstream agentic workflows where reading order determines context.
\end{enumerate}

\subsection{Semantic Formatting: Infographic with Heading Hierarchy}
\label{app:qual-formatting}

\begin{figure}[!ht]
\centering
\begin{subfigure}[t]{0.48\textwidth}
    \centering
    \includegraphics[width=0.72\textwidth]{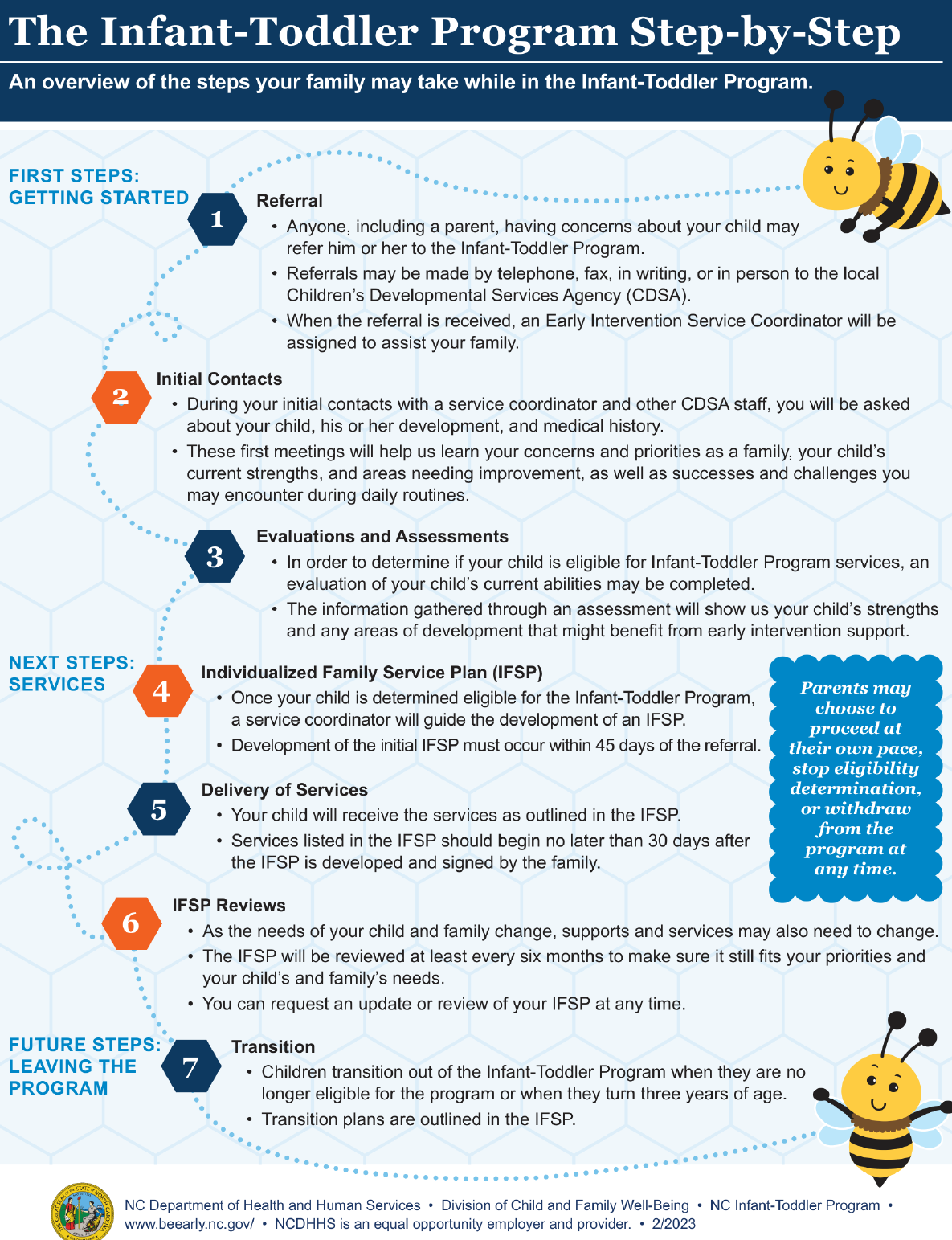}
    \caption{Source page (NC DHHS infographic). 3-level hierarchy with bold step titles and italic callout.}
    \label{fig:fmt-source}
\end{subfigure}
\hfill
\begin{subfigure}[t]{0.48\textwidth}
    \centering
    \includegraphics[width=\textwidth,page=1]{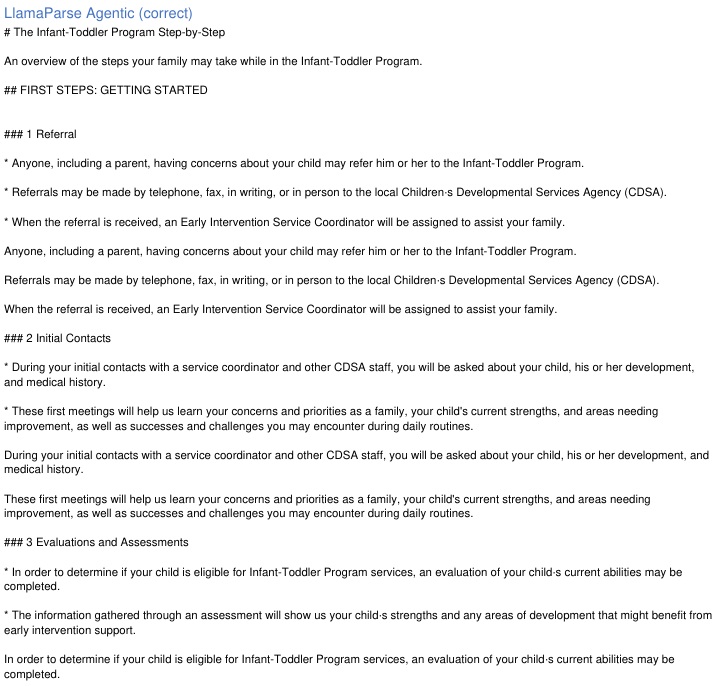}
    \caption{LlamaParse Agentic (1.000).}
    \label{fig:fmt-ours}
\end{subfigure}

\vspace{0.6em}
\begin{subfigure}[t]{0.31\textwidth}
    \centering
    \includegraphics[width=\textwidth,page=1]{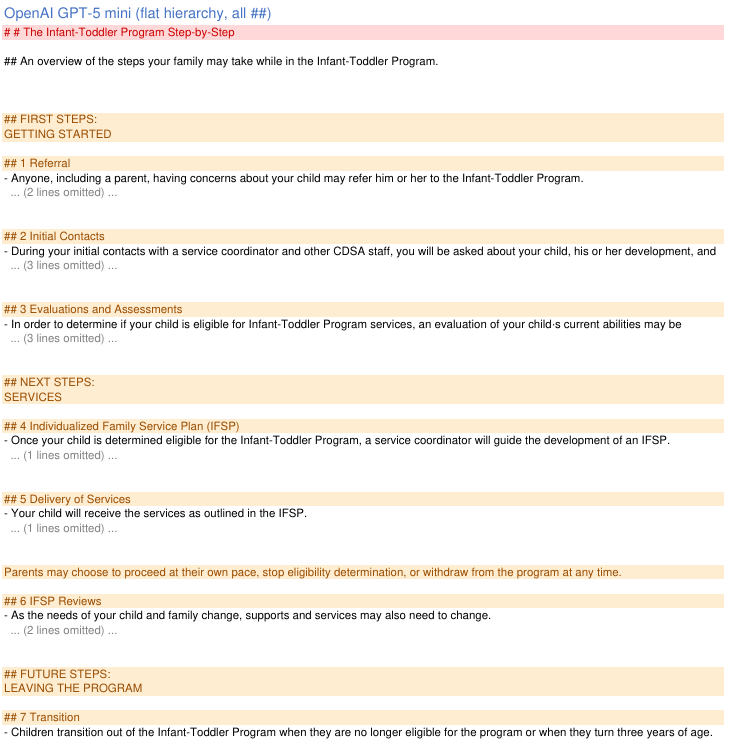}
    \caption{GPT-5 Mini (0.829).}
    \label{fig:fmt-gpt5}
\end{subfigure}
\hfill
\begin{subfigure}[t]{0.31\textwidth}
    \centering
    \includegraphics[width=\textwidth,page=1]{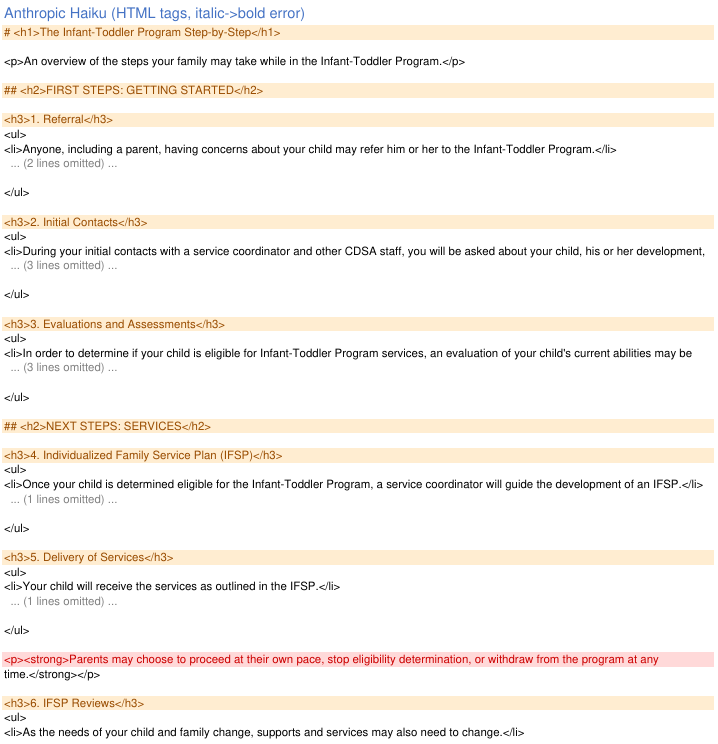}
    \caption{Haiku~4.5 (0.174).}
    \label{fig:fmt-haiku}
\end{subfigure}
\hfill
\begin{subfigure}[t]{0.31\textwidth}
    \centering
    \includegraphics[width=0.67\textwidth,page=1]{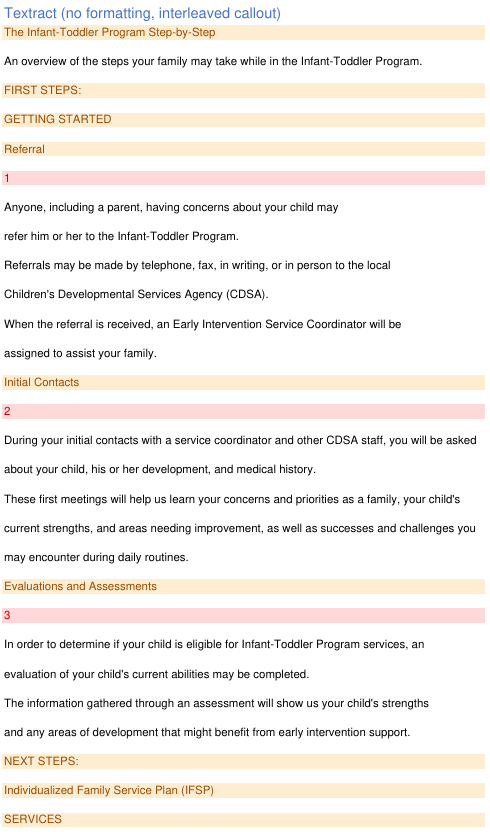}
    \caption{Textract (0.000).}
    \label{fig:fmt-textract}
\end{subfigure}
\caption{Source page and provider outputs for the infographic example. (a)~Source page with 3-level heading hierarchy. (b)--(e)~Rendered outputs. LlamaParse Agentic preserves hierarchy and bold titles; GPT-5 Mini flattens headings; Haiku~4.5 mixes HTML with Markdown; Textract produces plain text.}
\label{fig:fmt-outputs}
\end{figure}

We illustrate the Semantic Formatting evaluation on a single-page infographic (\texttt{text\_misc/stepbystep}, \Cref{fig:fmt-source}): an NC DHHS guide with a clear 3-level heading hierarchy (main title, section headers in all caps, numbered step titles in bold) and one italic callout.
Most providers extract the text content correctly, but fail to preserve heading structure and inline styling.
This contrast between high content faithfulness and low semantic formatting is precisely why both metrics are needed: correct text is not useful if downstream agents cannot distinguish section headers from body text, or bold labels from plain content.

\paragraph{Metric overview.}
Semantic Formatting (\Cref{sec:semantic-formatting}) averages two sub-scores: \emph{text styling} (are bold and italic markers preserved?) and \emph{title accuracy} (are headings detected with correct hierarchy?). This document has 7 \texttt{is\_bold} rules for step titles, 1 \texttt{is\_italic} rule for the callout, 11 \texttt{is\_title} rules, and 1 \texttt{title\_hierarchy} rule.

\paragraph{Results overview.}
\Cref{tab:fmt-scores} shows per-provider scores. LlamaParse Agentic achieves a perfect 1.0; all competitors lose heading hierarchy, text styling, or both.

\begin{table}[h]
\centering
\small
\begin{tabular}{@{}lrrr@{}}
\toprule
\textbf{Provider} & \textbf{Semantic Formatting} & \textbf{Text Styling} & \textbf{Title Accuracy} \\
\midrule
LlamaParse Agentic & \textbf{1.000} & \textbf{1.000} & \textbf{1.000} \\
GPT-5 Mini         & 0.829 & 1.000 & 0.658 \\
Haiku~4.5          & 0.174 & 0.000 & 0.348 \\
Textract           & 0.000 & 0.000 & 0.000 \\
\bottomrule
\end{tabular}
\caption{Per-provider Semantic Formatting scores on the infographic example. GPT-5 Mini preserves bold styling but flattens the heading hierarchy; Haiku~4.5 uses HTML tags instead of Markdown; Textract produces plain text with no formatting at all.}
\label{tab:fmt-scores}
\end{table}

\paragraph{Provider output comparison.}
\Cref{fig:fmt-outputs} shows the rendered outputs. Despite containing largely the same text content, the four outputs differ dramatically in structural markup.

\begin{itemize}
    \item \textbf{LlamaParse Agentic} (\Cref{fig:fmt-ours}, 1.000): Correct 3-level hierarchy (\texttt{\#} main title, \texttt{\#\#} section headers, \texttt{\#\#\#} numbered steps). All 7 step titles bold, italic callout preserved.

    \item \textbf{GPT-5 Mini} (\Cref{fig:fmt-gpt5}, 0.829): Text styling is near-perfect: all 7 bold step titles and the italic callout are correctly marked. The remaining gap is structural: the model flattens the 3-level heading hierarchy, reducing the title accuracy component even though the underlying text content is faithfully extracted.

    \item \textbf{Haiku~4.5} (\Cref{fig:fmt-haiku}, 0.174): Outputs HTML tags (\texttt{<h3>}, \texttt{<strong>}) inside Markdown instead of using Markdown syntax. The bold detection rules expect \texttt{**text**}, so the HTML-wrapped step titles score zero. The italic callout is marked as \texttt{<strong>} (bold) instead of \texttt{<em>} (italic).

    \item \textbf{Textract} (\Cref{fig:fmt-textract}, 0.000): Pure OCR output with no Markdown syntax. No headings, no bold, no italic. Step numbers are detached from titles (``1'' on one line, ``Referral'' on the next), and the italic callout is interleaved with adjacent main text due to row-wise reading of the side box.
\end{itemize}

\paragraph{Takeaway.}
This example highlights the gap between content extraction and structural understanding:
\begin{enumerate}
    \item \textbf{Content faithfulness $\neq$ semantic formatting.} Both Haiku~4.5 and Textract extract most of the text correctly (high content faithfulness), yet score 0.174 and 0.000 on semantic formatting. The text is there, but the structure is lost. For an agentic workflow that needs to navigate a document by section or identify key terms by formatting, this distinction is critical.
    \item \textbf{Subtle errors compound.} Haiku~4.5 uses HTML tags inside Markdown, a small implementation choice that zeroes out the entire text styling score. Textract strips all formatting entirely. Neither failure is an OCR error; both providers ``see'' the text correctly but fail to encode its structure.
\end{enumerate}

\subsection{Visual Grounding: Corporate Annual Report}
\label{app:qual-layout}

\begin{figure}[!ht]
\centering
\begin{subfigure}[t]{0.48\textwidth}
    \centering
    \includegraphics[width=\textwidth]{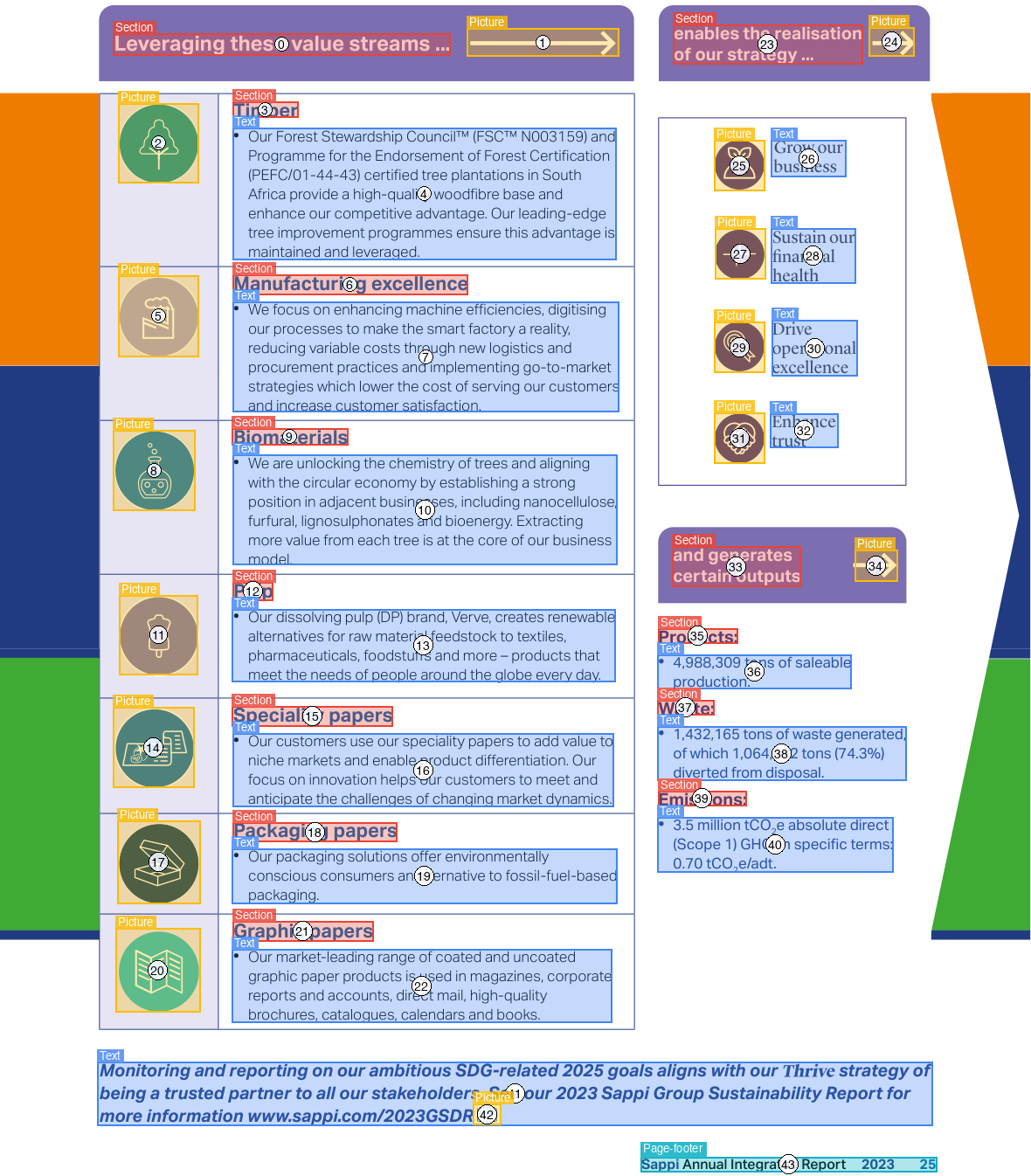}
    \caption{Ground truth (44 elements across 4 classes).}
    \label{fig:layout-source}
\end{subfigure}
\hfill
\begin{subfigure}[t]{0.48\textwidth}
    \centering
    \includegraphics[width=\textwidth]{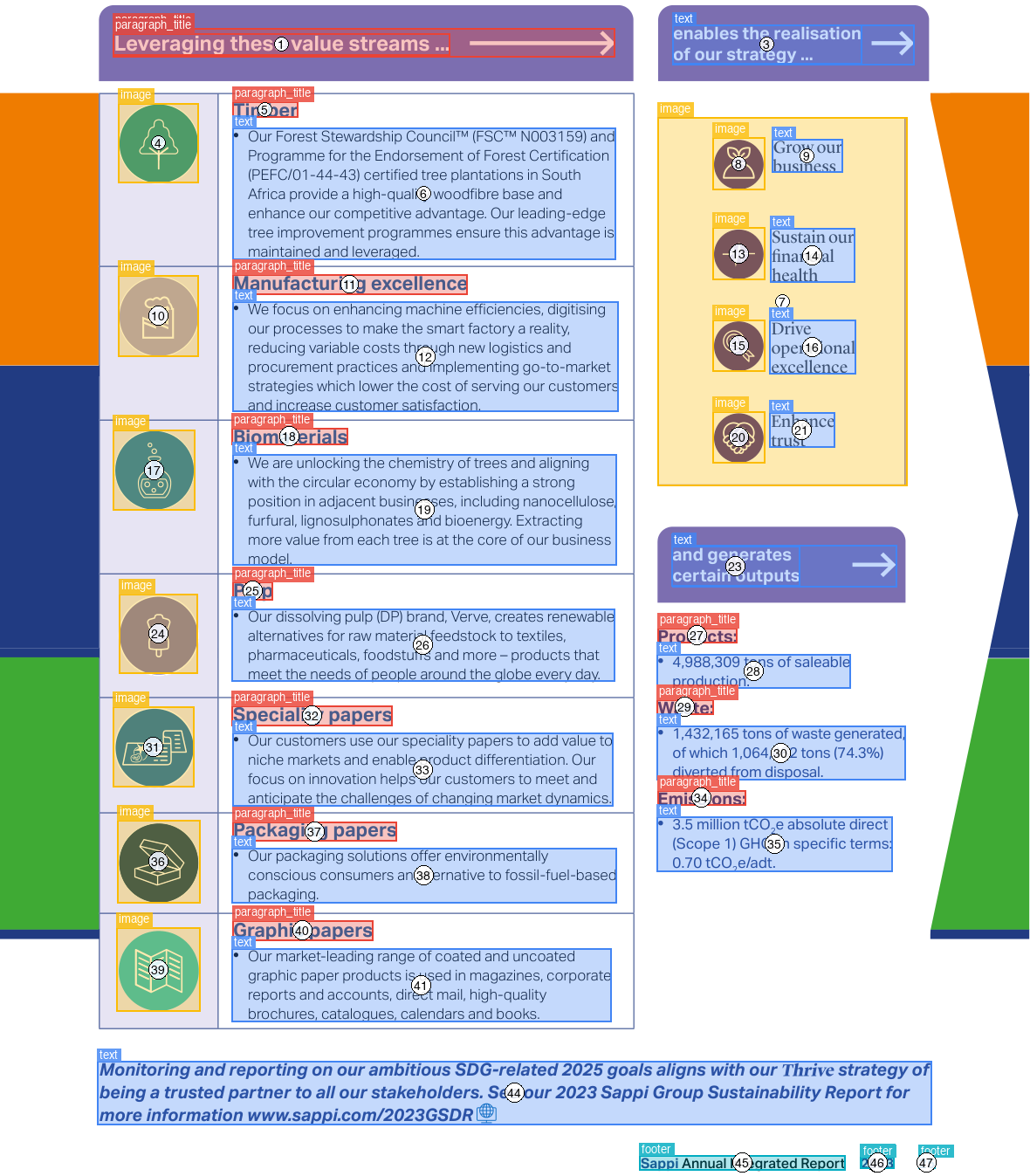}
    \caption{LlamaParse Agentic (86.4\%).}
    \label{fig:layout-ours}
\end{subfigure}

\vspace{0.6em}
\begin{subfigure}[t]{0.31\textwidth}
    \centering
    \includegraphics[width=\textwidth]{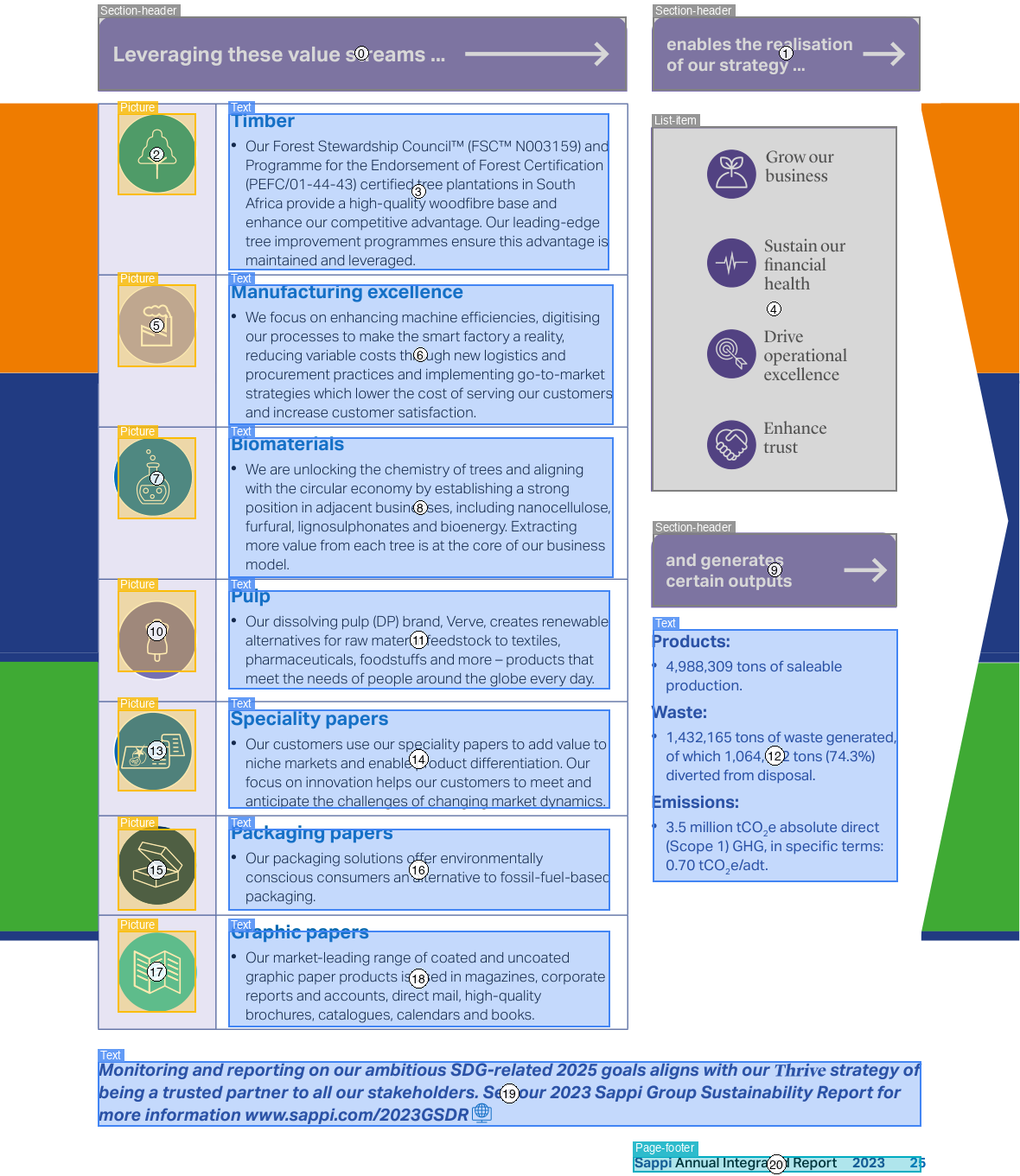}
    \caption{Gemini 3 Flash (43.2\%).}
    \label{fig:layout-gemini}
\end{subfigure}
\hfill
\begin{subfigure}[t]{0.31\textwidth}
    \centering
    \includegraphics[width=\textwidth]{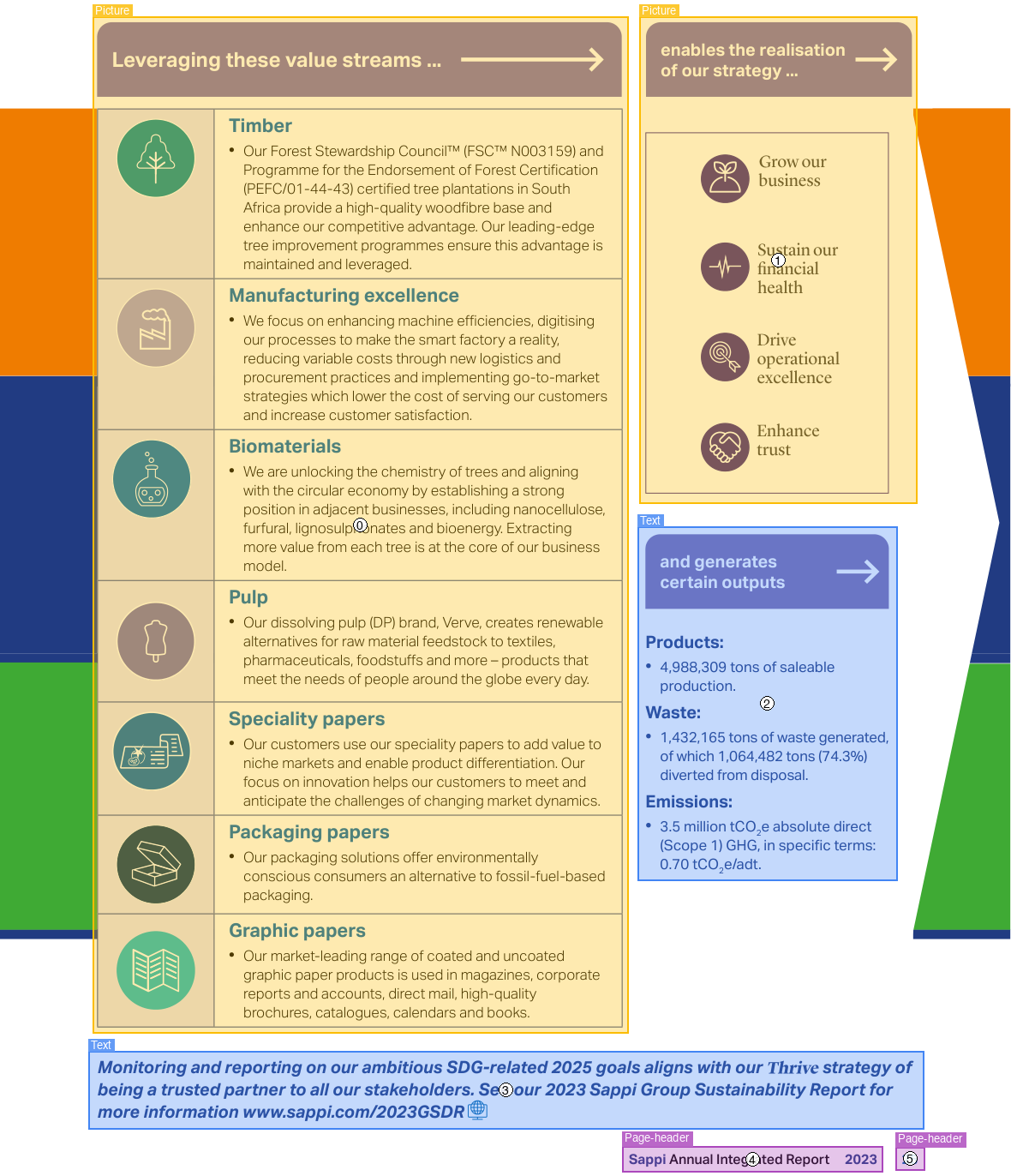}
    \caption{\landingainame{} (2.3\%).}
    \label{fig:layout-landingai}
\end{subfigure}
\hfill
\begin{subfigure}[t]{0.31\textwidth}
    \centering
    \includegraphics[width=\textwidth]{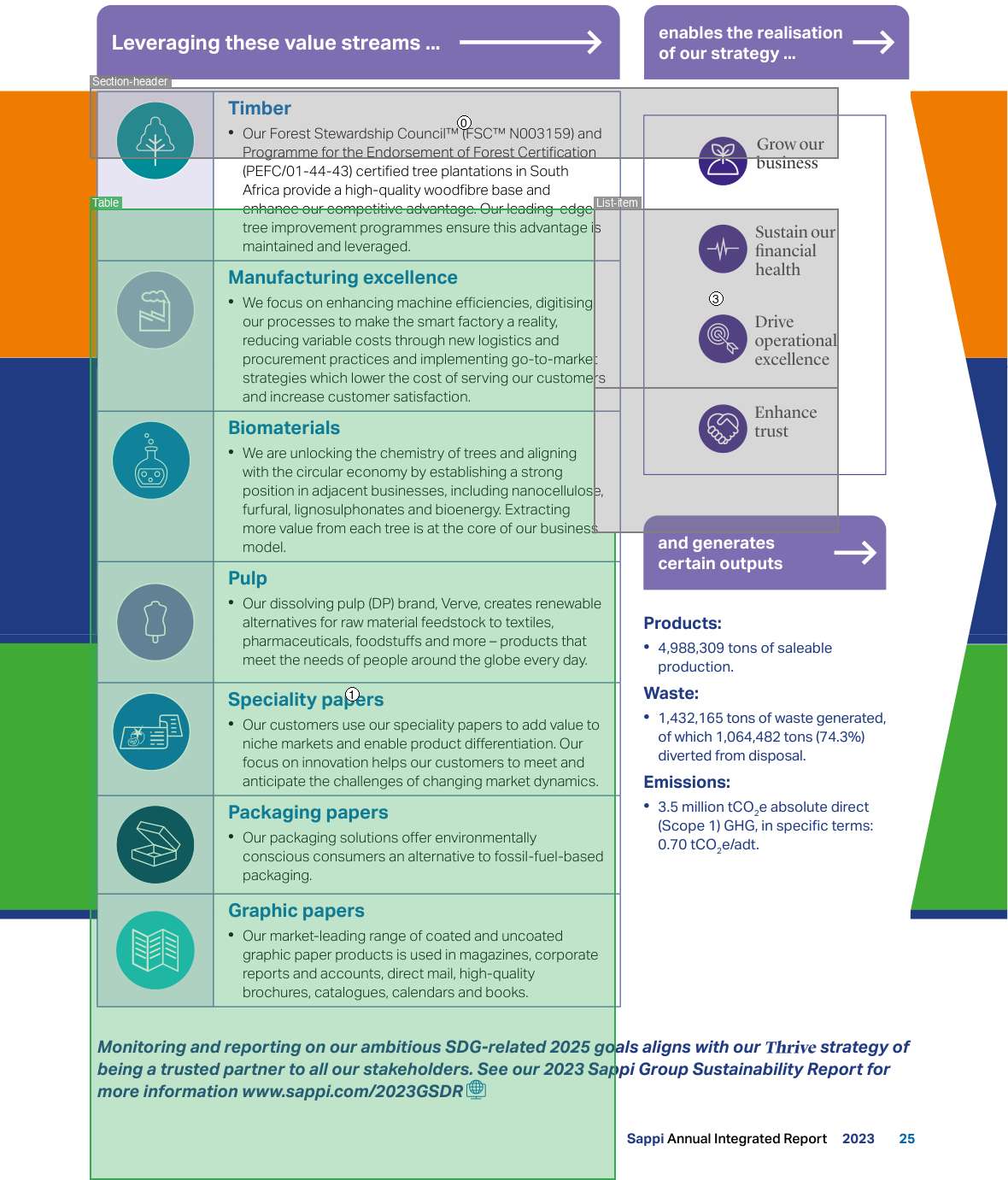}
    \caption{Haiku~4.5 (0.0\%).}
    \label{fig:layout-haiku}
\end{subfigure}
\caption{Ground truth and predicted layout overlays for the Sappi annual report page. (a)~Ground truth with 44 elements across \texttt{Picture}, \texttt{Text}, \texttt{Section}, and \texttt{Page-footer} classes. (b)--(e)~Provider predictions. LlamaParse Agentic produces 48~predictions closely matching GT; Gemini detects only 21; \landingainame{} and Haiku~4.5 produce only 6~each.}
\label{fig:layout-outputs}
\end{figure}

We illustrate the Visual Grounding evaluation on a visually rich page from the 2023 Sappi Annual Integrated Report (\texttt{sappi\_annual\_report}, \Cref{fig:layout-source}).
The page contains 44 annotated layout elements spanning four semantic classes: 15~\texttt{Picture} elements (circular certification and sustainability icons), 15~\texttt{Text} blocks (section descriptions and data summaries), 13~\texttt{Section} headers (``Timber,'' ``Manufacturing excellence,'' etc.), and 1~\texttt{Page-footer}.
This dense, multi-region layout with icons, colored sidebars, and mixed text/graphic content is challenging for visual grounding systems.

\paragraph{Metric overview.}
The Element Pass Rate (\Cref{sec:grounding}) evaluates each ground truth element against the parser's predictions via three sequential checks, all of which must pass for the element to count as correctly detected:
\begin{enumerate}
    \item \textbf{Localization:} The predicted bounding box must overlap the ground truth box with Intersection-over-Area $\geq 0.50$ from the ground truth perspective and $\geq 0.20$ from the prediction perspective.
    \item \textbf{Classification:} The predicted element type must match the ground truth class (e.g., \texttt{Picture}, \texttt{Text}, \texttt{Section}).
    \item \textbf{Attribution:} For elements with annotated text content (30 of 44 here), the extracted text must achieve token-level F1 $\geq 0.80$ against the ground truth text.
\end{enumerate}
Reading order is evaluated separately: for each element that passes both localization and attribution, the system checks whether its relative position is consistent with the ground truth reading order within a local neighborhood of 3 elements.

\paragraph{Results overview.}
\Cref{tab:layout-scores} shows per-provider scores on this example. LlamaParse Agentic achieves 86.4\% element pass rate with near-perfect localization and attribution; the three competitors shown fail to detect most elements.

\begin{table}[h]
\centering
\small
\begin{tabular}{@{}lrrrr@{}}
\toprule
\textbf{Provider} & \textbf{Element Pass Rate} & \textbf{Localization} & \textbf{Classification} & \textbf{Attribution} \\
\midrule
LlamaParse Agentic      & \textbf{86.4\%} & \textbf{93.2\%} & \textbf{86.4\%} & \textbf{100.0\%} \\
Gemini 3 Flash          & 43.2\% & 45.5\% & 45.5\% & 62.1\% \\
\landingainame               & 2.3\%  & 4.5\%  & 2.3\%  & 3.4\%  \\
Haiku~4.5               & 0.0\%  & 0.0\%  & 0.0\%  & 0.0\%  \\
\bottomrule
\end{tabular}
\caption{Per-provider Element Pass Rate and sub-metric scores on the Sappi annual report page (44 ground truth elements). The element pass rate requires all three sub-checks to pass simultaneously. Gemini 3 Flash detects some elements but misses over half; \landingainame{} and Haiku~4.5 produce only 6 predictions each and fail almost entirely.}
\label{tab:layout-scores}
\end{table}

\paragraph{Provider output comparison.}
\Cref{fig:layout-outputs} shows the predicted bounding box overlays from each provider. The number of predicted elements varies dramatically: LlamaParse produces 48 predictions (close to the 44 ground truth elements), while the three competitors produce far fewer.

\begin{itemize}
    \item \textbf{LlamaParse Agentic} (\Cref{fig:layout-ours}, 86.4\%): Produces 48 predictions for 44 ground truth elements. Localization is strong at 93.2\%, and attribution is perfect at 100.0\%---every detected element contains the correct text content. The 86.4\% classification rate reflects a few element-type confusions (e.g., section headers misclassified as text). Reading order reaches 79.3\%, with minor ordering inconsistencies in the densely packed right-side data columns.

    \item \textbf{Gemini 3 Flash} (\Cref{fig:layout-gemini}, 43.2\%): Detects only 21 elements out of 44, missing more than half the page content. The model entirely fails on \texttt{Section} elements (F1${}=0.0$): none of the 13 section headers (``Timber,'' ``Manufacturing excellence,'' ``Products,'' etc.) are detected as distinct elements. It captures some pictures and text blocks but merges others into oversized bounding boxes. Attribution drops to 62.1\%, indicating that even for located elements, the extracted text often does not match the ground truth content.

    \item \textbf{\landingainame} (\Cref{fig:layout-landingai}, 2.3\%): Produces only 6 predictions---large bounding boxes that each span multiple ground truth elements. Of the 44 ground truth elements, 42 remain unmatched entirely. The few predicted boxes have low spatial overlap with individual GT elements and contain merged text from adjacent regions. Reading order scores 0.0\% since almost no elements pass the localization and attribution prerequisites.

    \item \textbf{Haiku~4.5} (\Cref{fig:layout-haiku}, 0.0\%): Also produces only 6 predictions, but none achieve sufficient bounding box overlap with any ground truth element. All 44 GT elements are unmatched: localization, classification, and attribution all score exactly 0.0\%. The predicted boxes are coarse page-level regions that do not correspond to individual layout elements, and the model fails to decompose the complex page into its constituent parts.
\end{itemize}

\paragraph{Takeaway.}
This example illustrates two key properties of the visual grounding evaluation:
\begin{enumerate}
    \item \textbf{Granularity matters.} The evaluation rewards fine-grained element detection: each of the 44 ground truth elements must be individually localized, correctly classified, and verified for text content. Producing a few large bounding boxes that cover the page area scores near zero because those boxes do not correspond to individual semantic elements. Both \landingainame{} and Haiku~4.5 detect roughly the right page regions but fail completely because they cannot decompose those regions into constituent elements.
    \item \textbf{Cascading requirements expose compound failures.} The three-stage evaluation (localization $\rightarrow$ classification $\rightarrow$ attribution) means that errors compound: an element that is well-localized but misclassified still fails. Gemini~3 Flash localizes about half the elements but scores only 43.2\% because many localized elements are misclassified or contain wrong text. This cascading design reflects the real-world requirement that a layout element is only useful if it is in the right place, has the right type, \emph{and} contains the right content.
\end{enumerate}

\end{document}